\pdfoutput=1

\documentclass{article}

\usepackage{arxiv}

\usepackage[colorlinks=true, allcolors=black]{hyperref}

\input{preamble}

\hypersetup{
pdftitle={Iconic Gesture Semantics},
pdfauthor={Andy Lücking, Alexander Henlein, Alexander Mehler},
pdfkeywords={Iconic Gesture Semantics, Vector Space Semantics, Spatial Semantics,
Speech–Gesture Integration, Multimodal Meaning, Dual Coding, Exemplification, Extemplification, Perceptual Classification, Speech–Gesture Integration, Semantic Theory}
}

\begin{document}

\title{The Spatial Semantics of Iconic Gestures}

\author{Andy Lücking \\
	Department of Computer Science\\
	Goethe University Frankfurt\\
	Frankfurt a/M., D-60325, Germany \\
	\texttt{luecking@em.uni-frankfurt.de} \\
	\And
	Alexander Henlein \\
	Department of Computer Science\\
	Goethe University Frankfurt\\
	Frankfurt a/M, D-60325, Germany \\
	\texttt{henlein@em.uni-frankfurt.de} \\
	\AND
	Alexander Mehler \\
	Department of Computer Science\\
	Goethe University Frankfurt\\
	Frankfurt a/M, D-60325, Germany \\
	\texttt{mehler@em.uni-frankfurt.de} \\
}

\date{November 28, 2025}


\maketitle

\begin{abstract}%
The current multimodal turn in linguistic theory leaves a crucial question unanswered: what \emph{is} the meaning of iconic gestures, and how does it compose with speech meaning? 
We argue for a separation of linguistic and visual levels of meaning and introduce a spatial gesture semantics that closes this gap.
Iconicity is differentiated into three aspects: Firstly, an interpretation of the form of a gesture in terms of a translation from kinematic gesture annotations into vector sequences (\emph{iconic model}).
Secondly, a truth-functional evaluation of the iconic model within spatially extended domains (\emph{embedding}).
Since a simple embedding is too strong, we identify a number of transformations that can be 
applied to iconic models, namely rotation, scaling, perspective fixation, and quotation of handshape.
%
%
%
Thirdly, the linguistic description or classification of an iconic model (\emph{informational evaluation}). 
Since the informational evaluation of an iconic gesture is a heuristic act, it needs a place in a semantic theory of visual communication.
Informational evaluation 
lifts a gesture to a quasi-linguistic level that can interact with verbal content. 
This interaction is either vacuous, or regimented by usual lexicon-driven inferences discussed in dynamic semantic frameworks.
\end{abstract}




\tableofcontents     


\section{Introduction}
\label{sec:main-introduction}

Speech and gesture interact semantically, as has been pointed out frequently \citep[see][for two seminal sources]{McNeill:1992,Kendon:2004}, and is, for instance, most clearly shown by examples where the gesture can be construed in such a way that it adds to speech meaning, as in (\nextx), where the gesture can be understood to provide shape information specifying the type of the staircases talked about:\footnote{The example is taken from dialogue V10 from the SaGA corpus \citep{Luecking:Bergmann:Hahn:Kopp:Rieser:2010}, starting at minute 3:19.} 
\ex 
\enquote{Ich g[laube das \staircases[1.5cm] sollen TREP]pen sein} (\textit{I think that should be staircases}; capitalization indicates main stress of the first syllable of the noun \textit{Treppen} \enquote*{staircases}, square brackets indicate the temporal alignment of speech and gesture) 
\label{ex:spiral-treppen}
\xe 

If the gesture is construed in such a way that it makes the concept \textit{wounded} or \textit{spiral} salient, then there is a strong inclination to interpret the noun \textit{staircases} in terms of its hyponym \textit{spiral staircases}.
The modest conditional formulation -- \emph{if the gesture is construed in such a way that \ldots} -- is indeed part of the iconic gesture semantics developed in the following. 
We argue that the \emph{informational evaluation} of a gesture lifts its visual properties to quasi-linguistic status, which can interact with natural language semantics and reasoning in the first place (we argue further that this reasoning is mainly driven by lexical knowledge).
This is because gestures and words are quite different things:
%
It is widely assumed that the meaning of natural languages are conventionalized, largely arbitrary systems, while iconic gestures are driven by non-arbitrary, visuo-spatial properties.
The importance of differentiating between description and depiction is also emphasized in the Cambridge Element by \citet{Davidson:2025-depiction}.
%
This ambivalence -- the verbal and nonverbal characteristics of iconic gestures -- has also divided gesture semantics over the past 25 years or so:\footnote{We consider \citet{Kuehnlein:1999} and \citet{Rieser:2004} to be a useful starting points for formal multimodal semantics.} 
On the one hand, there are gesture semantics that assume a shared semantic representation of speech and gesture, namely the predicate constants of the formal semantic representation language \citep{Lascarides:Stone:2009:a}.
This is obviously a word-like construal of iconic gesture meaning. 
On the other hand, \citet{Giorgolo:2010-phd} developed a mereotopological model of the interpretation of iconic gestures, accounting for the nonverbal, visual nature of iconic gestures.\footnote{The status of other works is less clear. \citet{Rieser:Poesio:2009}, for instance, pursue a labelling approach but restrict the gestural predicates to spatial ones, which give rise to partial ontology descriptions \citep[see also][]{Hahn:Rieser:2010}. Hence, they use labels to describe spatial configurations.  
\citet{Luecking:2013:a} employs a spatial, vector-based model, but only to the effect of deriving verbal labels therefrom. This shows that both, labelling and spatial, views should be kept apart and combined in a unified approach, which is one of the goals of this paper. We discuss pictorial approaches in \cref{subsec:projective-semantics}.} 

The difference between both approaches can be illustrated by means of the gesture from (\lastx):
The first approach -- \emph{labelling theories} -- assigns semantic predicates to the gesture which express its meaning(s)\footnote{Since gestures are considered vague or underspecified, a single gesture usually is compatible to several predicates \citep{Lascarides:Stone:2009:a}.} (\nextx a), the second one -- \emph{spatial theories} -- models the meaning of the gesture in terms of a spatial structure, or iconic model (\nextx b).
\pex 
\a labelling theories: $\textit{spiral}(x) \vee \textit{curled}(x) \vee \textit{twined}(x) \vee\ldots $
\a spatial theories: \coil
\xe 

We conjecture that this ambivalence is due to the  \enquote{twofoldedness} of gestures diagnosed in gesture research: \enquote{as we see \emph{them}, we see something \emph{in them}.} \citep[p.~286; original emphasis]{Streeck:2008}.
We aim at an explanation and a reconciliation of this ambiguity, since from a theoretical perspective, this situation is unsatisfactory for several reasons: 

Firstly, labelling theories -- which, following the format of  \citet{Schlenker:2018} and colleagues \citep[e.g.,][]{Ebert:Ebert:Hoernig:2020}, currently appear to be predominant -- lack a principled way of assigning predicates to iconic gestures; semantic representations are brought about \enquote{by hand}, merely following (not necessarily justified, as will be shown below) interpretive assumptions.\footnote{\enquote{It should be emphasized that we will not seek to explain how a gesture [\ldots]
comes to have the content that it does} \citep[p.~296]{Schlenker:2018}. As will become evident, even the \enquote{that it does} part is not necessarily justified nor innocent.} 

Secondly, spatial theories (including pictorial semantics based on  \citet{Greenberg:2021}, see \cref{subsec:projective-semantics}) do not provide a computational procedure for deriving iconic models from kinematic gesture representations and combine it with speech meaning.

Thirdly, the meaning representations in (\lastx a) and (\lastx b) are obviously related in a non-arbitrary way.
However, the relation between labelling and spatial theories has not been addressed in gesture semantics so far.
%
%
We provide the basis of a unified iconic gesture semantics that takes these desiderata into account.
%

Our starting point is a spatial theory, namely a vector space model, following work by \citet{Zwarts:1997,Zwarts:Winter:2000,Zwarts:2017}  (\cref{sec:main-spatial-gesture-semantics}).
Our first contribution is a formal notion of \emph{gesture space} modelled as an oriented vector space in terms of an adaptation of vector space semantics. 
We then show how kinematic gesture representations can be interpreted in terms of vector sequences in gesture space (addressing the second issue raised above).

Our second contribution is a shift in semantic architecture: we propose in \cref{sec:main-informational-evaluation} -- following work in computational semantics -- to employ vector spaces as models of \emph{intensions} of certain lexical items. 
This move paves the way for explaining the relation between labelling and spatial theories (third issue): iconic models and semantic predicates are linked \textit{via} a semiotic relation reminiscent of \emph{exemplification} \citep{Goodman:1976}.
We will argue that in order to maintain the semiotic nature of iconic gestures (i.e., avoiding treating them as \enquote{mere} objects or events), a modified notion of exemplification is needed. 
This is perhaps the most consequential contribution:
%
The notion of meaning in mainstream possible worlds semantics, $\llbracket \cdot \rrbracket$, is such that the meaning $\llbracket \alpha \rrbracket$ of a word $\alpha$ when applied to a world $w$ returns $\alpha$'s extension in $w$.
%
The notion of meaning emerging from extended exemplification (\enquote{extemplification}
), to the contrary, has to be such that when applied to an object (in some world) it returns a linguistic label for that object.
A consequence of this view is that the extemplification part of iconic gesture semantics cannot be fully spelled out within a Frege/Montague framework. 
We therefore offer a heuristic that is primarily intended for labeling approaches to interpreting gestures within possible worlds semantics.

\section{Spatial Gesture Semantics}
\label{sec:main-spatial-gesture-semantics}

If gestures contribute meaning in the same way as words, either at-issue or non-at-issue, we would expect this meaning to be just as addressable as the meaning of words.
Addressing meaning is a primary task of clarification interaction. 
Research on clarification requests have shown, among others, that reprise fragments are a common means of requiring information about the source \citep{Purver:2004-phd,Purver:Ginzburg:2004}.
Reprise fragment clarification requests obey strict categorical and phonological parallelism with their source \citep{Ginzburg:Cooper:2004}.
This is shown in (\nextx).
\pex 
\a A: Do you fear him? \\ B: Fear? / \#Afraid?
\a A: The teacher phoned him. \\ B: Him? / \#He? \\ C: Phoned? / \#Called?
\a A: Were you cycling yesterday? \\ B: Cycling? / \#Cycled?
\xe 

Such parallelism constraints generalize multimodally \citep{Ginzburg:Luecking:2021-clarifications}. 
\pex 
\a B: You have to move your legs like this [\textit{moves right hand up and down in a wave-like manner}].\\ 
A: [\textit{moves right hand up and down in a wave-like manner, raises eye-brows}] (constructed from a kids TV show)
\a A: I hear you're busy $\langle$laughter$\rangle$ [$=$ little giggle]. \\ 
B: $\langle$laughter$\rangle$? ($=$ low arousal laughter with rising contour). (attested example)
\xe 

To address the antecedent manual gesture (\lastx a) respectively the laughter (\lastx b), B has to reproduce the formal properties of A's initial non-verbal signals.
For instance, moving the hand in (\lastx a) from right to left instead of up and down fails to instantiate the intended multimodal clarification request.

Parallelisms constraints also govern denials: the sub-utterance triggering an objection is identified by categorical and phonological uptake:
\ex 
A: The teacher phoned him. \\
B: No, that's not true, the teacher didn't phone / \#phoning / \#call him!
\xe 
B objects A on the grounds that B assumes that the teacher either did not contact the referent of \textit{him}, or contacted him by means other than telephone. 
Such denial tests play an important role in detecting a gesture's meaning contribution, if any.

\subsection{The need for a spatial semantics}
\label{sec:introduction}


%
The second gesture in (\nextx) -- the one associated with the noun \textit{Dach} \enquote*{roof} -- can be understood to provide shape information of the roof under discussion (the datum also gives the previous context, including the gesture associated with the noun \textit{Stangen} \enquote*{poles}; we use two dots, \enquote{..}, to indicate a brief pause; an English translation is given in brackets following the datum):\footnote{The example is taken from dialogue V11 from the SaGA corpus \citep{Luecking:Bergmann:Hahn:Kopp:Rieser:2010}, starting at minute 2:45. Here and in the following we try to use attested, empirical data.} 
%
\ex 
so'ne dreiteilige Eingangstür .. mit nem schwarzem Vordach  .. das Vordach sind so  \includegraphics[width=1.5cm]{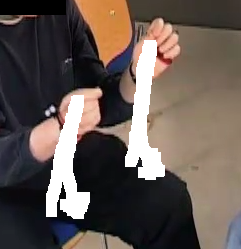}[\textit{up and down movement of both hands}]  zwei .. äh .. Stangen mit nem \dachdrueber{[\textit{straight movement to right of right hand}]} Dach drüber halt \\
(\textit{a three-part entrance door .. with a black canopy .. the canopy is simply \textup{[first gesture]} two .. uh .. poles with a \textup{[second gesture]} roof over them })
\label{ex:roof}
\xe 

Intuitively, the first gesture in (\lastx) depicts the two poles talked about and does not add much if anything to speech meaning,
the second one arguably depicts the shape of the roof talked about, or depicts that the two poles are connected.
%
%
But the latter, linguistic descriptions are acts of interpretation in their own right: gestures as visual, bodily movements 
do not make any \emph{linguistic} contribution.
%
Clear evidence for this is obtained from denial tests: it is simply not possible to target linguistic at-issue (\nextx a) or non-at-issue (\nextx b) 
contributions of gestures (question marks indicate examples whose interpretation is problematic), whereas a multimodal repair with another gesture \emph{is} possible (\nextx c). 
\pex {[\ldots]} with a \dachdrueber roof over them \label{ex:roof2}
\a \ljudge{?} No, that's not true. The roof (i) is not \placeholder / (ii) actually is \placeholder 
\a \ljudge{?} Wait a minute. The roof (i) is not \placeholder / actually is \placeholder
\a No, that's not true, the roof (i) is not \includegraphics[width=1.5cm]{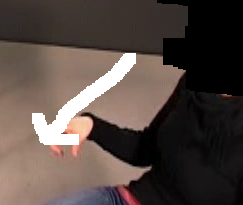} / (ii) actually is \includegraphics[width=1.5cm]{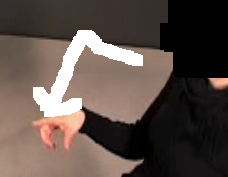}.
\xe 

The marked examples are problematic: 
%
%
What can be said in the places marked with the placeholder \enquote*{\placeholder}? 
There is no clear answer to these questions, since the gesture has not asserted or presupposed 
anything (asserting and presupposing 
being linguistic acts).
Furthermore, any linguistic expression that fills the placeholder slot would trivially violate the parallelism constraint known from clarification interaction (see above). 
The application of linguistic denial tests to non-linguistic, visual communication means like gestures requires first and foremost a linguistic interpretation of the gestures.
This can be shown using data from dialogue, where interlocutors A and B first agree on a linguistic interpretation of the gesture (\nextx b,c), which can then be rejected (\nextx d) (or accepted).
\pex \label{ex:roof-dialogue}
\a A: {[\ldots]} with a \dachdrueber roof over them 
\a B: By \includegraphics[width=1.5cm]{dach-gerade_from-V11-6-43.png}, do you mean that (i) the roof is a flat one / (ii) extends to the right?
\a A: Yes.
\a B: But that's not true. The roof (i) is not flat / (ii) does not extend to the right. 
\xe 

There is, in other words, a \enquote{semiotic gap} between the visual realm of iconic gestures and the symbolic realm of language.
%
This gap can be bridged, however, by agreeing on a linguistic interpretation of a gesture in question, as in (\lastx).
%

As evinced in (\ref{ex:roof2}-c), gestures can be addressed on the visual level.
The multimodal pair of noun and gesture can be denied with a gestural uptake (\ref{ex:roof2}-c(i)), or figure as the object of a gestural repair (\ref{ex:roof2}-c(ii)). 
Hence, an interaction of speech and gesture in clarification interaction is possible as long as the gesture keeps its \enquote{visual innocence}, that is, as long as the gesture is not treated as a linguistic expression, but matched by another gesture. 

Evidence for keeping apart a visual and a linguistic level of meaning gains further support from the anaphoric potential of a multimodal utterance, that is, its possible anaphoric continuations. 
While verbal constituents can be picked out by pronouns in subsequent utterances, the gesture cannot, as shown in (\nextx) by example of the English translation of example (\ref{ex:roof}):\footnote{In formal semantics this is captured by keeping apart verbal and gestural discourse markers \citep[see][]{Lascarides:Stone:2009:a}.}
\pex {[\ldots]} two poles with a \dachdrueber roof over them \label{ex:roof-anaphora}
\a They ($=$ two poles) stand firm.
\a It ($=$ a roof) was covered in moss.
\a \ljudge{\#} It ($=$ \textit{gesture}) extends to the right. 
\xe 

 All in all, there is ample evidence that in addition to stable linguistic meaning, there is also a somewhat variable, \enquote{weaker} visual level of information.
The linguistic and visual meanings are not equivalent and one cannot simply switch from one to the other.
Accordingly, it is important to keep the two dimensions apart when trying to formalize the semantic integration of speech and gesture.
We do so in \cref{subsec:compositional-speech-gesture-integration} by distinguishing a linguistic and a visual dimension of logical forms.
The point of contact between the two are vector spaces (in terms of an embedding of a gesture's iconic model into the vector space projected from the referent of its affiliated expression).

Given the well-attested gap between the visual realm of gestures and the symbolic realm of linguistic expressions, the question arises as to how the equally well-attested observation of a semantic interaction between speech and gesture can be explained.
There are at least quite clear semantic intuitions about the circumstances under which a multimodal utterance is true that can guide semantic theorizing. 
Informally, the multimodal PP in (\lastx), for instance, is true of the modified NP if the extension \mng[e]{\text{NP}} of the NP in situation $e$ has a roof and the roof \enquote{looks like} the gesture. 
In the following, we develop a spell-out for the intuitive, informal notion of \textit{looking like} a gesture.
%
Based on the formal concept of vector space, a re-interpretation of kinematic annotation as descriptions of vectors is achieved: the \emph{iconic model} of an iconic gesture.\footnote{We use the term \enquote{model} to indicate the fact that the vectorial representation of a gesture abstracts over presumably irrelevant features such as skin colour etc.}
The iconic model of the roof gesture, for example, amounts to the abstract percept of a straight line running from right to left, as seen from the perspective of the hearer standing opposite. 
Visually:
\ex \label{ex:roof-line}
\straightline
\xe

To capture the visual contributions of gestures, we employ a spatial theory, namely a vector space semantics, following work by \citet{Zwarts:1997,Zwarts:Winter:2000,Zwarts:2017}.
%
%
%
The general architecture of the iconic gesture semantics is shown in \cref{fig:architecture1}.
Speech is interpreted by a denotation function within spatial models.
\begin{RevisionEnv}
Spatial models are models which impose spatial structure on the domain of entities and events.
The spatial structure is formulated in terms of vectors and vector sequences (paths) within vector spaces.
The kinematic representation of a gesture is translated via function \enquote*{$\gestvec$} into an iconic model.
The iconic model constrains the spatial projection of the entities talked about by the gestures verbal affiliate.
%
Iconicity is semantically differentiated in two aspects: a direct interpretation of a gesture's form via function \enquote*{$\gestvec$}, and an embedding of its output into spatial models. 
Overall, a spatial semantics is developed that respects the \enquote{visual innocence} of iconic gestures, in line with the above-given data. 
%
%
As a result, a semantic theory of iconic gestures developed that fulfills an urgent desideratum of linguistic theory.
The theory explains how visual information from gestures can enrich linguistic meaning while keeping both meaning dimensions apart. 
Each component of the theory will be introduced in the following sections. 
%

\begin{figure}
    \centering
    \begin{adjustbox}{max width=\linewidth}        
    \begin{tikzpicture}[
     Kasten/.style={rectangle, draw, very thick, minimum height=0.7cm},
     Const/.style={Kasten, fill=gray!30},
     Theory/.style={Kasten, rounded corners},
     node distance=1.3cm and 1.6cm,
     ]
        \node [Const] (gesture) {Gesture};
        \node [Const, right=of gesture] (verb) {Verbal constituent};
        \node [Theory, below=of gesture] (gesturespace) {Gesture space};
        \node [Theory, below=of gesturespace] (iconic) {Iconic model};
        \node [Theory, right=2cm of verb] (dom) {$D_e, D_s$}; 
        \path (iconic) -| (dom) node [midway, Theory] (vec) {$D_p, D_v$};

        \begin{scope}[->, thick]
        \draw (gesture) -- (gesturespace) node [midway, right] {$\gestvec$};
        \draw [dotted] (gesturespace) -- (iconic) node [midway, right] {produces};
        \draw [dashed] (gesture) -- (verb) node [midway, below] {affiliates};
        \draw (dom) -- (vec) node [midway, align=left, draw, fill=white] (vectors) {space\\ axis\\ place\\ axis-path \\ place-path};
        \draw (vec) -- (gesturespace) node [midway, above, sloped] {defines};
        \draw (vectors) -- (gesturespace)  node [midway, below] {defines};
        \end{scope}

        \node [rectangle, draw, rounded corners, fit=(dom) (vec) (vectors), inner sep=2ex, label={[align=left]right:{Spatial\\ Model}}] (model) {};
        
        \draw [->, thick] (verb) -- (verb -| model.west) node [midway, below] {\mng{\cdot}};
        \draw [->, thick] (iconic) -- (iconic -| model.west) node [midway, above] {embedding} node [midway, below, align=center] {\textit{via} scaling, rotation, \\ perspective, quotation};
     \end{tikzpicture}
     \end{adjustbox}
     \caption{Architecture of iconic gesture semantics. Verbal constituents are interpreted within spatial models. Spatial models relate entities ($D_e$) and events ($D_s$) onto points ($D_p$) and vectors ($D_v$) in terms of vector spaces, axes, places, and paths -- see \protect\cref{sec:spatial-theories} for details. The latter are used to define vectorial gesture spaces. Gestures are translated into these gesture spaces, producing formal iconic models that -- eventually modified by the operations of scaling, rotation, perspective, or quotation -- impose constraints on the spatial domain in which their affiliated verbal expressions are evaluated.}
    \label{fig:architecture1}
\end{figure}

\end{RevisionEnv}

\subsection{A brief gesture primer}
\label{sec:gesture-primer}

By \enquote*{gesture}, we refer to hand and arm movements that accompany speech and are related to the narrative. 
From these hand and arm movements, we are concerned with iconic ones (\cref{subsec:dimensions-of-classifying-gestures}). 
%
%
\Cref{subsec:conventionalization} discusses different degrees of conventionalization and explain them in terms of the number and/or strength of the coupling between lemmata and their imagistic content parts (anticipating results from \cref{sec:perceptual-classification}).
\Cref{subsec:empirical-data} collects some empirical examples that exemplify the range of gesture we aim to analyze.
\Cref{subsec:kinematic-gesture-representation} introduces a kinematic representation system for iconic gestures. 
Such a representation is required to satisfy the first demand of iconicity, namely an interpretation of gestures' forms.

\subsubsection{Dimensions of classifying gestures}
\label{subsec:dimensions-of-classifying-gestures}

We focus on iconic gestures, as opposed to emblematic, deictic, rhythmic, and discourse functional ones -- see \cref{fig:gesture-taxonomy} for an overview. 
We adopt Müller's quadrinomial nomenclature of \emph{modes of representation} of iconic gestures: 
%
%
\emph{Acting} classifies a miming action (i.e., pantomime). 
\emph{Molding} describes three-dimensional sculpturing, \emph{drawing} two-dimensional tracing.
In \emph{representing}, the hand/arm is a \Revision{stand-in} for an object.
%
%
\Revision{The modes of representation are semantically important since they provide information about a gesture's primary representational dimension and its most relevant kinematic features.
%
%
We will partly reconstruct modes of representation in terms of different kinds of vectors in spatial models, with a focus on {acting} and {drawing} ({molding} turn out to be a variant of {drawing} in semantic terms).
%
%
}

The semantic challenge posed by iconic gestures is that they are not regimented by fixed form--meaning associations, that is, a lexicon \citep[cf.][]{McNeill:1992}.
Hence, iconic gestures do not constitute a fixed class which can simply be interpreted by the interpretation function, \mng{\cdot}, for language.
%
%
A gesture occurrence, however, has a \enquote{docking point} in speech, namely its \emph{lexical affiliate} \citep{Schegloff:1984}.
%
The affiliate guides the interpretation of the gesture \citep{Hadar:Krauss:1991}.
The affiliates of the gestures in the initial example (\ref{ex:roof}), for example, are the nouns \textit{Stangen} \enquote*{poles} and \textit{Dach} \enquote*{roof}, respectively.
Note that while the affiliate was initially assumed to be a lexical item (i.e., a single word), it is lexical in only about 80\% of occurrences, the remainder exhibit syntactically more complex verbal attachment sites \citep{Mehler:Luecking:2012-pathways}.


\begin{figure}
\begin{adjustbox}{max width=\linewidth}
\begin{forest}
[
manual gesture
  [formal
    [emblematic]
    [deictic,
      [object]
      [direction]
      [abstract]
    ]
    [iconic, draw, l*=2
      [acting, name=acting, draw, dashed]
      [molding, name=molding, draw, dashed]
      [drawing, name=drawing, draw, dashed]
      [representing, name=representing]
    ]
    [metaphoric, l*=2, name=metaphor]
    [{rhythmic\\ (beat)}, align=center]
  ]
  [functional 
    [{dialogue\\ content}, align=center, l*=2]
    [{dialogue\\ management \\ (interactive)}, align=center, l*=2]
  ]
]
\end{forest}
\end{adjustbox}
\caption{Dimensions of classifying gestures. 
The focus is on iconic gestures, in particular on the molding and drawing dimensions, and to some extent also on acting.}
    \label{fig:gesture-taxonomy}
\end{figure}

\subsubsection{Conventionalization}
\label{subsec:conventionalization}

What we have said so far does not mean, however, that the interpretation of gestures is completely arbitrary.
In fact, gestures are classified according to their degree of conventionalization, a classification that has become known as  \emph{Kendon's continuum} \citep[37]{McNeill:1992} -- see \cref{fig:kendon-continuum} for slightly modified diagram.

\begin{figure}[htb]
\centering
\begin{adjustbox}{max width=\linewidth}
\begin{tikzpicture}
    \matrix [ampersand replacement=\&] (m) {
    \node [align=left] {co-speech \\ gestures};
    \& 
    \node [scale=2] {\faCaretRight};
    \& 
    \node [align=left] {recurrrent gestures, \\ gesture families};
    \& 
    \node [scale=2] {\faCaretRight};
    \& 
    \node [align=left] {pantomimes};
    \& 
    \node [scale=2] {\faCaretRight};
    \& 
    \node [align=left] {emblematic \\ gestures};
    \& 
    \node [scale=2] {\faCaretRight};
    \& 
    \node [align=left] {sign \\ languages};
    \\
    };

\begin{scope}[on background layer]
    \fill [MaterialGrey200] (m.north west) -- (m.north east) -- (m.south west) -- cycle;

    \fill [MaterialBrown100] (m.south west) -- (m.south east) -- (m.north east) -- cycle;
\end{scope}
\end{tikzpicture}
\end{adjustbox}
\caption{(Modified) Kendon's Continuum. To the right: increasing degree of conventionalization, decreasing language dependency}
\label{fig:kendon-continuum}
\end{figure}

A key finding is that the space between spontaneous accompanying gestures and fully conventionalized sign languages is populated by recurring gestures, i.e., gestures that share common formal characteristics across different uses  \citep{Bressem:Mueller:2014-away,Fricke:Bressem:Mueller:2014}; one example of this is the \textit{Away} gesture family.
Note that pantomimes occupy a near-to-emblematic, separate position in Kendon's continuum.

Our gesture semantic model introduced in \cref{sec:main-informational-evaluation} provides an explanation for the different degrees of conventionality: 
The stronger the match between a gesture's iconic model and the \cvm of a lexical item, the more conventionalized the gesture; the  less the number of concurrent matching {\cvm}s, the more conventionalized the gesture.
Hence, conventionalization is a probabilistic feature of gesture classification: the more conventionalized, the easier to informationally evaluate a gesture.
We conjecture that this explains observations regarding different informational status of pro-, co-, and post-speech gestures, which (incorrectly, see \cref{sec:introduction}) were ascribed to graded (non-)at-issueness \citep{Esipova:2019,Schlenker:2018}. 
A more realistic picture is outlined in \cref{subsec:issueness}, after the introduction of our proposal.

\subsubsection{Empirical data}
\label{subsec:empirical-data}

Some empirical examples that exemplify important kinds of usages of iconic gestures are collected in the following.
%
%
The gestures given in (\nextx) are drawing gestures that depict shape-related properties of the objects talked about (the \enquote{roof gesture} from (\ref{ex:roof}) is one of them).
A gloss following \enquote*{$\Rightarrow$} provides a brief verbal description of the spatial semantic content of the multimodal constituent.  
Where possible, we link to publicly available video files that show the temporal alignment of speech and gesture and the performance velocity of the gesture.
\pex \label{ex:gesture-examples-shape}
\a Stewart Robson on \textit{ESPN FC Extra Time} (first gesture): 
\par 
\textit{You know when they go on that wheel} \includegraphics[width=2cm]{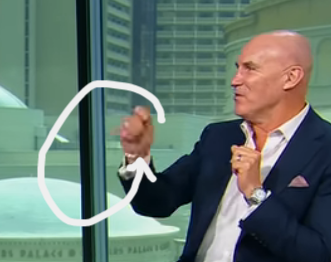} \textit{and throw the dagger would you ever like to see that go wrong}? \\ 
$\Rightarrow$ the wheel has a circular layout
\par 
(\url{https://www.youtube.com/watch?v=CiVS5_HKFY8&t=0h1m18s})
\label{ex:wheel} 


\a SaGA dialogue V11, starting at 2:32:
\par
dann ist das Haus halt so: / \textit{then the house is like this}: 
\includegraphics[trim=50 0 150 180, clip, width=2.6cm]{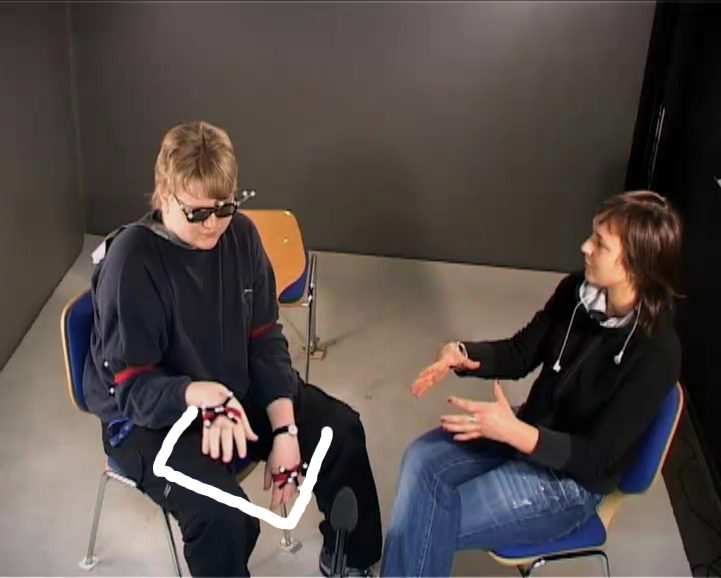} 
\par$\Rightarrow$ the house has a rectangular, U-shaped layout. 
\label{ex:u-shaped}
\xe 
Note that (\lastx b), in contrast to (\lastx a), cannot even be understood without the gesture. 
A gesture semantics must handle both cases.

The gestures in (\nextx) can be conceived as drawing gestures, too.
However, contrary to the object-related gestures collected in (\lastx), these are event gestures that depict movement paths of the event talked about.
\pex \label{ex:gesture-examples-event} 
\a SaGA dialogue V5, starting at 6:48 (an English translation is provided after the slash)
\par
und da musst du sofort, scharfer rechter Winkel, \includegraphics[width=2cm]{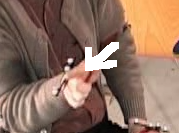} rechts rein / \textit{and you must enter immediately on the right, at an acute right angle} \\
$\Rightarrow$ the path of the movement runs straight to the right\footnote{There is an -- arguably harmless -- ambiguity here, namely whether the gesture depicts the path of the entering action, or the path/road on which the entering takes place. We assume that the gesture is affiliated to the verb phrase, amounting to an event-related rendering.}
\label{ex:rechts-rein}

\a Keanu Reeves at the \textit{Graham Norton Show}:
\par 
\textit{a car pulled out in front of me} \includegraphics[trim=0 10 0 50, clip, width=2cm]{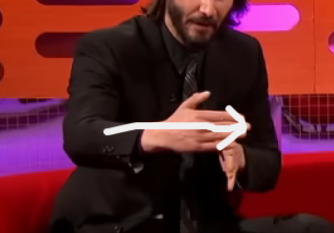}\\
$\Rightarrow$ the car pulled out in a straight line from the right
\par
(\url{https://www.youtube.com/watch?v=6VxnceG_eh4&t=0h13m39s})
\label{ex:pulled-out}
\xe


The gestures in (\nextx) are acting gestures (also called pantomime): the speaker imitates an action by acting as the actor in the action.
\pex \label{ex:gesture-examples-action}
\a Stewart Robson on \textit{ESPN FC Extra Time} (second gesture): 
\par 
\textit{You know when they go on that wheel and throw} \includegraphics[width=2cm]{gesture-throw-dagger} \textit{the dagger would you ever like to see that go wrong}? \\ 
$\Rightarrow$ throwing exhibits a certain handshape and movement trajectory 
\par
(\url{https://www.youtube.com/watch?v=CiVS5_HKFY8&t=0h1m18s})
\label{ex:throw-dagger}

\a Keanu Reeves at the \textit{Graham Norton Show}:
\par 
\includegraphics[trim=0 20 10 20, clip, width=2.5cm]{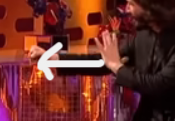} \textit{no one could hit}
\\
$\Rightarrow$ fist and straight movement path away from speaker's body
\par
(\url{https://www.youtube.com/watch?v=6VxnceG_eh4&t=0h1m6s})
\label{ex:hit}
\xe 

As evinced in (\lastx), the acting or pantomimic gesture dimension is not a fully spatial one: it involves an interpretation of handshapes on top of a spatial interpretation of movement paths.
We will account for acting gestures by imposing a quotational condition for handshapes in addition to a spatial iconic model derived from the motion component (\cref{subsec:gesture-vector}).
The quotational requirement seems to be a distinguishing property of gestures that are primarily located within the acting dimension.

\subsubsection{Kinematic gesture representation}
\label{subsec:kinematic-gesture-representation}


%
The alphabet provides a ready-made transcription system for written text, and phonetic transcription systems for spoken language.
%
%
But how to represent iconic gestures? 
%
%
There are several transcription systems for gestures around, we illustrate gesture annotation according to the format used by  \citet{Kopp:Tepper:Cassell:2004,Luecking:Bergmann:Hahn:Kopp:Rieser:2010}.
%
%
In fact, \enquote{annotation playing for gestures the same role as syntax representation plays for linguistic utterances} \citep[p.~123]{Rieser:2015}, \Revision{that is, providing the backbone of their semantic interpretation}.

%
%
The formal description of a gestural movement is given for each hand in terms of its \enquote{anatomical logic}, namely the \emph{handshape}, the orientations of the \emph{palm} and the \emph{back of the hand} (\textsc{boh}), the movement trajectory (if any) of the \emph{wrist}, and the relation between both hands (synchronicity, \textsc{sync}).
The handshape can be transcribed according to a handshape nomenclature -- we use 
the Multimodal Multidimensional (M3D) gesture labeling system \citep{Rohrer:et:al:2023-m3d}.
%
%
The orientations (\textsc{orient}) of the palm and back of the hand are specified with reference to the speaker's body (e.g.,\ \textit{PAB} encodes \enquote{palm away from body} and \textit{BUP} encodes \enquote{back of hand upwards}). 
\Revision{Specifying both palm and back of hand is redundant, but useful for practical purposes such as monitoring consistency.}
Movement trajectories of the whole hand are specified with respect to the wrist in terms of the described path, its direction (\textsc{dir}), and the extent of the movement.
Position and extent are given with reference to the \emph{gesture space} \citep[p.~86--89]{McNeill:1992}, see \cref{fig:gesture-space}.
%
%
%
%
%
%
%
\Revision{Movement annotations can be concatenated by means of the symbol \enquote*{$>$}. 
Concatenated directions provide annotations for gesture movements that involve a change in direction, as, for instance, the shape demonstration in (\ref{ex:u-shaped}).}

\begin{figure}[htb]
    \centering
  \begin{tikzpicture}[scale=0.9, 
  Dist/.style={below, font=\sffamily\footnotesize}]
  \tdplotsetmaincoords{80}{110}
  \begin{scope}[tdplot_main_coords]
    \coordinate (UHL) at (0,0,0); 
    \coordinate (UHR) at (0,5,0); 
    \coordinate (VUR) at (5,5,0); 
    \coordinate (VUL) at (5,0,0); 
    
    \filldraw [draw=black, fill=GU-Sandgrau!80!black] (UHL) -- (UHR) --
    (VUR) -- (VUL) -- cycle; 
    \foreach \x in {1,2,3,4} {
        \draw [dashed, white] (0,\x,0) -- (5,\x,0);
      }
      \foreach \y in {1,...,4} {
        \draw [dashed, white] (\y,5,0) -- (\y,0,0);
      }
      
      \coordinate (OHL) at (0,0,5);
      \coordinate (OVL) at (5,0,5);
      \filldraw [draw=black, fill=GU-Sandgrau] (UHL) -- (OHL) --
      (OVL) -- (VUL) -- cycle; 
      \foreach \z in {1,...,4} {
        \draw [dashed, white] (\z,0,0) -- (\z,0,5);
      }
      \foreach \u in {1,2,3,4} {
        \draw [dashed, white] (0,0,\u) -- (5,0,\u);
      }
      
      \coordinate (OHR) at (0,5,5);
      \coordinate (OVR) at (5,5,5);
      \filldraw [draw=black, fill=GU-Sandgrau!90!black] (UHL) -- (UHR) --
      (OHR) -- (OHL) -- cycle; 
      \foreach \v in {1,2,3,4} {
        \draw [dashed, white] (0,0,\v) -- (0,5,\v);
        \draw [dashed, white] (0,\v,0) -- (0,\v,5);
      }


      \Block{2}{2}{2}{1}{GU-Purple}
      \Block{2}{2}{3}{2}{GU-Purple}
      \Block{2}{2}{4}{3}{GU-Purple}
      \Block{2}{3}{2}{1}{GU-Purple}
      \Block{2}{3}{3}{2}{GU-Purple}
      \Block{2}{3}{4}{3}{GU-Purple}
      \Block{2}{4}{2}{1}{GU-Purple}
      \Block{2}{4}{3}{2}{GU-Purple}
      \Block{2}{4}{4}{3}{GU-Purple}
      \Block{3}{2}{2}{1}{GU-Purple}
      \Block{3}{2}{3}{2}{GU-Purple}
      \Block{3}{2}{4}{3}{GU-Purple}
      \Block{3}{3}{2}{1}{GU-Purple}
      \Block{3}{3}{3}{2}{GU-Purple}
      \Block{3}{3}{4}{3}{GU-Purple}
      \Block{3}{4}{2}{1}{GU-Purple}
      \Block{3}{4}{3}{2}{GU-Purple}
      \Block{3}{4}{4}{3}{GU-Purple}
      \Block[CBL]{4}{2}{2}{1}{GU-Purple}
      \Block[CL]{4}{2}{3}{2}{GU-Purple}
      \Block[CUL]{4}{2}{4}{3}{GU-Purple}
      \Block[CB]{4}{3}{2}{1}{GU-Purple}
      \Block[CC]{4}{3}{3}{2}{GU-Purple}
      \Block[CU]{4}{3}{4}{3}{GU-Purple}
      \Block[CBR]{4}{4}{2}{1}{GU-Purple}
      \Block[CR]{4}{4}{3}{2}{GU-Purple}
      \Block[CUR]{4}{4}{4}{3}{GU-Purple}


      \draw [->, thick] (4,2.5,2.5) -- (6,2.5,2.5) node [below=0.8cm] {back};
      \draw [->, thick] (2.5,4,2.5) -- (2.5,5.3,2.5) node [below] {right};
      \draw [->, thick] (2.5,2.5,4) -- (2.5,2.5,4.5) node [above] {up};
      \draw [<->, thick] (4,4,1) -- (1,4,1) node [Dist, at start] {N} node [Dist, midway] {M} node [Dist, at end] {F};
    \end{scope}

    \matrix [matrix of nodes, ampersand replacement=\&, column 1/.style={nodes={text width=1cm, align=left}}, column 2/.style={nodes={text width=3.5cm, align=left}}, xshift=8cm, yshift=1.6cm] {
      CBL: \& center below left \\
      CL: \& center left\\
      CUL: \& center upper left \\
      CB \& center below \\
      CC: \& center center \\
      \ldots \& \ldots \\
      N: \& near\\
      M: \& middle\\
      F: \& far \\
    };
  \end{tikzpicture}
  \caption{Gesture space model. Three-dimensional extension of the two-dimensional model of \protect{\citet{McNeill:1992}} (taken from \protect\citet{Luecking:2016}).}
\end{figure}

\Revision{As an example, consider the representation of the horizontal gesture from our first example in (\ref{ex:roof}), re-used in (\ref{ex:roof2}), (\ref{ex:roof-dialogue}), (\ref{ex:roof-anaphora}), and (\ref{ex:roof-line}).
The right hand with the index finger extended is moved to the left.
We consider the slight upward movement part of the preparation phase, which brings the gesture to its starting position in gesture space.
}
The corresponding annotation is shown in the attribute--value matrix (\textsc{avm}) in \cref{fig:horizontal-gesture-representation}.
The feature value \textit{none} tells us that the left hand is in rest position (while the right hand is active, \enquote*{RHA}).
Such information is captured as synchronization information (\textsc{sync}), which also hosts the starting and ending slot of a gesture movement in gesture space (\textsc{sloc} respectively \textsc{eloc}).
%
%
%
The change of direction involved in the shape demonstration from example (\ref{ex:u-shaped}) is shown in the annotation in \cref{fig:u-shape-gesture-representation}, which makes use of a concatenation of wrist movement values.

\begin{figure}
\begin{floatrow}
\ffigbox[0.45\textwidth]{%
\footnotesize
\begin{avm}
  \[\asort{right hand}
    handshape & D \\ 
    palm & \[orient & PDN\\
      path & 0 \\
      dir & 0 \] \\
    boh & \[orient & BUP \\
      path & 0  \\
      dir & 0\] \\
    wrist & \[
      path & line \\
      dir & ML \\
      extent & medium\]  \\
    sync & \[config & RHA \\
      rel-mov & none \\
      s-loc & CC-M\\ 
      e-loc & CR-M \]
  \]
\end{avm}
}{%
    \caption{Kinematic representation of the horizontal gesture from example (\ref{ex:roof}).}
    \label{fig:horizontal-gesture-representation}
    }
\ffigbox[0.45\textwidth]{%
\footnotesize
\begin{avm}
  \[\asort{right hand}
    handshape & O \\ 
    palm & \[orient & PDN\\
      path & 0 \\
      dir & 0 \] \\
    boh & \[orient & BUP \\
      path & 0  \\
      dir & 0\] \\
    wrist & \[
      path & line$>$line$>$line \\
      dir & MR$>$MB$>$ML \\
      extent & large\]  \\
    sync & \[config & RHA \\
      rel-mov & none \\
      s-loc & CBR-F\\ 
      e-loc & CBR-N \]
  \]
\end{avm}
}{%
    \caption{Kinematic representation of the shape demonstration from example (\ref{ex:u-shaped}).}
    \label{fig:u-shape-gesture-representation}
    }
\end{floatrow}
\end{figure}

Is a form-based gesture representation system fine-grained enough and captures the continuous nature of gesture movements? 
The problem is that for any two annotation labels, there is a potential third one in between, \textit{ad infinitum}. 
This means that iconic gestures exhibit continuity (they are, as is often said, analog) and are not discrete (they are not digital). 
For these reasons, among others, \citet{Greenberg:2021} suggests treating gestures as concrete paintings and interpreting them in terms of an analogical interpretive function of the gesture form.
However, the interpretation of the gesture form is never explained in detail.
We do not believe that this is merely a minor technicality, but rather an important part of an iconic gesture semantics of substance.
We fill this gap by a direct interpretation of discrete annotation labels (in fact, this idealization is abandoned below in favor of continuous pose estimation representations from computer vision and computational semantics, see \cref{subsec:ai}). 
More importantly, Greenberg's approach, as explained in \cref{subsec:projective-semantics}, does not capture the semiotic nature of (certain types of) iconic gestures.
It therefore remains inherently incomplete anyway.

Even if continuity contains a grain of truth, the iconic reality is cruder than reflections on analog/discrete symbolic representation systems would suggest.
Firstly, the performance of a gesture is often rather vague or fuzzy
\citep{Lawler:Hahn:Rieser:2017}: (omitted) movement aspects do not have to be meaningful at all. 
For example, \citet{Rieser:2011} discusses gestural depictions of a pointed Gothic church window, whereby the gesture trajectory is quite rounded.
A literal, continuous interpretation of gestures will certainly not be able to capture the looseness of gesture production.

Secondly, visual perception also recognizes a Just Noticeable Difference (JND) \citep[even if there are individual differences, cf.][]{Mollon:et:al:2017}. 
This means that not every kinematic difference is perceived, or if it is perceived, does not necessarily lead to a categorical difference -- see (\nextx) for a related example. 

The last point relates, thirdly, to categorical perception: in contrast to continuous perception, categorical perception is much more general and is very likely related to how the neural networks in our brain process perceptual stimuli  \citep{Harnad:2003-perception}.\footnote{A concrete example: a common type of neuron is the so-called leaky integrate-and-fire neuron (LIF). While a LIF continuously integrates its input, its output is sensitive to a threshold value. That is, the firing behaviour of a LIF is nonlinear \citep{Tal:Schwartz:1997}.}
This means that even continuous signals are assigned to a specific category (possibly with an interindividual range of indifference; for example, think of the JND in acoustic phonetics).

Fourthly, a detailed and potentially continuous interpretation of gesture movements is probably only significant in certain contrasting contexts in which the formal difference between two or more gestures is important. 
A simple example of this can be found in (\nextx) using the clamp gesture (thumb and index finger are extended; represented by open rectangles), which indicates distance or size.  
\ex 
not that \tikz[baseline] \draw [black, thick] (0,0.3) -- (0,0) -- (1,0) -- (1,0.3); wide, but this  \tikz[baseline] \draw [black, thick] (0,0.3) -- (0,0) -- (1.1,0) -- (1.1,0.3); wide 
\xe 

The speaker of (\lastx) corrects the size of an entity from the first to the second gesture; the gesture indicates the respective size. 
Note that the right distance is actually greater than the left distance; this could go unnoticed, underscoring categorical perception and JNDs.
%

In the following, it is shown how kinematic gesture representations can be mapped onto vector sequences and interpreted within vector space models. 
%

\subsection{Vector space semantics}
\label{sec:spatial-theories}

To capture the visual content of an iconic gesture, we adopt a vector space semantics based on the work by \citet{Zwarts:1997} (\cref{subsec:vector-space-semantics}). 
%
In any case, some kind of \enquote{preprocessing} is required,  based on the kinematic gesture representation from  \cref{subsec:kinematic-gesture-representation}, which produces the visuo-spatial content of gestures -- \emph{iconic models} (\cref{subsec:gesture-vector}). 
%
The basic compositional interaction of iconic models with speech is discussed in \cref{subsec:compositional-speech-gesture-integration}.
The semantic framework used to formulate the discussion is a standard version of situation semantics in line with \citet{Kratzer:1989}; we return to theoretical repercussions in \cref{subsec:repercussions}.

\subsubsection{Basic vector space semantics}
\label{subsec:vector-space-semantics}

%
%
%
%

Following \citet{Zwarts:Winter:2000}, we assume that vectors \Revision{and vector sequences} (concatenations, or \enquote{paths}, of vectors) are primitive spatial entities in natural language models. 
Vectors are given within a vector space $V$ over the real numbers $\mathbb{R}$.\footnote{We assume a standard Euclidean space, but this is not a necessity: spaces defined in terms of polar instead of Cartesian coordinates are formally as well as conceptually potent alternatives \citep{Zwarts:Gaerdenfors:2016}.}
Vector spaces are closed under vector addition and scalar multiplication. 
Two domains are defined from this ontology: 
\begin{itemize}
\item The domain of points: $D_p = V$ (each point is defined by a vector's endpoint)
\item The domain of vectors: $D_v = V \times V$ (the Cartesian product of $V$)
\end{itemize}

Vector space population: For each point $p \in D_p$ there is a vector space $V_p \subseteq D_v$ ($p$ is the zero-vector, or \Revision{origin} of $V_p$). 

%

$D_e$ (entities), $D_s$ (events) and $D_v$ (vector sequences) are related by a couple of functions, some of which are briefly introduced in the following (we basically follow the vector space model introduced by \citealt{Zwarts:1997,Zwarts:2003}): 
\begin{itemize}
\item The vector space located at a concrete object in $e$ denoted by an NP $\alpha$ is given by \enquote*{$\vecspace(\mng[e]{\alpha})$}.


\item Place and axis vectors determine spatial relationships: 
\begin{itemize}
\item $\place(x,\bvec{v})$: $x$ is placed at the end of \bvec{v}; $\place(\bvec{v},x)$: the starting point of \bvec{v} is placed at $x$; $\place(\bvec{u},\bvec{v})$: the starting point of \bvec{u} is placed at the end of \bvec{v} -- see \cref{fig:place-vector}.
\item $\axis(x,\bvec{v})$ object $x$ has an axis \bvec{v} -- see \cref{fig:axis-vector}.
\end{itemize}

\item Paths: sequences of axis or place vectors,  see \cref{fig:place-path}. Paths are defined as $n$-tuples of points, $n > 2$.
Paths are notated as $\bvec{v}[k]$ (index $k$ can be thought of as an indication of the length of the tuple). We use the letters $a$ and $z$ to denote the first and the last element of the tuple, respectively. 
The start of a path is indexed as $\bvec{v}[a]$, its end point as $\bvec{v}[z]$.\footnote{Hence, $\place(\bvec{w},\bvec{v}) \Leftrightarrow \bvec{w}[a] = \bvec{v}[z]$.}
%
\Revision{Paths can be constructed for axis and for place vectors, giving rise to axis-paths and place-paths, respectively.
\Cref{fig:place-path} shows a place-path, that is, object $x$'s movement along a place trajectory.
Axis-paths are needed to express \textit{eigenmovement} of objects, for instance, rotation.
Paths are elements of $D_v$.}
Note that paths are non-temporal entities. 
They receive a temporal interpretation only if index $k$ is mapped to points or intervals in time.
\end{itemize}

\begin{figure}[htb]
\begin{floatrow}
\ffigbox[\FBwidth]{%
    \begin{tikzpicture}
    \draw [gray] (-1.2,-1) rectangle (2.75,3);
      \draw[vector] (0,0.1) node [below left, rectangle, draw, align=center, label=below:{$\place(\bvec{v},y)$}] {$y$} -- (1,2) node [midway, auto] {$\bvec{v}$} node [above] {$\place(\bvec{w},\bvec{v})$};
      \draw [vector] (1,2) -- (1.5,0.5) node [midway,auto]
      {$\bvec{w}$} node [below right, rectangle, draw, label=below:{$\place(x,\bvec{w})$}] {$x$};
    \end{tikzpicture}
}{%
\caption{Place vector}
\label{fig:place-vector}
}    
\ffigbox[1.1\FBwidth]{%
    \begin{tikzpicture}
    \draw [gray] (-0.8,-1) rectangle (1.8,3);
      \draw (0,-0.2) rectangle (1,2.2) node [above] {$x$};
      \draw[vector] (0.5,-0.2) -- (0.5,2.2) node [midway,auto]
      {$\mathbf{v}$};
      \draw [white] (-0.7,0) -- (1.7,0);
    \end{tikzpicture}
}{%
\caption{Axis vector}
\label{fig:axis-vector}
}
\ffigbox[\FBwidth]{%
   \begin{tikzpicture}
   \draw [gray] (-0.5,-0.5) rectangle (2.6,3.5);
      \draw[dotted] (0,3) .. controls (0,3) and (1.5,1.5) .. (2,0);
      \path (2,0) node [below,rectangle,draw] {$x$};
      \draw[vector] (2,0) -- (0,3);
      \draw[vector] (2,0) -- (1,1.8);
      \draw[vector] (2,0) -- (1.5,1.1);
    \end{tikzpicture}
}{%
\caption{Place-path}
\label{fig:place-path}
}
\end{floatrow}
\end{figure}

\begin{itemize}
\item Vectors pointing in the same direction constitute a \Revision{\emph{level}}.
The set of vectors varying only in the direction of one of their three dimensions make up a \emph{plane}.
 

\item The \emph{orthogonal complement} $A^\perp$ of a level or plane $A$ is the set of vectors orthogonal to those in $A$ ($A^\perp = \{\bvec{v} \in D_v \mid \forall\bvec{w} \in A: \bvec{v} \perp \bvec{w}\}$), see \cref{fig:orthogonal}. 

\item Each vector space $V$ provides three, mutually perpendicular \emph{orienting levels}.  Intuitively, these levels correspond to the directions \textit{up} (\textsc{up}), \textit{forward} or \textit{front} (\textsc{ft}), and \textit{right} (\textsc{rt}). These levels in addition to their corresponding inverses, give rise to an \emph{oriented vector space}, see \cref{fig:oriented-axes}. The orienting levels are determined by external reference frames, or by intrinsic or functional properties of reference objects. 

\item Given orienting levels $A$ and $A'$, a vector \bvec{v} can be decomposed into its \emph{projections} onto the levels, $\bvec{v}_A$ and $\bvec{v}_A'$. \cref{fig:orthogonal-components} shows the orthogonal components of \bvec{v} on the \textsc{up} and the \textsc{rt} levels.
\end{itemize}

\begin{figure}[htb]
\begin{floatrow}
\ffigbox[\FBwidth]{%
\begin{tikzpicture}
\draw [gray] (-0.65,-1.5) rectangle (1.5,1.5);
    \draw [axis] (0,-1) -- (0,1) node [midway, left] {A};
    \foreach \comp in {-0.9,-0.8,...,0.9} {\draw [vector, black!70] (0,\comp) -- (1,\comp);}
\end{tikzpicture}
}{%
\caption{Orthogonal complements of an axis}
\label{fig:orthogonal}
}
\ffigbox[\FBwidth]{%
    \begin{tikzpicture}
    \draw [gray] (-2,-1.5) rectangle (1.8,1.5);
      \fill (0,0,0) circle (2pt);
      \draw[axis] (0,0,0) -- (0,1,0) node [above]
      {\textsc{up}};
      \draw[axis] (0,0,0) -- (0,-1,0) node [below]
      {$-$\textsc{up}}; 
      \draw[axis] (0,0,0) -- (1,0,0) node [right]
      {\textsc{rt}};
      \draw[axis] (0,0,0) -- (-1,0,0) node [left]
      {$-$\textsc{rt}};
      \draw[axis] (0,0,0) -- (0,0,-1) node [above right]
      {\textsc{ft}};
      \draw[axis] (0,0,0) -- (0,0,1) node [below left]
      {$-$\textsc{ft}};
    \end{tikzpicture}
}{%
\caption{Orienting half-axes and their inverses}
\label{fig:oriented-axes}
}
\ffigbox[\FBwidth]{%
\begin{tikzpicture}
\draw [gray] (-1.3,-1.5) rectangle (1.4,1.5);
    \fill (-0.5,-1) circle (2pt);
    \draw [axis] (-0.5,-1.2) -- (-0.5,1.2) node [left] {\textsc{up}};
    \draw [axis] (-0.6,-1) -- (0.7,-1)  node [right] {\textsc{rt}};
    \draw [vector] (-0.5,-1) -- (0.3,0.8) node [above right] {\bvec{v}};
    \draw [vector] (-0.5,-1) -- (0.3,-1) node [below] {$\bvec{v}_\textsc{rt}$};
    \draw [vector] (-0.5,-1) -- (-0.5,0.8) node [left] {$\bvec{v}_\textsc{up}$};
    \draw [dashed] (-0.5,0.8) -- (0.3,0.8) -- (0.3,-1);
\end{tikzpicture}
}{%
\caption{Projections of a vector}
\label{fig:orthogonal-components}
}
\end{floatrow}
\end{figure}

\begin{RevisionEnv}
How do spatial domains look like?
Three examples are given in \crefrange{fig:space-of-glass}{fig:throw}. 
%
\Cref{fig:space-of-glass} shows the vector space associated with an object, namely an upright wine glass. 
%
%
%
The spatial relationship between two objects is shown in \cref{fig:near}.
The ambulance is located at a place vector emanating from the hospital.
%
%
\Cref{fig:throw} illustrates place paths.
The movement involved in a throwing event is a tuple of $k$ points giving rise to a vector sequence $\bvec{v}[k]$.
There is also a second trajectory, namely a place path within the vector space of the thrown object. 
\end{RevisionEnv}

\begin{figure}[htb]
\floatsetup{valign=c, heightadjust=all}
    \begin{subfloatrow}
    \ffigbox[0.4\textwidth]{%
    \begin{tikzpicture}[3d view]
      \draw[axis, gray] (0,0,0) -- (0,2,0); 
      \draw[axis, gray] (0,0,0) -- (0,-2,0); 
      \draw[axis, gray] (0,0,0) -- (1.2,0,0); 
      \draw[axis, gray] (0,0,0) -- (-1.2,0,0); 
      \draw[axis, gray] (0,0,0) -- (0,0,-2); 
      \draw[axis, gray] (0,0,0) -- (0,0,2); 
      \node [rectangle, inner sep=0.2pt, draw, scale=6, thick, text=gray] (bb) at (0,0,0) {\faWineGlass};
      \draw [white] (bb.north) -- (bb.south) node [near start, sloped, anchor=north] {axis};
    \end{tikzpicture}
    }{%
    \caption{Spatial projection of a wine glass. The glass is located at the center of a vector space, 
    and has a main axis.}
    \label{fig:space-of-glass}
    }
    \ffigbox[0.5\textwidth]{%
    \begin{tikzpicture}[3d view]
          \draw[axis, gray] (0,0,0) -- (0,2,0); 
      \draw[axis, gray] (0,0,0) -- (0,-2,0); 
      \draw[axis, gray] (0,0,0) -- (1.5,0,0); 
      \draw[axis, gray] (0,0,0) -- (-1.5,0,0); 
      \draw[axis, gray] (0,0,0) -- (0,0,-2); 
      \draw[axis, gray] (0,0,0) -- (0,0,2); 
    \node [scale=6, inner sep=0pt] (hospital) at (0,0,0) {\faHospital};
    \node [scale=2, inner sep=0pt] (ambulance) at (3,-1,-1) {\faAmbulance};
        \draw [vector] (hospital) -- (ambulance) node [below, midway, sloped] {$\place(a,\bvec{u})$};
    \end{tikzpicture}
    }{%
    \caption{Spatial projection of two objects being spatially related (bounding boxes and axes not shown). The ambulance $a$ is near the hospital if the length of \bvec{u} is below some threshold $\tau$.}
    \label{fig:near}
    }
    \end{subfloatrow}
    \begin{subfloatrow}
    \ffigbox[\textwidth]{%
    \begin{tikzpicture}[3d view]
              \draw[axis, gray] (0,0,0) -- (0,2,0); 
      \draw[axis, gray] (0,0,0) -- (0,-2,0); 
      \draw[axis, gray] (0,0,0) -- (1.5,0,0); 
      \draw[axis, gray] (0,0,0) -- (-2,0,0); 
      \draw[axis, gray] (0,0,0) -- (0,0,-2); 
      \draw[axis, gray] (0,0,0) -- (0,0,2); 
        \node [inner sep=0pt, rotate=100, scale=3] (hand) {\faHandHolding};
        \node [left=0.6cm of hand.north, scale=1.5, inner sep=0pt] (ball)  {\faBaseballBall};
        \draw [inversevector] (hand.south) -- +(-20:0.75cm) node [midway, above, sloped] {$\bvec{v}[k]$};
    \end{tikzpicture}
    }{%
    \caption{Spatial projection of a throwing event. The movement of the hand \enquote{leaves} a place path vector sequence in the domain.}
    \label{fig:throw}
    }
    \end{subfloatrow}
\end{figure}

Why do we need vector denotations \Revision{at all}?
Vectors provide a mathematical model for \Revision{the interpretation of} spatial prepositions such as \textit{above} \citep{Zwarts:1997}.
The differently shaded areas in \cref{fig:above} show three readings of \textit{above} (of a glass), which can all be defined in terms of sets of vectors from the vector space located at the object in question \Revision{that have a non-zero orthogonal projection to the \bvec{UP} level}.\footnote{A more elegant solution is to define a probability distribution over the space surrounding the object denoted by the NP \citep{O_Keefe:1996}.}

\begin{figure}[htb]
\begin{floatrow}[2]
\ffigbox[\FBwidth+2cm]{%
        \begin{adjustbox}{scale=0.7}
        \begin{tikzpicture}
            \node [scale=2] (glass) at (2,2) {\faWineGlass*};
            \fill [gray!30] (0,2.4) rectangle (4,5);
            \fill [gray!50] (0,5) -- (1.8,2.4) -- (2.2,2.4) -- (4,5) -- cycle;
            \fill [gray!70] (1.8,2.4) rectangle (2.2,5);
        \draw (0,1) rectangle (4,5);
        \end{tikzpicture}
        \end{adjustbox}
        }%
        {%
        \caption{\textit{above}(NP) $=$ \\ 
        $\{\bvec{v} \in \text{space}(\mng[e]{\text{NP}}) \mid \card{\bvec{v}_\bvec{UP}} = \card{\bvec{v}} \vee \card{\bvec{v}_\bvec{UP}} > \card{\bvec{v}_{\bvec{RT}/\bvec{-RT}}} \vee \card{\bvec{v}_\bvec{UP}} > 0\}$.
        }
 \label{fig:above}
 }
\ffigbox[\FBwidth+2cm]{%
\begin{adjustbox}{scale=0.7}
        \begin{tikzpicture}
        \draw (0,0) rectangle (4,4);
            \node [scale=3] (walker) at (1,1) {\faWalking};
            \draw [thick] (1.5,0.5) .. controls (2,0) and (2.5,0.5) .. (3.5,1) .. controls (4,1.5) and (3,2) .. (2,2) .. controls (1.5,2.5) and (2,3) .. (3,3);
            \draw [thick, dotted] (3,3) -- (3.5,3);
        \end{tikzpicture}
        \end{adjustbox}
        }%
        {%
        \caption{$\lambda x . \lambda e [\text{walk}(e) \wedge \text{ag}(e) = x \wedge \exists \bvec{v} [\text{path}(e,\bvec{v})]]$.}
        \label{fig:walk-semantics}
        }
    \end{floatrow}
\end{figure}


Vectors (that is, spatial entities) are also part of the truth conditions of predicates that denote  movements in space.
For instance, a walking event is a walking event only if there is a path of movement, be it undirected, as in \cref{fig:walk-semantics}, or directed \citep{Krifka:1998-telicity}:
%
\pex 
\a Mary walked from the university to the capitol.
\a \Revision{$\exists e [\text{walk}(e) \wedge \text{ag}(e) = \text{mary} \wedge \text{source}(e) = u \wedge \text{goal}(e) = c \wedge \exists\bvec{v}[\place(\bvec{v},u) \wedge \place(c,\bvec{v})]]$}
\xe 

(\lastx b) is the existential closure of the lambda abstraction over events.
It consists of conjoined linguistic (related to event $e$) and visual (related to \bvec{v}) semantic contributions. 
%
The sentence in (\lastx a) is true if the walking event in question took place, in which case there is a path $\bvec{v}$. 
The path vector sequence adds information about the way in which the destination is reached: the path distinguishes, for example, between going directly to the goal and taking detours.
Accordingly, \enquote{Mary walked the shortest route from the university to the capitol.} is true if there is an $e$ and an $x$ such that (\lastx b) holds, and there is no \bvec{w} which is shorter than \bvec{v}.




\subsubsection{Iconic models in vectorial gesture space}
\label{subsec:gesture-vector}


\Revision{A main obstacle for any theory of iconic gesture semantics is to provide an interpretation of the form of a gesture.
In this section, we provide a mapping from kinematic gesture representations (\cref{subsec:kinematic-gesture-representation}) to iconic models, that is, to vectorial representations in vector space (\cref{subsec:vector-space-semantics}). 
The iconic models are then evaluated in a spatial domain (\cref{subsec:compositional-speech-gesture-integration}), providing constraints for the vector spaces associated with a gesture's affiliate expressions in speech. 
For this purpose, we take our computational hint from \citet{Luecking:2016}.}
The basic idea is twofold but simple:
\begin{itemize}
    \item The notion of oriented vector space is used as a formal model of a speaker's gesture space.
    \item The annotation predicates that represent gesture kinematics (\cref{subsec:kinematic-gesture-representation}) are re-interpreted in terms of vector \Revision{descriptions} in vectorial gesture space.
\end{itemize}
This two-step process gives rise to a computational procedure to derive vector-based iconic models from the form of a gesture -- the first characterizing feature of iconicity (the second one is the intersection of these iconic models with the models of the spatial configurations of the objects and events talked about, cf. \cref{fig:architecture1}).

A gesture space \enquote*{$\gestspace(s)$} is anchored at each speaker $s$, see \cref{fig:gesture-space}.
%
%
\Revision{The gesture space is aligned with the orienting levels of the speaker's vector space, that is, it inherits the orientation along} the anatomical planes which hypothetically transect the speaker's body, see \cref{fig:planes}.
%
\pex Anatomical planes
\a \emph{Sagittal plane}  $\coloneqq \{\bvec{v} \in
D_v \mid \bvec{v} \perp \textsc{up} \cup \textsc{rt} \vee \bvec{v} \perp -\textsc{up} \cup -\textsc{rt}\}$

\a \emph{Transverse plane}  $\coloneqq \{\bvec{v} \in D_v \mid  \bvec{v} \perp \textsc{ft} \cup \textsc{up} \vee \bvec{v} \perp -\textsc{ft} \cup -\textsc{up}\}$ 

\a \emph{Coronal} (or \emph{Frontal}) \emph{plane}  $\coloneqq \{\bvec{v} \in D_v \mid  \mathbf{v} \perp \textsc{rt} \cup \textsc{ft} \vee \bvec{v} \perp -\textsc{rt} \cup -\textsc{ft}\}$
\xe 


\begin{figure}
\centering
\begin{tikzpicture}[3d view]
\node [circle, inner sep=0pt, minimum size=6pt, gray, shading=ball] at (0,0,0) {};1111
\draw [gray] (-1,1,-1) -- (1,1,-1) -- (1,1,1) -- (-1,1,1) -- cycle;
\draw [gray] (-1,0,-1) -- (1,0,-1) -- (1,0,1) -- (-1,0,1) -- cycle;
\draw [gray] (-1,-1,-1) -- (1,-1,-1) -- (1,-1,1) -- (-1,-1,1) -- cycle;
\draw [gray] (1,-1,-1) -- (1,1,-1);
\draw [gray] (1,-1,0) -- (1,1,0);
\draw [gray] (1,-1,1) -- (1,1,1) node [above, text=black] {$\gestspace(s)$};
\draw [gray] (0,-1,-1) -- (0,1,-1);
\draw [gray] (0,-1,0) -- (0,1,0);
\draw [gray] (0,-1,1) -- (0,1,1);
\draw [gray] (-1,-1,-1) -- (-1,1,-1);
\draw [gray] (-1,-1,0) -- (-1,1,0);
\draw [gray] (-1,-1,1) -- (-1,1,1);
\draw [gray] (-1,1,0) -- (1,1,0);
\draw [gray] (-1,0,0) -- (1,0,0);
\draw [gray] (-1,-1,0) -- (1,-1,0);
\draw [gray] (0,1,-1) -- (0,1,1);
\draw [gray] (0,-1,-1) -- (0,-1,1);

\coordinate (s) at (-1.5,0,0);
\begin{scope}[vector, black]
\draw (s) -- (-1.5,0,2) node [at end, right] {\bvec{UP}};
\draw (s) -- (-1.5,0,-2) node [at end, left] {\bvec{-UP}};
\draw (s) -- (-1.5,2,0) node [at end, left] {\bvec{-RT}};
\draw (s) -- (-1.5,-4.5,0) node [at end, below] {\bvec{RT}};
\draw (s) -- (-3.5,0,0) node [at end, left] {\bvec{-FT}} node [near end, below=0.6cm, align=center] {$\vecspace(s)$};
\draw (s) -- (2.5,0,0) node [at end, right] {\bvec{FT}};
\end{scope}

\node [scale=8, text=gray!50] (speaker) at (-1.5,0,0) {\faFemale};
\node [rectangle, rounded corners, fill=gray!50, text width=1cm, minimum height=0.2cm, anchor=west, rotate=3] at (-1.5,0,0) {}; 
\node [circle, inner sep=0pt, minimum size=6pt, shading=ball] at (-1.5,0,0) {};
\draw [vector, thick] (-1.5,0,0) -- (0,0,0) node [midway, below, sloped, inner sep=1pt] {$\place(\bvec{v},s)$};
\end{tikzpicture}
\caption{Gesture space: the orienting half axes of the speaker's vector space \enquote*{$\vecspace(s)$} align with the place vector of the speaker's gesture space \enquote*{$\gestspace(s)$} and define the indexical frame of reference for relative locations and directions (\textit{right}, \textit{left}, \textit{in front of}, \textit{below}, \ldots).}
\label{fig:gesture-space}
\end{figure}

\begin{figure}
\subfloat[\label{fig:sagittal-plane}][Sagittal plane]{%
\begin{adjustbox}{max width=0.3\textwidth}
    \begin{tikzpicture}[every node/.append style={font=\footnotesize}]
    \draw [gray] (-2,-1.5) rectangle (1.8,1.5);
      \fill (0,0,0) circle (2pt);
    \draw[axis] (0,0,0) -- (0,1,0) node [above]
      {\bvec{UP}};
      \draw[axis] (0,0,0) -- (0,-1,0) node [below]
      {\bvec{-UP}}; 
      \draw[axis] (0,0,0) -- (1,0,0) node [right]
      {\bvec{RT}};
      \draw[axis] (0,0,0) -- (-1,0,0) node [left]
      {\bvec{-RT}};
      \draw[axis] (0,0,0) -- (0,0,-1) node [above right]
      {\bvec{FT}};
      \draw[axis] (0,0,0) -- (0,0,1) node [below left]
      {\bvec{-FT}};
      \filldraw [fill=gray, draw=black, opacity=0.5] (0,-1,-1) -- (0,1,-1) -- (0,1,1) -- (0,-1,1) -- cycle;
    \end{tikzpicture}
\end{adjustbox}
}\hfill
\subfloat[\label{fig:transverse-plane}][Transverse plane]{%
\begin{adjustbox}{max width=0.3\textwidth}
\begin{tikzpicture}[every node/.append style={font=\footnotesize}]
    \draw [gray] (-2,-1.5) rectangle (1.8,1.5);
      \fill (0,0,0) circle (2pt);
      \draw[axis] (0,0,0) -- (0,1,0) node [above]
      {\bvec{UP}};
      \draw[axis] (0,0,0) -- (0,-1,0) node [below]
      {\bvec{-UP}}; 
      \draw[axis] (0,0,0) -- (1,0,0) node [right]
      {\bvec{RT}};
      \draw[axis] (0,0,0) -- (-1,0,0) node [left]
      {\bvec{-RT}};
      \draw[axis] (0,0,0) -- (0,0,-1) node [above right]
      {\bvec{FT}};
      \draw[axis] (0,0,0) -- (0,0,1) node [below left]
      {\bvec{-FT}};
      \filldraw [fill=gray, draw=black, opacity=0.5] (1,0,1) -- (-1,0,1) -- (-1,0,-1) -- (1,0,-1) -- cycle;
    \end{tikzpicture}
\end{adjustbox}
}\hfill
\subfloat[\label{fig:coronal-plane}][Coronal plane]{%
\begin{adjustbox}{max width=0.3\textwidth}
\begin{tikzpicture}[every node/.append style={font=\footnotesize}]
    \draw [gray] (-2,-1.5) rectangle (1.8,1.5);
      \fill (0,0,0) circle (2pt);
      \draw[axis] (0,0,0) -- (0,1,0) node [above]
      {\bvec{UP}};
      \draw[axis] (0,0,0) -- (0,-1,0) node [below]
      {\bvec{-UP}}; 
      \draw[axis] (0,0,0) -- (1,0,0) node [right]
      {\bvec{RT}};
      \draw[axis] (0,0,0) -- (-1,0,0) node [left]
      {\bvec{RT}};
      \draw[axis] (0,0,0) -- (0,0,-1) node [above right]
      {\bvec{FT}};
      \draw[axis] (0,0,0) -- (0,0,1) node [below left]
      {\bvec{-FT}};
\filldraw [fill=gray, draw=black, opacity=0.5] (1,1,0) -- (1,-1,0) -- (-1,-1,0) -- (-1,1,0) -- cycle;
    \end{tikzpicture}
\end{adjustbox}
}
\caption{Oriented planes in vector space.}
\label{fig:planes}
\end{figure}

With the formal notion of gesture space as vector space, a mapping from kinematic gesture representation of simple, uni-directional movements to vector \Revision{descriptions} is straightforward: the directions encoded in gesture annotation predicates are translated to orienting levels. 
%

%
Moving on to vector sequences, vectors are concatenated head-to-tail, while the concatenation obeys the trajectory annotation \emph{line} vs. \emph{arc}, which distinguishes roundish from angular paths. 
%
A minimal example is shown in (\nextx), where the iconic models emerging from vector sequence \Revision{description} $\bvec{UP}>_{\text{line}}\bvec{RT}$ respectively $\bvec{UP}>_{\text{arc}}\bvec{RT}$ are given.
\ex 
\begin{tikzpicture}[baseline=(anchor)]
    \draw [thick] (0,0) -- (0,1) node [midway, left] {\bvec{UP}};
    \draw [vector] (0,1) -- (2,1) node [at start, above left] {$>$} node [at start, below right] {line} node [midway, above] (anchor) {\bvec{RT}};

    \begin{scope}[xshift=4cm]
        \draw [vector] (0,0) arc[
            start angle=180,
            end angle=90,
            x radius=2,
            y radius=1
            ] node [near start, left] {\bvec{UP}} node [midway, above left] {$>$} node [midway, below right] {arc} node [near end, above] {\bvec{RT}};
    \end{scope}
\end{tikzpicture}
\xe 

The resulting vector sequence \Revision{descriptions}, $\bvec{z}=\bvec{UP}>_{\text{line}}\bvec{RT}$ and $\bvec{z'}=\bvec{UP}>_{\text{arc}}\bvec{RT}$, \Revision{both describe open paths}, that is $\bvec{z}[a]\neq\bvec{z}[z]$ and $\bvec{z'}[a]\neq\bvec{z'}[z]$.
Distinguishing open and closed trajectories is important and is brought about by comparing the starting and ending positions of vector trajectories in vectorial gesture space: both are part of kinematic gesture representations.
We notate $>_{\text{line}}$ as $\perp$ and $>_{\text{arc}}$ as $\circ$.
The algebraic derivation of vectorial representations from the kinematic annotation of a gesture \enquote*{$\gestvec(\gamma)$} is given in (\nextx).
Following formal gesture representations (\cref{subsec:kinematic-gesture-representation}), input and output are attribute--value matrices (\textsc{avm}s).
%
%
\pex Gesture vectorization function $\gestvec(\gamma)$ as model for $\phi$. \label{ex:gesture-vectorization} 
\a $\gestvec(\text{handshape}\ \cornerquote{$\alpha$}) =$ \begin{avm}\[hs & \cornerquote{$\alpha$}\]\end{avm}
\a $\gestvec(\bvec{u}>_{\text{line}}\bvec{v}) =$ \begin{avm}\[traj & $\bvec{u}\perp\bvec{v}$\]\end{avm}
\a $\gestvec(\bvec{u}>_{\text{arc}}\bvec{v}) =$ \begin{avm}\[traj & $\bvec{u}\circ\bvec{v}$\]\end{avm}
\a $\gestvec(\text{s-loc,e-loc}) = \begin{cases}
    \text{sync  traj}[a]=\text{traj}[z] & \text{if s-loc} = \text{e-loc} \\ 
    \text{sync  traj}[a]\neq\text{traj}[z] & \text{else}
\end{cases}$ 
\xe 

%
\Revision{Handshape annotation is just copied (\lastx a).}
Vectorization applies progressively over movement annotations (\lastx b,c).
Condition (\lastx d) is the closure condition, which checks whether a given movement trajectory brings about a closed or an open path. 
%
%
\Revision{The structure of a basic iconic model therefore is an \textsc{avm} with values for the three features handshape (\enquote*{hs}), gesture sequence or trajectory (\enquote*{traj}), and syncronization (\enquote*{sync}).}

Let us briefly exemplify the 
\begin{RevisionEnv}
vectorial interpretation of some of the example gestures from \cref{subsec:empirical-data} by means of (\lastx).
The relevant annotation of the \enquote{roof gesture}, \includegraphics[width=1.5cm]{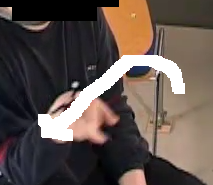}, from (\ref{ex:roof}) is shown in (\nextx), abbreviating the full representation from \cref{fig:horizontal-gesture-representation}.
The output of the vectorization function is the description of a straight vector on the \bvec{RT} plane (the M3D label for the stretched index finger is \enquote{D}).
\ex 
\footnotesize
\begin{avm}
$\gestvec$(%
  \[\asort{right hand}
    handshape & D \\
    wrist & \[
      path & line \\
      dir & MR \\
      extent & small\]  \\
    sync & \[config & RHA \\
      rel-mov & none \\
      s-loc & CC-M\\ 
      e-loc & CR-M \]
  \])
$=$
\[
hs & D\\
traj & $\bvec{RT}$ \\ 
sync & traj[a] $\neq$ traj[z]
\]
\end{avm}
\label{ex:roof-vector}
\xe

The wheel gesture from (\ref{ex:wheel}), \includegraphics[width=1.5cm]{gesture-wheel.png}, describes a full circle on the Sagittal plane. 
The path is a closed one, as indicated by sync's feature value.
%
%
%
\ex 
\footnotesize
\begin{avm}
$\gestvec$(%
  \[\asort{right hand}
    handshape & D \\
    wrist & \[
      path & arc$>$arc$>$arc$>$arc \\
      dir & MU$>$MF$>$MD$>$MB \\
      extent & medium\]  \\
    sync & \[config & BHA \\
      rel-mov & none \\
      s-loc & CC-M\\ 
      e-loc & CC-M \]
  \])
$=$
\[
hs & D\\
traj & $\bvec{UP}\circ\bvec{FT}\circ\bvec{-UP}\circ\bvec{-FT}$ \\ 
sync & traj[a] $=$ traj[z]
\]
\end{avm}
\label{ex:wheel-vector}
\xe

The demonstration of the layout of the house from (\ref{ex:u-shaped}), \includegraphics[trim=50 0 150 180, clip, width=1.5cm]{mmsubcat2.png}, amounts to a U-shaped trajectory description that is drawn with an open hand (M3D label \enquote{O}).
\ex 
\footnotesize
\begin{avm}
$\gestvec$(%
  \[\asort{right hand}
    handshape & O \\
    wrist & \[
      path & line$>$line$>$line \\
      dir & MR$>$MB$>$ML \\
      extent & large\]  \\
    sync & \[config & RHA \\
      rel-mov & none \\
      s-loc & CB-F\\ 
      e-loc & CB-N \]
  \])
$=$
\[
hs & C \\
traj & $\bvec{RT}\perp\bvec{-FT}\perp\bvec{-RT}$ \\ 
sync & traj[a] $\neq$ traj[z]
\]
\end{avm}
\label{ex:u-vector}
\xe

In sum, iconic models -- vector sequences with handshapes -- can be derived from gesture annotations by means of vectorization function \enquote*{$\gestvec$}.
Iconic models are the semantic contributions of gestures and impose spatial constraints on the evaluation of multimodal utterance, to which we turn shortly in \cref{subsec:compositional-speech-gesture-integration} (cf. also the architecture of spatial semantics given in \cref{fig:architecture1}). 
In most cases, however, the iconic models do not apply verbatim, that is, in exactly the orientation and size as they are represented by the gesture in gesture space.\footnote{This raises the question of how precise a gesture can actually be (see also the discussion in \cref{subsec:kinematic-gesture-representation}). Given that humans developed a lot of precision instruments like rulers and compasses, the bare hands are probably good at sketching spatial gists -- which is arguably sufficient for everyday communication, but falls short of drawing neat geometrical structures and indicating precise lengths.}
%
%
For instance, the layout gesture in (\ref{ex:u-vector}) describes a vector in gesture space that has a physical length of about 40\,cm, which is certainly much shorter than the actual layout of the house under discussion. 
Accordingly, we add some spatial operations on iconic models that implement the needed degree of interpretative freedom.

\paragraph{Rotation and scaling}
\label{subsubsec:rotation-and-scaling}

It seems justified to assume that gestures do not depict the real-world sizes of the objects and events talked about.
That means formally that iconic models can be subject to scalar multiplication before they intersect with gesture spaces from spatial domains of speech.
Multiplication of a three-dimensional vector \bvec{v} with scalar $k$ is defined as in (\nextx).
\ex 
$\bvec{v} = \langle x,y,z\rangle, k \in \mathbb{N}$, then $\bvec{v}k = \langle xk, yk, zk\rangle$
\xe 

Since the output of $\gestvec(\gamma)$ is an \textsc{avm} which has a vector description only as a value for the feature \enquote*{traj}, we notate scalar multiplication of an iconic model as $\gestvec(\gamma).\text{traj} \cdot k$, where \enquote*{.traj} is the path to the value of the attribute labelled \enquote{traj}.
The result of this operation is a description of a vector of virtually any length. 
This is an overgeneralization, but a harmless one: it seems plausible to assume that outrageous vectors and vector sequences simply do not occur in realistic models of a model-theoretic semantics, or if they do occur, then only in worlds that are very dissimilar to the actual one.  

Similarly, the orienting axes of the spatial domains of evaluation do not have to coincide with the orientation of gesture space along the anatomical planes of the speaker.
In fact, the opposite is to be expected, as long as the viewpoint of the speaker does not matter (on this shortly).
Change of orientation is equivalent to rotation.
Since gesture spaces on our account are three-dimensional vector spaces, three types of rotation are possible, namely rotation around the $x$, $y$, or $z$ axis (that is, around the \bvec{-FT}/\bvec{FT}, \bvec{-RT}/\bvec{RT}, and \bvec{-UP}/\bvec{UP} levels, respectively).
The corresponding rotation matrices for some angle $\theta$ are given in (\nextx):

\pex \label{ex:rotation-matrix} 
\a $R_x(\theta) = 
\begin{bmatrix}
    1 & 0 & 0 \\ 
    0 & \cos\theta & -\sin\theta \\
    0 & \sin\theta & \cos\theta
\end{bmatrix}$

\a $R_y(\theta) = 
\begin{bmatrix}
    \cos\theta & 0 & \sin\theta \\
    0 & 1 & 0 \\
    -\sin & 0 & \cos\theta
\end{bmatrix}$

\a $R_z(\theta) = 
\begin{bmatrix}
    \cos\theta & -\sin\theta & 0 \\ 
    \sin\theta & \cos\theta & 0 \\
    0 & 0 & 1 
\end{bmatrix}$
\xe

We notate the rotation of an iconic model as follows: $\gestvec(\gamma).\text{traj} \cdot R_d(\theta)$, where $d$ is one of the dimensions $x,y,z$.

But which axis should be rotated, if not all of them? 
We hypothesize -- backed by our empirical examples -- that gestures exhibit a horizontal--vertical anisotropy:
\pex \textbf{Horizontal--Vertical Anisotropy} \label{ex:anisotropy}
\a If a gesture moves through two dimensions, then the \enquote{unused} third dimension is free to rotate. Example: the \enquote{roof gesture} (\ref{ex:roof}) moves through the $\langle x,y \rangle$ subspace, so rotation around the $z$-axis, if at all, is preferable.

\a A three-dimensional gesture is more likely to rotate horizontally than vertically. Example: the representational \enquote{bowl gesture} (\ref{ex:gesture-example-bowl}) consists in a three-dimensional handshape configuration. Rotating this shape around the $z$-axis seems more likely than tilting it in one direction or other. 
\xe 

Therefore, a rotation around the $z$-axis is preferable in both cases, that is, the horizontal plane of gesture space is kept aligned with the horizontal planes of the vector spaces of the interlocutors.
It is tempting to explain the anisotropy in iconic gestures by the anisotropy in neurocognition and vision \citep{Hinterecker:et:al:2018,Hansen:Essock:2004}, but we must leave it at speculation.
%

\paragraph{Perspective}
\label{subsubsec:perspective}

Speakers and the origins of their gesture spaces are connected by a place vector that is aligned with the \bvec{FT} level -- see \cref{fig:gesture-space}.
Objects and events carry an oriented vector space -- see \crefrange{fig:space-of-glass}{fig:throw}.
%
%
Accordingly, the orientation of the place vector in relation to the anatomical planes already determines speaker perspective and defines the indexical reference frame for relative locations. 
Gestures such as event drawings often exploit speaker perspective: if the gesture is performed on the \bvec{RT} dimension in gesture space, the depicted event happened on the right of the speaker, if the gesture is performed on the \bvec{FT} dimension, it happened in front of the speaker, and so forth.
If the perspective is fixed by the speaker's viewpoint, then rotation is blocked and the intersection of the gesture vector or vector sequence and the spatial domain is orientationally faithful to the iconic model.
A perspectival iconic model is defined as follows: $\gestvec(\gamma).\text{traj} \cdot R_d(0)$ (i.e., a model with zero rotation).
We often simply omit the multiplication with a rotation matrix to notate a perspectival iconic model.


\paragraph{Handshape quotation}
\label{subsubsec:quoting-handshapes}

\begin{figure}
\begin{floatrow}[2]
    \ffigbox[\FBwidth]{%
    \begin{adjustbox}{max width=0.48\textwidth}
    \begin{tikzpicture}[]
              \draw[axis, gray] (0,0,0) -- (0,2,0); 
      \draw[axis, gray] (0,0,0) -- (0,-2,0); 
      \draw[axis, gray] (0,0,0) -- (2,0,0); 
      \draw[axis, gray] (0,0,0) -- (-2,0,0) node [at end, below] {\bvec{FT}};
      \draw[axis, gray] (0,0,0) -- (0,0,-2); 
      \draw[axis, gray] (0,0,0) -- (0,0,2); 
        \node [inner sep=0pt, rotate=100, scale=3] (hand) {\faFistRaised};
        \draw [vector] (hand.north) -- +(180:1cm); 
        \node [scale=5, rotate=100] at (-2,0.2,0) {\Knife};
    \end{tikzpicture}
    \end{adjustbox}
    }{%
    \caption{Throwing a dagger with fingers pointed at each other.}
    \label{fig:throw-fist}
    }
    \ffigbox[\FBwidth]{%
    \begin{adjustbox}{max width=0.48\textwidth}
    \begin{tikzpicture}[]
              \draw[axis, gray] (0,0,0) -- (0,2,0); 
      \draw[axis, gray] (0,0,0) -- (0,-2,0); 
      \draw[axis, gray] (0,0,0) -- (2,0,0); 
      \draw[axis, gray] (0,0,0) -- (-2,0,0) node [at end, below] {\bvec{FT}};
      \draw[axis, gray] (0,0,0) -- (0,0,-2); 
      \draw[axis, gray] (0,0,0) -- (0,0,2); 
        \node [inner sep=0pt, rotate=110, scale=3] (hand) {\faHandHolding};
        \draw [vector] (hand.north) -- +(180:1cm);
        \node [scale=5, rotate=100] at (-2,0.2,0) {\Knife};
    \end{tikzpicture}
    \end{adjustbox}
    }{%
    \caption{Throwing a dagger like a ball with open hand (arguably very difficult anyway).}
    \label{fig:throw-open}
    }
    \end{floatrow}
\end{figure}

The acting mode of representation classifies miming gestures within the iconic dimension.
One example is discussed in terms of throwing a dagger in (\ref{ex:throw-dagger}): 
\includegraphics[width=2cm]{gesture-throw-dagger}.
The vectorization is as follows:
\begin{avm}
\[
hs & P \\
traj & $\bvec{FT}$ \\ 
sync & traj[a] $\neq$ traj[z]
\]
\end{avm}. 
The special status of pantomime or acting can be made more precise in terms of spatial gesture semantics: A purely vectorial interpretation of gestural movement -- which seems to offer the right degree of resolution for other modes of representation -- goes only halfway for acting and does not, for instance, distinguish between the regular throwing of a dagger with a grip through the fingers (\cref{fig:throw-fist}, assuming the fist icon represents a knife-throwing hand) and throwing with the palm (\cref{fig:throw-open}, which is probably very difficult to accomplish).
The missing piece for acting gestures obviously is an interpretation of the handshape.
The point of miming is that the mime uses his physical actions to denote physical actions of the same kind.
That is, miming is a form of direct quotation. 
%
On top of the vectorial interpretation therefore is a quotational component which indicates that the handshape of the gesture quotes the handshape of the mimed action. 
%
A quotation operation in sign language has been fomulated by \citet{Davidson:2015-quotation} in terms of
a \textit{demonstration} relation.
We restrict such a demonstration, \enquote*{HSQ} (\textit{handshape quotation}), to handshapes:\footnote{In (\ref{ex:hsq}), HSQ replaces the preposition \textit{like} from Davidon's original formulation (p.~487). The \emph{demonstration} relation is due to the influential account of quotation of  \citet{Clark:Gerrig:1990} that involves demonstrative reference. \citet{Ginzburg:Cooper:2014-quotation} have argued that demonstrations in such cases are grounded in similarity, an approach that was later (presumably unaware of this source) applied to nonverbal gestures \citep{Ebert:Hinterwimmer:2022}.} 
\ex 
$\mng{\text{HSQ}} = \lambda g . \lambda e [\textit{demonstration}(g,e)]$
\label{ex:hsq}
\xe 

In (\lastx), $g$ is the actual gesture, respectively the handshape used in gesturing, and $e$ is the handshape of the quoted action. 
if $g_1$ is handshape \enquote*{P} (M3D's label for \textit{fingers pointed together} from the dagger throwing gesture), then $\mng{\text{HSQ}}(g_1) = \lambda e [\textit{demonstration}(g_1, e)]$, i.e., the set of events that \enquote{are like} \enquote*{P}.
This additional quotational requirement for acting gestures is sufficient to evaluate the multimodal utterance from (\ref{ex:throw-dagger}) as false in situations such as  \cref{fig:throw-open}, where situations such as in \cref{fig:throw-fist} remain potential denotations, as desired.

%
The quotational analysis also works well in the context of non-existing events such as the negation example from (\ref{ex:hit}), where the utterance \enquote{no one could hit} is accompanied by a punching gesture.
The movement presupposes a place path, the gesture is supposed to mimic the handshape used in hitting. 
This meaning can be represented as follows (ignoring tense): 
\ex 
$\exists e [\neg\exists x . \text{hit}(e,x) \wedge \exists \bvec{v}[k][\placepath(\bvec{v}[k],e)] \wedge \textit{demonstration}(\text{F},e)]$, where \enquote*{F} is the handshape label for \textit{fist}. 
\xe 

(\lastx) reads as \textit{no one could hit like this}, where \enquote{like this} is demonstrated by the speaker/gesturer and picks out all events where there is no hitting at all (no hitting agent), where hitting does not leave a path, or where hitting looks different from \enquote{F}, which is intuitively correct. 
In general, it is possible to quote hypothetical physical actions, just as it is to quote hypothetical utterances \citep{Sams:2010-quotation,Fotiou:2024-quotation}.
Handshape quotation is expressed for iconic models as follows: $\mng{\text{HSQ}}(\gestvec(\gamma).\text{hs})$.


\end{RevisionEnv}
%
%

\subsubsection{Compositional speech--gesture integration}
\label{subsec:compositional-speech-gesture-integration}

The previous sections developed step-by-step a spatial iconic gesture semantics that is needed in linguistic theory in light of multimodal studies.
Here it is briefly shown how the visual level of meaning interacts with the linguistic level of meaning. 
As the compositional \enquote{backbone}, we assume a multimodal grammar  \citep{Alahverdzhieva:Lascarides:Flickinger:2017,Luecking:2013:a}.
Such grammar extensions combine an iconic gesture with its lexical or phrasal affiliate.
%
Schematically, a multimodal utterance $\alpha[\beta/\gamma]$ consisting of a sentence $\alpha$, a gesture $\gamma$ and its affiliate $\beta$ is true, iff $\alpha$ is true and there is an \Revision{embedding} of the iconic model of $\gamma$ -- \Revision{possibly transformed by scaling or rotation, and possibly additionally constrained by perspective or quotation} -- into the spatial configuration \enquote*{$\vecspace(\mng[e]{\beta})$} projected from $\mng[e]{\beta}$.\footnote{For the sake of simplicity, we assume that a multimodal utterance involves just one gesture, but the set-up scales up straightforwardly.}

%
%
\begin{RevisionEnv}
As argued in detail in \cref{sec:introduction}, visual information and linguistic meaning are not on a par.
A straightforward solution is to keep apart compositional linguistic semantics and a visual component in the lexicon.\footnote{This move is also supported by similar arguments from dual coding and lexical semantics \citep[\S6.4]{Pustejovsky:Batiukova:2019-lexicon}.}
%
%
Within the following tree-structure derivation, the dimensions are indicated by the labels \enquote*{[ling]} and \enquote*{[vis]}, respectively.
Technically, they are operators that \enquote{encapsulate} the terms in their scope.\footnote{This is related to the formal treatment of discourse markers in multimodal discourse of \citet{Lascarides:Stone:2009:a}: Discourse markers for gestures are introduced in the scope of an operator $[\mathcal{G}]$; $[\mathcal{G}]$ ensures that gestural discourse markers only change the output context, but not the input context. }
%
%
The following rules of functional application and set union apply: 
\pex \label{ex:rules}
\a $\kappa$ is a branching node and $\{\alpha, \beta\}$ the set of its daughters.
\par 
(i) If $[\text{ling}]\mng{\beta}$ is a function whose domain contains $[\text{ling}]\mng{\alpha}$, then $[\text{ling}]\mng{\kappa} = [\text{ling}]\mng{\beta}([\text{ling}]\mng{\alpha})$.
\par 
(ii)
$[\text{vis}]\mng{\kappa} = [\text{vis}]\mng{\alpha} \cup [\text{vis}]\mng{\beta}$ -- the vector representations of the daughters are merged into the set of visual meanings of the mother node.  

\a Existential closure: $S$ is a sentence node. Then there is a node $S'$ whose daughters are $S$ and $C = [\text{ling}]\lambda R. \exists e [R(e)]$ and $D = [\text{vis}]\lambda R. \exists \cornerquote{\bvec{v}} [R(\cornerquote{\bvec{v}})]$ such that 
\par
(i) $[\text{ling}]\mng{S'} = C([\text{ling}]\mng{S})$, and
\par 
(ii) for each lambda term $\nu$ in $[\text{vis}]\mng{S}$, $D(\nu)$ is part of $[\text{vis}]\mng{S'}$.

\a $\omega$ is a multimodal node and $\alpha$ its speech, $\gamma$ its gesture daughter. Then $[\text{vis}]\mng{\omega} \ni [\text{vis}]\mng{\alpha}([\text{vis}]\mng{\gamma})$ for some $[\text{vis}]\mng{\alpha}$ in the set of $\alpha$'s visual meanings, $[\text{vis}]\mng{\gamma}$  is in the domain of $[\text{vis}]\mng{\alpha}$. Since affiliation is primarily lexical (see \cref{sec:gesture-primer}), $[\text{vis}]\mng{\alpha}$ is usually a singleton set anyway.

\a $[\text{vis}]\mng{\gamma} \text{ is in the domain of } [\text{vis}]\mng{\alpha}$ if $[\text{vis}]\mng{\gamma} \subset D_v$ and 
\par 
(i) the vector space of $\alpha$ is of type $\vecspace(e), e \in D_s$ and $\gamma$ is a drawing or acting gesture
\par 
(ii) the vector space of $\alpha$ is of type $\vecspace(x), x \in D_e$ and $\gamma$ is a drawing or molding gesture
\xe


The rules in (\lastx) derive  the compositional meaning of speech as usual. 
Lexical items may have a non-empty \enquote*{[vis]} dimension that contributes a vector meaning.
Given the spatial models introduced in \cref{subsec:vector-space-semantics}, these vector meanings are usually trivially true.
However, when combined with an iconic gesture, the vectorial meaning is descriptively enriched, leading to \enquote{stronger} truth conditions (i.e., they cancel out more circumstances of evaluation). 
Multimodal integration that captures the semantic interaction of speech and gesture while maintaining the \enquote{visual innocence} of the gesture as specified in (\lastx) is exemplified in \cref{fig:mm-integration} by using an example closely related to \textit{dagger-throwing} from (\ref{ex:throw-dagger}). 
%
For simplification and reasons of space, the indefinite \textit{a dagger} and the pronoun \textit{he} are both treated as expressions of type $e$.\footnote{Generalized quantifiers in object position cannot be interpreted immediately in common semantic formalisms (type coercion or floating must first be applied). We consider our simplification to be unproblematic, since QNPs in object position can be interpreted as expected in newer theories on pluralities and quantification \citep{Luecking:Ginzburg:2022-rtt}.}

\begin{figure}[htb]
\begin{adjustbox}{max width=\linewidth}
\begin{forest}
baseline
[{S$'$ \\ {[ling]}$\exists e [\text{throw}(e) \wedge \text{ag}(e) = \text{he} \wedge \text{th}(e) = \text{dagger}]$ \\ {[vis]}$\{\placepath(e,[\bvec{FT} \cdot k \wedge \mng{\text{HSQ}}(K)])$, \\ $\exists \bvec{u} \in \vecspace(\text{dagger}) [\axis(\bvec{u},y)]\}$}, align=center
[{$[\text{ling}]\lambda R. \exists e[R(e)]$ \\ $[\text{vis}]\lambda R. \exists \cornerquote{\bvec{v}} [R(\cornerquote{\bvec{v}})]$.}, align=center [existential closure]]
[{S \\ {[ling]}$\lambda e [\text{throw}(e) \wedge \text{ag}(e) = \text{he} \wedge \text{th}(e) = \text{dagger}]$ \\ {[vis]}$\{\placepath(e,[\bvec{FT} \cdot k \wedge \mng{\text{HSQ}}(K)])$, \\ $\lambda \bvec{u} \in \vecspace(\text{dagger}) [\axis(\bvec{u},y)]\}$}, align=center
[{PRON \\ $[\text{ling}]\text{he}$}, align=center [he]]
[{VP \\ {[ling]}$\lambda x. \lambda e [\text{throw}(e) \wedge \text{ag}(e) = x \wedge \text{th}(e) = \text{dagger}]$ \\ {[vis]}$\{\placepath(e,[\bvec{FT} \cdot k \wedge \mng{\text{HSQ}}(K)])$, \\ $\lambda \bvec{u} \in \vecspace(\text{dagger}) [\axis(\bvec{u},y)]\}$}, align=center
[{MM \\ {[ling]}$\lambda y. \lambda x. \lambda e [\text{throw}(e)$ \\ $\wedge \text{ag}(e) = x \wedge \text{th}(e) = y]$ \\ {[vis]}$\{\placepath(e,[\bvec{FT} \cdot k \wedge \mng{\text{HSQ}}(K)])\}$}, align=center 
[{V \\ {[ling]}$\lambda y. \lambda x. \lambda e [\text{throw}(e)$ \\ $\wedge \text{ag}(e) = x \wedge \text{th}(e) = y]$ \\ {[vis]}$\{\lambda \bvec{v} \in \vecspace(e) [\placepath(e,\bvec{v})]\}$}, align=center [throw]]
[{$\gamma$\textsubscript{acting}  \\  {[vis]}$[\bvec{FT} \cdot k \wedge \mng{\text{HSQ}}(K)]$}, align=center]
]
[{NP \\ {[ling]}$\text{dagger}$ \\ {[vis]}$\{\lambda \bvec{u} \in \vecspace(\text{dagger}) [\axis(\bvec{v},y)]\}$}, align=center [a dagger]]
]
]
]
\end{forest}
\end{adjustbox}
\caption{Derivation of \textit{He threw a dagger} $+$ throwing gesture $\gamma_\text{acting}$.}
\label{fig:mm-integration}
\end{figure}

The iconic model in \cref{fig:mm-integration} is perspectival, so no rotation is involved.
The dimension of rotation follows the anisotropy heuristics (\ref{ex:anisotropy}), that is, unless otherwise specified, the iconic model is only rotated around the $z$-axis.
The top node in (\lastx) is true if there is an event of \enquote{he} throwing a dagger, throwing is done with handshape F in the speaker's line of sight, and the dagger has an axis (the latter not being specified further).

In general, a multimodal utterance $\alpha[\beta/\gamma]$ including an iconic gesture $\gamma$ and its affiliate $\beta$ is true iff  $\mng{[\text{ling}]\alpha} = 1$ and the (potentially transformed) iconic model of $\gamma$ can be embedded within the spatial projection of $\mng[e]{\beta}$, and, if applicable, it holds that $\mng{\text{HSQ}}(\gestvec(\gamma).\text{hs})$.
Note that the rules in (\ref{ex:rules}) provides us with a first formal notion of \emph{multimodal well-formedness}: a speech--gesture mismatch occurs if there is no \bvec{v} which embeds the iconic model into the spatial configuration projected from the verbal affiliate.
Hence, spatial theories give rise to a well-behaved, truth-functional semantics of iconic gestures as a conservative extension of a Montagovian semantic framework.

\subsubsection{Analyzing some examples}
\label{sec:examples}

Let us apply the formal system more or less informally to some of the examples collected in \cref{subsec:empirical-data}.
\newcommand\simplecuboid[3]{%
\fill[gray!50!white] (tpp cs:x=0,y=0,z=#3)
-- (tpp cs:x=0,y=#2,z=#3)
-- (tpp cs:x=#1,y=#2,z=#3)
-- (tpp cs:x=#1,y=0,z=#3) -- cycle;
\fill[gray!80!white]
(tpp cs:x=0,y=0,z=0)
-- (tpp cs:x=0,y=0,z=#3)
-- (tpp cs:x=0,y=#2,z=#3)
-- (tpp cs:x=0,y=#2,z=0) -- cycle;
\fill[gray!30!white] (tpp cs:x=0,y=0,z=0)
-- (tpp cs:x=0,y=0,z=#3)
-- (tpp cs:x=#1,y=0,z=#3)
-- (tpp cs:x=#1,y=0,z=0) -- cycle;}

\newcommand\door[3]{%
\fill[green!70!black] (tpp cs:x=0,y=0.5,z=0)
-- (tpp cs:x=0,y=0.5,z=#3)
-- (tpp cs:x=0,y=#2,z=#3)
-- (tpp cs:x=0,y=#2,z=0) -- cycle;}

\newcommand\window[3]{%
\fill[blue!50!white] (tpp cs:x=#1,y=#2,z=#3)
-- (tpp cs:x=#1+0.5,y=#2,z=#3)
-- (tpp cs:x=#1+0.5,y=#2,z=#3+0.5)
-- (tpp cs:x=#1,y=#2,z=#3+0.5) -- cycle;}

\newcommand\locatedcuboid[6]{%
\fill[gray!50!white] (tpp cs:x=#1,y=#2,z=#3+#6)
-- (tpp cs:x=#1+#4,y=#2,z=#3+#6)
-- (tpp cs:x=#1+#4,y=#2+#5,z=#3+#6)
-- (tpp cs:x=#1,y=#2+#5,z=#3+#6) -- cycle;
\fill[gray!30!white] (tpp cs:x=#1,y=#2,z=#3+#6)
-- (tpp cs:x=#1+#4,y=#2,z=#3+#6)
-- (tpp cs:x=#1+#4,y=#2,z=#3)
-- (tpp cs:x=#1,y=#2,z=#3) -- cycle;
\fill[gray!80!white] (tpp cs:x=#1,y=#2,z=#3+#6)
-- (tpp cs:x=#1,y=#2+#5,z=#3+#6)
-- (tpp cs:x=#1,y=#2+#5,z=#3)
-- (tpp cs:x=#1,y=#2,z=#3) -- cycle;}

\newcommand{\locatedxwindow}[6]{%
\fill[blue!50!white] (tpp cs:x=#1,y=#2,z=#3+#6)
-- (tpp cs:x=#1+#4,y=#2,z=#3+#6)
-- (tpp cs:x=#1+#4,y=#2,z=#3)
-- (tpp cs:x=#1,y=#2,z=#3) -- cycle;}

\newcommand{\locatedywindow}[6]{%
\fill[blue!60!gray] (tpp cs:x=#1,y=#2,z=#3+#6)
-- (tpp cs:x=#1,y=#2+#5,z=#3+#6)
-- (tpp cs:x=#1,y=#2+#5,z=#3)
-- (tpp cs:x=#1,y=#2,z=#3) -- cycle;}

\Crefrange{fig:flat-roof}{fig:u-shaped-house} depict two spatial projections of houses, which have to be considered when interpreting the demonstration from example (\ref{ex:u-shaped}), \includegraphics[trim=50 0 150 180, clip, width=1.5cm]{mmsubcat2.png}, producing a U-shaped iconic model, 
\begin{avm}
\[
hs & C \\
traj & $\bvec{RT}\perp\bvec{-FT}\perp\bvec{-RT}$ \\ 
sync & traj[a] $\neq$ traj[z]
\]
\end{avm}. 
%
%
The iconic model describes the axis-path of the house talked about (modulo rotation and scaling). 
The gesture, therefore, excludes spatial projections with a straight main axis, as those in \cref{fig:flat-roof}; it restricts the spatial models to those involving houses exhibiting a layout as shown in \cref{fig:u-shaped-house}.

\begin{figure}[htb]
\begin{tikzpicture}[3d view]
      \draw[axis, gray] (2,1.5,0.7) -- (2,1.5,-1); 
      \draw[axis, gray] (2,1.5,0.7) -- (2,6.5,0.7); 
      \draw[axis, gray] (2,1.5,0.7) -- (5.5,1.5,0.7); 
    \simplecuboid{4}{3}{1.3}
    \door{3}{1.25}{1}
    \window{1}{0}{0.5}
    \window{2}{0}{0.5}
    \window{3}{0}{0.5}
    \draw[axis, gray] (2,1.5,1.4) -- (2,1.5,2.5); 
    \draw[axis, gray] (2,0,0.7) -- (2,-3.5,0.7); 
    \draw[axis, gray] (0,1.5,0.7) -- (-1.5,1.5,0.7); 
    \draw [white, very thick] (0,1.5,1.4) -- (4,1.5,1.4);
\end{tikzpicture}
    \caption{Flat-roofed building. White line showing main axis vector.}
    \label{fig:flat-roof}
 \end{figure}
 

\begin{figure}[htb]
\begin{tikzpicture}[3d view]
      \draw[axis, gray] (0,0,0.5) -- (0,5,0.5); 
      \draw[axis, gray] (0,0,0.5) -- (4,0,0.5); 
      \draw[axis, gray] (0,0,0.5) -- (0,0,-1); 
      \locatedcuboid{0}{1}{0}{2}{1}{1}
      \locatedcuboid{0}{-2}{0}{2}{1}{1}
      \locatedcuboid{-1}{-2}{0}{1}{4}{1}
      \locatedxwindow{-0.75}{-2}{0.3}{0.5}{0}{0.5}
      \locatedxwindow{0.25}{-2}{0.3}{0.5}{0}{0.5}
      \locatedxwindow{1.25}{-2}{0.3}{0.5}{0}{0.5}
      \locatedywindow{-1}{-1.5}{0.3}{0}{0.5}{0.5}
      \locatedywindow{-1}{-0.5}{0.3}{0}{0.5}{0.5}
      \locatedywindow{-1}{0.5}{0.3}{0}{0.5}{0.5}
        \locatedywindow{-1}{1.5}{0.3}{0}{0.5}{0.5}
      \draw[axis, gray] (0,-2,0.5) -- (0,-5,0.5);  
     \draw[axis, gray] (-1,0,0.5) -- (-2.5,0,0.5); 
       \draw[axis, gray] (0,0,1) -- (0,0,2.5);
      \draw [white, very thick] (2,1.5,1) -- (-0.5,1.5,1) -- (-0.5,-1.5,1) -- (2,-1.5,1);
      \end{tikzpicture}
\caption{U-shaped building. White line showing main axis vector.}
\label{fig:u-shaped-house}
\end{figure}

An \enquote{event drawing} gesture was given in (\ref{ex:pulled-out}), namely \includegraphics[width=1.5cm]{gesture-pulling-out.png}. 
The event talked about is a car pulling out \emph{in front of} the speaker, fixing a speaker perspective by forcing zero rotation. 
The gesture's iconic model describes a vector in the direction of the \bvec{-RT} level,
\begin{avm}
    \[
hs & K \\
traj & $\bvec{-RT}$ \\ 
sync & traj[a] $\neq$ traj[z]
\]
\end{avm}, that is, towards left in gesture space.
The direction in this case is important, since the speaker introduces a viewpoint (namely, in front of him).
From possible \textit{car-pulling-out}-events -- two of which are displayed in \crefrange{fig:car-right}{fig:car-left} -- the gesture cancels those of the perspectival type of \cref{fig:car-left}, the speaker's relative position is marked by \enquote*{$S$}.

\begin{figure}
\begin{floatrow}[2]
\ffigbox[\FBwidth]{%
\begin{tikzpicture}[3d view]
      \draw[axis, gray] (0,0,0) -- (0,1.5,0); 
      \draw[axis, gray] (0,0,0) -- (0,-2.5,0); 
      \draw[axis, gray] (0,0,0) -- (1.5,0,0) node [right] {\bvec{RT}};
      \draw[axis, gray] (0,0,0) -- (-1.5,0,0); 
      \draw[axis, gray] (0,0,0) -- (0,0,-1.5) node [at end, right, black] {$S$};
      \draw[axis, gray] (0,0,0) -- (0,0,1); 
      \node [scale=4, anchor=north, inner sep=0pt, fill=white] (0,0,0) {\faRoad};
      \node [scale=2, xscale=-1, inner sep=0pt] (car) at (3,0,0) {\faCarSide};
      \draw [vector] (car) -- +(180:2cm);
    \end{tikzpicture}
}{%
\caption{Car pulling out from right.}
\label{fig:car-right}
}
\ffigbox[\FBwidth]{%
\begin{tikzpicture}[3d view]
      \draw[axis, gray] (0,0,0) -- (0,1.5,0); 
      \draw[axis, gray] (0,0,0) -- (0,-2.5,0); 
      \draw[axis, gray] (0,0,0) -- (1.5,0,0); 
      \draw[axis, gray] (0,0,0) -- (-1.5,0,0) node [left] {$-\bvec{RT}$};
      \draw[axis, gray] (0,0,0) -- (0,0,-1.5) node [at end, right, black] {$S$};
      \draw[axis, gray] (0,0,0) -- (0,0,1); 
      \node [scale=4, anchor=north, inner sep=0pt, fill=white] (0,0,0) {\faRoad};
      \node [scale=2, inner sep=0pt] (car) at (-3,0,0) {\faCarSide};
      \draw [vector] (car) -- +(0:2cm);
      \end{tikzpicture}
}{%
\caption{Car pulling out from left.}
\label{fig:car-left}
}
\end{floatrow}
\end{figure}

The action gesture \enquote{throw a dagger}
(\ref{ex:throw-dagger}), \includegraphics[width=1.5cm]{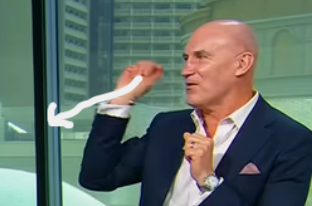}, finally, not only introduces a speaker perspective, but also a handshape quotation.
The derivation of this example is shown in \cref{fig:mm-integration}.
Accordingly, the corresponding iconic model, 
\begin{avm}
\[
hs & K \\
traj & $\bvec{FT}$ \\ 
sync & traj[a] $\neq$ traj[z]
\]
\end{avm}, 
applies without rotation (but scaling is still possible) and excludes circumstances of evaluation where a different handshape is involved, as in \cref{fig:throw-open}.
The gesture selects for throwing events of the type in \cref{fig:throw-fist}, where fingers are pointed together, and the dagger is thrown straight ahead.


\end{RevisionEnv}

\subsection{Related work}
\label{sec:related-work}

The vector space semantics takes some inspiration from mereotopological approaches, which, however, so far remain computationally incomplete and presumably overestimate the precision of gestural representations (\cref{subsec:spatial-frames}).
The same applies to 
projective or pictorial semantics,  
which was proposed as semantics for iconic gestures by \citet{Schlenker:2019-gestural}.
Furthermore, \cref{subsec:projective-semantics} argues that projection is a non-starter for iconic gestures anyway, in particular acting ones. 
%

\subsubsection{Mereotopological domains}
\label{subsec:spatial-frames}

%
%
%
The spatial gestures semantics developed by \citet{Giorgolo:2010-phd} makes use of a rather powerful mereotopological language for modeling a spatial domain.
%
On this account, an iconic gesture is supposed to be mapped via a procedure $\phi$ onto an iconic space, that is, a mereotopological rendering of the gesture's kinematic representation. 
The iconic space bears an equivalence relation (under some fixed perspective) to the spatial domain -- capturing the iconic gestures' visual contribution. 
The integration of speech and gesture is intersective: a gesture assigns the arguments of its verbal affiliate (which can be an $n$-ary predicate or an $n$-ary predicate modifier -- a grammar of affiliation is presupposed on this account) a common subspace in the spatial domain and requires this subspace to be equivalent to the iconic space, that is, the output of $\phi$.
However, a spell-out for $\phi$ (in contrast to our \enquote*{$\gestvec(\gamma)$}) is not provided, leaving the account incomplete.
One reason for this shortcoming probably is that the mereotopological approach does not distinguish different modes of representation and primary dimensions of depiction (cf. \cref{sec:gesture-primer}), which have proven to be important for gesture studies as well as for our vector space semantics of iconic gestures. 
%
Additionally, we speculate that a detailed mereotopology is too precise for manual gestures, which are often somewhat \enquote{sloppy}.



\subsubsection{Applying projection semantics?}
\label{subsec:projective-semantics}

A truth-functional approach to the meaning of figurative drawings\footnote{\enquote{Canonical examples of pictures include architectural and engineering drawings, figurative paintings, functional illustrations sketches and illustrations, photographs, as well as many kinds of maps.} \citep[p.~849]{Greenberg:2021}} has been developed by \citet{Greenberg:2011,Greenberg:2021} in terms of projection semantics, or pictorial semantics.
It has been suggested that a projection semantics offers an account for sign language, too, namely \enquote{iconological semantics} \citep{Schlenker:Lamberton:2024-iconological}.
\citet[744]{Schlenker:2019-gestural} suggests to apply projection semantics to iconic gestures.
However, this proposal does not go far enough, as we will show below.
A projection is a geometric projection in the sense of the artist's technique to construct perspectival paintings.\footnote{See, for instance, Albrecht Dürer's \textit{Underweysung der Messung}, in particular illustrated in the final image of the fourth volume (available at   \url{https://de.wikisource.org/wiki/Underweysung_der_Messung,_mit_dem_Zirckel_und_Richtscheyt,_in_Linien,_Ebenen_unnd_gantzen_corporen/Viertes_Buch}).} 
Projection semantics uses such geometric projections to interpret pictures. 
A picture on this account expresses a content (a pictorial space), if the picture is a projection of that content (relative to a viewpoint, that is, a pair of projection source and picture plane) -- see \cref{fig:projection-system} for a simple example.\footnote{Formally: $\mng{P}_{S,c} \subseteq \{\langle w, v \rangle \mid \textit{proj}_S(w,v) = P\}$ (the denotation of a picture $P$ in context $c$ is a subset of the world--viewpoint pairs $\langle w, v \rangle$ (scenes) such that there is a projection from the world--viewpoint pairs onto $P$ relative to a system of depiction $S$).
}
Projection semantics is semantic because it offers a notion of accuracy: a picture is accurate iff (abbreviates if and only if) there is a viewpoint that provides a geometric projection from the pictorial space to the picture.
Viewpoints can straightforwardly be embedded in standard possible world semantics: they provide a geometric system on top of the worlds already part of the model's domains. 
Further refinements can, or need to, be given in terms of different systems of depiction (in particular \enquote{impurely projective} ones like caricature; \cref{fig:projection-system} only shows simple line drawing) and depth constraints, among others \citep[cf.][]{Greenberg:2021}.

\begin{figure}
    \centering
\begin{tikzpicture}
    \node [circle, inner sep=0pt, minimum size=2pt, fill=black, label=above:source] (source) at (0,0) {};

    \node [circle, inner sep=0pt, minimum size=1cm, ball color=black, label=below:world] (world) at (5,-1) {};

    \begin{scope}[black!60, dashed]
    \draw (source) -- (world.north);
    \draw (source) -- (world.south) node [sloped, near start, below] {projection lines};
    \end{scope}

    \filldraw[draw=black, fill=none] (3,0) -- (4,0.5) node [at end, above] {picture plane} -- (3.5,-1.5) -- (2.5,-2) -- cycle;

 \node [circle, draw, minimum size=0.65cm, xscale=0.8] at (3.3,-0.65) {};
\end{tikzpicture}    
\caption{Geometric projection: The pair of projection source (determining perspective) and picture plane (determining orientation) is a viewpoint. A viewpoint and a world define a scene. The set of all such scenes is the pictorial space, the content of a picture.}
    \label{fig:projection-system}
\end{figure}

Does projection semantics also provide an adequate framework for analyzing iconic gestures?
A correspondence between paintings and gestures is obtained straightforwardly: the gesture plays the role of the picture plane which displays the projection source. 
The content of the gesture, as with pictures, then depends on the depiction system to be at work. 
Given the basic figurative projection function of projection semantics, this means that the content of a gesture is the set of world--viewpoint pairs where a (possibly different, let us assume) gesturer performs that gesture, seen from the viewpoint in question. 
That is, the content of a gesture is in turn a gesture! 
Such a verbatim interpretation obviously misses the semiotic point of a gesture: the content of a gesture surely is not a look-alike gesturing situation.
In other words, projection semantics, when applied to iconic gestures, fails to distinguish between the gesture as a physical action and the content of the gesture.
%

Projective semantics tries to avoid this pitfall by claiming that projections are selective: The input for the projection function does not have to be a complete object or scene, but can be limited to certain features of it. 
In the visual domain, for example, a projection can project only the outline of an object, ignoring its color.\footnote{If the projection function merely selects what is appropriate in a given context, one might of course ask how useful this concept actually is.}
Such selectivity is needed, most prominently, for acting gestures: a gesture mimicking throwing (see \ref{ex:throw-dagger-exemplification} below) does not actually involve a thrown object.
A gestural projection must therefore abstract over the theme of the throwing action.
However, such a projection cannot be distinguished from a selective projection of a real throwing action, involving an object.
here, too, projective semantics fails to capture the semiotic nature of iconic gestures.


Apart from this fatal semiotic flaw, geometric projections of oriented worlds are confronted with perceptual challenges. 
Consider an iconic gesture that can be interpreted as \textit{rolling}: the index finger rotates in circles while the wrist is moved rightwards. 
The decisive feature is that there is a part of the rotation movement of the index finger that runs backwards (i.e., in the opposing direction of the wrist movement), as illustrated in \cref{subfig:perzmov} (see also \citealp[p.~1088]{Bressem:2013}).
This configuration, however, is a purely perceptual one; it can never be projected onto a physical movement (\crefrange{subfig:rolling-motion}{subfig:pointb}), where \enquote{going back} is simply impossible. 
Thus, geometric projections make wrong predictions with respect to gestures where the perceptual image diverges from its physical origin.

However, this problem is due to the extensional nature of projections.
In \cref{sec:main-informational-evaluation} we develop an intensional, lexical approach that avoids this drawback.
Crucially, the visual aspects of the lexical entry of \textit{rolling} can indeed carry rotational information, which is true for the iconic model of a rotational rolling gesture.

\begin{figure}[tb]
  \subfloat[\label{subfig:rolling-motion}]{%
  \begin{adjustbox}{scale=0.9}
    \begin{tikzpicture}
      \draw[-stealth'] (0,0) arc (150:30:2cm) arc (150:90:2cm);
      \draw (-0.5,0) -- (5.2,0);
      \draw (1,0.5) circle (0.5cm) node [fill, circle, inner sep=1pt,
      label=right:{\footnotesize A}] {};
      \draw (2.5,0.5) circle (0.5cm);
      \node [fill, circle, inner sep=1pt, label=left:{\footnotesize
        B}] at (0.51,0.57) {};
      \node [fill, circle, inner sep=1pt] at (3,0.55) {};
    \end{tikzpicture}
    \end{adjustbox}
  }\hfill%
  \subfloat[\label{subfig:pointb}]{%
  \begin{adjustbox}{scale=0.9}
    \begin{tikzpicture}
      \draw[-stealth'] (0,0) arc (150:30:2cm) arc (150:90:2cm);
    \end{tikzpicture}
    \end{adjustbox}
  }\hfill%
  \subfloat[\label{subfig:perzmov}]{%
  \begin{adjustbox}{scale=0.9}
    \begin{tikzpicture}
      \draw[-stealth'] (0,0) arc (270:-40:0.5cm);
      \draw[-stealth'] (0,0) node [fill, inner sep=1pt, circle,
      label=right:{\footnotesize B}] {} -- (0,0.5) node [fill, inner
      sep=1pt, circle, label=above:{\footnotesize A}] {} --(2,0.5); 
    \end{tikzpicture}
    \end{adjustbox}
  }
  \caption{Perceiving a rolling disc as simultaneous movements of points $A$ and $B$ \protect\cref{subfig:rolling-motion}. The movement trajectory of point $B$ in isolation corresponds to a cycloidal trace \protect\cref{subfig:pointb}. The shared movement of $A$ and $B$ -- a horizontal rightward movement --  is the perceptual reference frame for both individual movements. The resulting \emph{percept} is that $B$ circles around $A$ \protect\cref{subfig:perzmov}. (Reproduced after \protect\citealp[p.~207]{Johansson:1973})
 }
  \label{fig:rolling}
\end{figure}

%

\subsection{Discussion}
\label{sec:discussion-spatial}

The research questions that we must leave open here include the following.

\subsubsection{Static representing gestures}
\label{sec:interpreting-handshape}

The static representing gesture in (\nextx) stands for the discourse referent introduced by the NP \textit{kleinere Schale} \enquote*{smaller bowl} in speech.
While uptake of a discourse referent in speech  
is the defining feature of representing gestures, the handshape provides spatial information about the shape of the object talked about (namely being bowl-like).
But unlike acting gestures, this representing gesture does not quote a handshape; rather the handshape introduces an iconic model in its own right.
That means that gestures can produce an iconic model even when the gesture is displayed statically.
\ex SaGA dialogue V21, starting at 4:38:
\par 
\includegraphics[trim=10 0 0 20, clip, width=3cm]{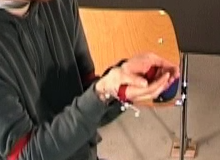} und 'ne kleinere Schale obendrauf  / \textit{and a smaller bowl on top} \\ 
$\Rightarrow$ layout of hands represents main axis of bowl object
\label{ex:gesture-example-bowl}
\xe 

The semantic resource to account for a spatial handshape interpretation is a vectorial interpretation of handshape geometry.
%
%
%
%
%
%
For example, the bowl gesture somehow introduces an arc-shape axis vector, which then can be interpreted in our system as usual.
Since annotation systems for manual gestures usually distinguish only comparatively small number of handshape labels (see \cref{subsec:kinematic-gesture-representation}), a lexical approach might be applicable here, that is, pairing handshape types with vectorial representations so that the vectorial description of a handshape can be read from this lexicon.
A computational solution outlined in \cref{subsec:ai} is to rely on pose detection algorithms developed in computer vision.

\subsubsection{QUD management}
\label{sec:qud-management}

We will also not discuss examples such as (\nextx), which involve question-under-discussion management \citep{Laparle:2021}. 
%
These gestures look like iconic gestures, but pertain to information management rather than 
content.
Accordingly, their proper treatment has to be spelled out within dynamic discourse theories, not static semantics. 
\pex \label{ex:gesture-examples-discourse}
\a Daniel Levitin, \textit{Ted talk}: \label{ex:locate-left-right}
\par 
\textit{some of them are obvious} \includegraphics[width=2.5cm]{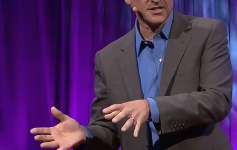}, \textit{some of them are not so obvious} \includegraphics[width=2.5cm]{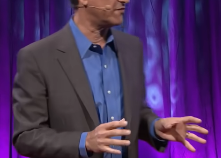} \\
$\Rightarrow$ \textit{on the one hand\ldots on the other hand} figure
\par 
(\url{https://www.youtube.com/watch?v=8jPQjjsBbIc&t=0h3m55s})

\a Kathy Griffin \citep[taken and simplified from][p.~177]{Laparle:2022-phd}:
\par 
\textit{Anyway um .. so so then} \includegraphics[width=3.5cm]{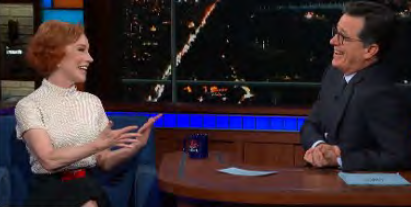} \textit{I decided to promote my own shows} \\
$\Rightarrow$ palm up open hand gesture  towards Colbert (host on the right) shows \enquote{re-engagement with main discourse} (\textit{op. cit.}, p.~177\psq)
\par
(\url{https://www.youtube.com/watch?v=UWKIljRfltM&t=0h7m17s})
\xe


\subsubsection{Two-handed gestures and the internal structure of objects}
\label{sec:two-handed-gestures}

\begin{figure}[tb]
  \begin{minipage}[m]{0.3\textwidth}
    \adjincludegraphics[valign=m, angle=270, trim={9cm 4cm 7cm 17cm}, clip, width=\linewidth]{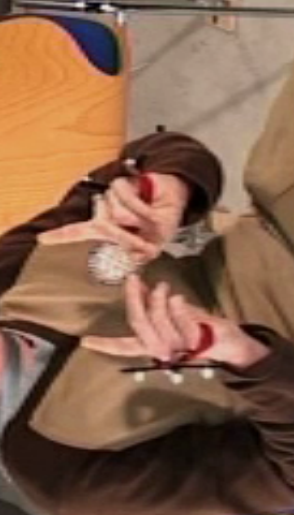}
  \end{minipage}%
  \hfill%
  \begin{minipage}[m]{0.6\textwidth} \raggedright
    \enquote{die rechte Kirche die hat zwei spitze Türme} \\
    \emph{the church to the right it has two pointed towers} 
  \end{minipage}
  \caption{Talking about two church towers (from SaGA dialogue V24, 6:25).}
  \label{fig:two-towers}
\end{figure}

The empirical examples collected in \cref{subsec:empirical-data} mostly involve one-handed gestures.
But there are also two-handed ones.
Take, for instance, the example in \cref{fig:two-towers}, discussed by \citet{Luecking:Ginzburg:2022-rtt}.
The authors claim (intuitively plausible) that each pointing finger represents one of the two church towers talked about. 
This is problematic for a plural semantics that introduces only a plural discourse referent and a cardinality constraint \enquote{two} (accordingly, the authors developed a theory of plurality and quantification that is compatible with gestures).
This example shows that in particular two-handed gestures call for a decomposition of real-world objects (and presumably also events), so that the iconic models of gestures can map onto the internal structures.
    
\subsubsection{Time and intensity}
\label{sec:time}

Obviously, especially event gesture like that in the \textit{car-pulled-out} example (\ref{ex:pulled-out}) not only have spatial but also a temporal interpretation. 
As hinted at in \cref{subsec:vector-space-semantics}, a temporal interpretation of a vector or vector sequence is achieved by mapping points or intervals of time to $\bvec{v}[k]$. 
In addition to spelling this out in an empirically adequate way, many follow-up questions arise for multimodal meaning, for instance, with regard to the interplay of verb aspect and vectors: activities have a different temporal course than accomplishments.
Similarly, gestures that are affiliated to event denoting nouns are expected to reveal the event structure inherent in these nouns.
Careful semantic analyses of these issues are still pending.
The topic of gestures and intensity or expressivity is already a field research \citep[see][]{Ebert:Walter:2026}.

\subsubsection{\enquote{Energy spaces}}
\label{sec:energy-spaces}

The iconic gesture semantics developed above is confined to representational gestures, that is, gestures that exhibit a rather pictorial (aka iconic) content.
%
%
However, a lot of gestures occurring in natural language interactions have a strong impact on the beat dimension and/or are bound up with functional information structuring interpretations (see \cref{sec:qud-management}). 
A mere spatial interpretation seems to miss the point of such a gesture.
Intuitively, they seem to be of a more \enquote{somatic} origin.
\Revision{\citet{Pouw:Fuchs:2022}, for instance, provide evidence for biomechanical linkages of gestural body movements with respiration.}
The vector space semantics, we argue, can be adapted to such gestures, too, following ideas of \citet{Talmy:1988} \Revision{and \citet{Goldschmidt:Zwarts:2016-force}.}
These authors envisaged the use of \emph{force vectors} instead of spatial place or path ones. 
One of the examples discussed by Talmy is the verb \textit{climbing}.
According to his analysis, the semantics of \textit{climb} is captured in terms of two forces: one pulling downwards, one striving upwards. 
Mathematical vector spaces are ontologically neutral. 
That is, the same formal devices can be used to model \enquote{energy spaces} consisting of force vectors. 
Following this idea, a domain extension is straightforward: in addition to spatial vector spaces, each individual $d \in D_e$ is assigned two force spaces, a pulling and an attracting one. 
Speakers occupy the respective \enquote{center of gravity} -- see \crefrange{fig:repulsion-space}{fig:attractor-space}.
Talmy's analyses of \textit{climbing} can be made precise in terms of an energy space spanned by the orthogonal projections of force vectors onto the downwards and upwards pulling ones in repulsion space.

\begin{figure}[htb]
\begin{floatrow}[2]
\ffigbox[\FBwidth]{%
    \begin{tikzpicture}
    \draw [gray] (-2,-1.5) rectangle (1.8,1.5);
      \fill (0,0,0) circle (2pt);
      \draw[vector] (0,0,0) -- (0,1,0) node [above]
      {};
      \draw[vector] (0,0,0) -- (0,-1,0) node [below]
      {}; 
      \draw[vector] (0,0,0) -- (1,0,0) node [right]
      {};
      \draw[vector] (0,0,0) -- (-1,0,0) node [left]
      {};
      \draw[vector] (0,0,0) -- (0,0,-1) node [above right]
      {};
      \draw[vector] (0,0,0) -- (0,0,1) node [below left]
      {};
    \end{tikzpicture}
}{%
\caption{Repulsion space}
\label{fig:repulsion-space}
}
\ffigbox[\FBwidth]{%
    \begin{tikzpicture}[every path/.append style={shorten <=4pt}]
    \draw [gray] (-2,-1.5) rectangle (1.8,1.5);
      \fill (0,0,0) circle (2pt);
      \draw[inversevector] (0,0,0) -- (0,1,0) node [above]
      {};
      \draw[inversevector] (0,0,0) -- (0,-1,0) node [below]
      {}; 
      \draw[inversevector] (0,0,0) -- (1,0,0) node [right]
      {};
      \draw[inversevector] (0,0,0) -- (-1,0,0) node [left]
      {};
      \draw[inversevector] (0,0,0) -- (0,0,-1) node [above right]
      {};
      \draw[inversevector] (0,0,0) -- (0,0,1) node [below left]
      {};
    \end{tikzpicture}
}{%
\caption{Attractor space}
\label{fig:attractor-space}
}
\end{floatrow}
\end{figure}

Force vectors are arguably involved in verbal construction like \textit{on the one hand \ldots\ on the other hand}: the two poles referred to are pulled apart by force vectors drawing in opposing directions.
Accordingly, this verbal construction is often accompanied by two placing gestures that locate the two poles in two different hemispheres of gesture space. 
Another point in this case could be gestures of uncertainty: being unsure corresponds to a lack of direction in energy space. 
Meandering through force vectors manifests itself in manual wiggling movements.
Of course, this is speculative to a great extent, but a line of theoretical research worth to be explored in future work.

\section{Informational Evaluation}
\label{sec:main-informational-evaluation}

This section is concerned with labeling approaches: \emph{how to derive predicates for describing the semantic contribution of gestures} -- what we call \emph{informational evaluation}.
%
Once a gesture is informationally evaluated, the multimodal utterance of which the gesture is a part can be construed by using this evaluation -- what we call \emph{conditioned interpretation} (because the interpretation of a multimodal utterance is conditioned on the informational evaluation of the gesture).
Obviously, informational evaluation is a \emph{heuristic act}, carried out by the interlocutors, or the working semanticist, as described in \cref{subsec:who-is-interpreting}. 
Accordingly, it needs a place in a semantic theory of iconic gestures. 
This is even more pressing because the \enquote{meaning} of gestures is often conflated with their informational evaluation in terms of linguistic labels \citep[see][]{Lascarides:Stone:2009:a,Schlenker:2019-gestural}.\footnote{Unfortunately, the use of labels is not innocent because it influences speaker judgments \citep[cf.][]{Hunter:2019}.} 
The main aim of this section is to develop such a semantic heuristic for gesture interpretation.

\subsection{The semantic challenge(s) of gesture}
\label{subsec:visuo-spatial}

One of the challenges of these gestures for semantic theories is that iconic gestures are not regimented by fixed form--meaning associations (i.e., a lexicon; cf. \citealt{McNeill:1992}).\footnote{It is noteworthy that the iconic gestures we are concerned with have been described as spontaneous hand and arm movements and called \enquote{gesticulations} by \citet{Kendon:1988}, in contrast to conventionalized gestures like emblems (e.g., \enquote*{thumbs-up}) or the signs of a sign language.} 
Hence, iconic gestures do not constitute a fixed class of items which can simply be interpreted by the interpretation function, \denotation{$\cdot$}, for language (or some related function for gestures). 
With regard to example (\ref{ex:spiral-treppen}) we already considered \textit{wounded} or \textit{spiral} as linguistic interpretations of the gesture, but \textit{circular}, \textit{upwards}, \textit{helical}, \textit{conchoidal}, \textit{twisted} would also do, as would \textit{slender}, \textit{tight}, \textit{narrow}, or \textit{ascending}, or even ignoring the gesture at all (we come back to this shortly).  
Hence, a single gesture gives rise to a plethora of possible informational evaluations, and the linguistic understanding of the whole multimodal utterance varies with each evaluation.
For instance, if the gesture is informationally evaluated 
to mean \textit{helical}, then the utterance is about helical staircases, if the evaluation amounts to \textit{tight}, then the utterance is about tight staircases, and so on.

It should be noted that example (\ref{ex:spiral-treppen}) is already exceptional.
A gesture occurrence is semantically related to a sub-utterance of the utterance it co-occurs with, its \emph{lexical affiliate} \citep{Schegloff:1984}.\footnote{The lexical affiliate is lexical, that is, a \enquote{single word}, in only about 80\% of occurrences, however, the remainder exhibit syntactically more complex verbal attachment sites \citep{Mehler:Luecking:2012-pathways}.}
%
%
%
In many, if not most, instances the gesture remains \enquote{informationally vacuous} in the sense that it just depicts its affiliate expression, as, for instance, shown in (\nextx) (SaGA dialogue V5, 4:44, the speaker talks about the round window in a church tower).
\ex 
{[\ldots]} mit ner Rosette \includegraphics[trim=100 330 500 160, clip, width=2cm]{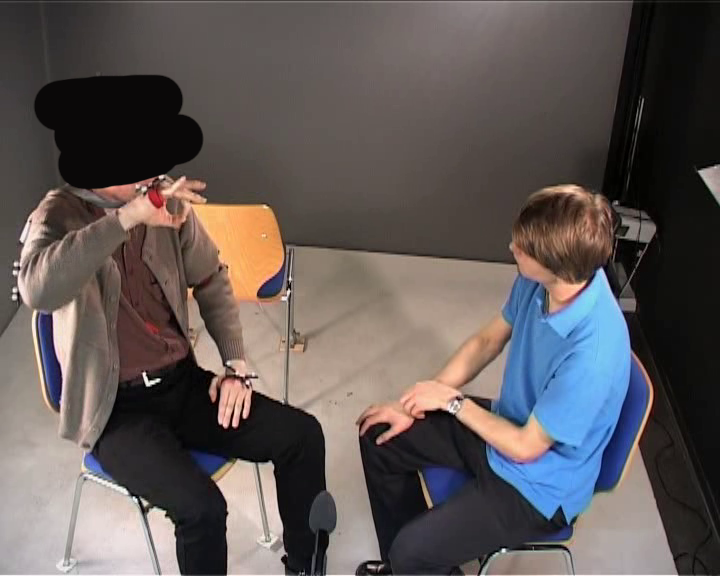} .. nen Rundfenster / \textit{with a rosette .. a round window}
\label{ex:affiliate}
\xe 

The distinguishing property of a round window is of course roundness, which is depicted by a ring gesture. 
Examples involving non-static gestures are given in (\nextx):
\ex 
Stewart Robson on \textit{ESPN FC Extra Time}:\footnote{Watch example at \url{https://www.youtube.com/watch?v=CiVS5_HKFY8&t=0h1m18s}.} 
\label{ex:wheel-dagger}
\par 
\textit{You know when they go on that wheel} \includegraphics[width=2cm]{gesture-wheel.png} \textit{and throw} \includegraphics[width=2cm]{gesture-throw-dagger}  \textit{the dagger would you ever like to see that go wrong}? 
\par 
$\Rightarrow$ the circle gesture represents the wheel, and the throwing movement mimics the dagger throwing action
\xe 

This direct matching between a gesture and its affiliate is the default case of speech--gesture integration and is violated by the staircase gesture in (\ref{ex:spiral-treppen}), where the gesture depicts a property not overtly expressed in speech.
We show in \cref{sec:lexicon-speech-gesture-integration} how to account for such indirect depictions in terms of lexicon-based conditioned interpretation.

Sensitivity to an affiliate is also the reason why one and the same gesture is interpreted very differently when produced in conjunction with different words and phrases, as exemplified in (\nextx), varying a constructed example used by \citet{Esipova:2019}.\footnote{Instead of a photo, the gesture here is drawn by using Ressler's \emph{Sketch} system (\url{http://www.frontiernet.net/~eugene.ressler/}).}
\pex \label{ex:s-brings}
\a S. might bring her \gestlarge\ \bringbox{dog.} $\Rightarrow$\enquote{holding/touching dog} \label{ex:large-dog}
\a S. might bring her \gestlarge\ \bringbox{dog in a box.} $\Rightarrow$ \enquote{holding/touching box}
\a S. might bring her \gestlarge\ \bringbox{dog by carrying it.} $\Rightarrow$ \enquote{carrying dog}
\xe 

The different verbal contexts of the gesture lead to the slightly different interpretations given to the right of the \enquote{$\Rightarrow$}.

The most striking feature of iconic gestures, however, is the fact that they make no linguistic contribution \emph{without someone evaluating them informationally}.
This can be seen from denial tests, as shown in \cref{sec:introduction}.
The linguistic antecedents are missing because the gesture has not asserted,  presupposed, or (conventionally or conversationally) implicated anything (asserting, presupposing, and implicating being linguistic acts).

The missing antecedents can be established, however, if the interlocutors come up with a shared informational evaluation of the gesture \citep{Luecking:Mehler:Henlein:2024-classifier}.
This can be shown using data from dialogue, where interlocutors A and B agree on the linguistic interpretation of the gesture (\nextx b--c), (i) respectively (ii), which can then be rejected (\nextx d) (or accepted).
\pex \label{ex:spiral-staircases-dialogue}
\a A: {[\ldots]} that should be \staircases staircases 
\a B: By \coil[0.3] [\textit{B repeats A's gesture}], do you mean that the staircases are (i) tight / (ii) wounded?
\a A: Yes.
\a B: But that's not true. The staircases are not (i) tight / (ii) wounded. 
\xe

\subsection{Who is interpreting?}
\label{subsec:who-is-interpreting}

Before turning to a semantic characterization of the challenge posed by iconic gesture, we should briefly discuss \emph{when} the informational evaluation of an iconic gesture is at stake at all, and who brings it about, if at all. 
In (\ref{ex:spiral-staircases-dialogue}), for instance, one interlocutor, B, is the interpreter of A's gesture. 
This is one of two possibilities known from gesture research, the other is the gesture researcher his/herself.
That is, there are three kinds of informational evaluation situations: (i) clarification interaction \citep{Ginzburg:Luecking:2021-clarifications}, (ii) gesture uptake \citep{Gullberg:Kita:2009}, and (iii) the verbal description of meaning and function of gestures within gesture studies \citep[e.g.,][]{Hadar:2013}.
%
%
We already discussed (i) (cf. \cref{sec:introduction}).
Let us briefly turn to the other two.

With regard to (ii) it is somewhat surprising that addressees do not always informationally evaluate the speaker's gestures.
This has been tested, for instance, by \citet{Gullberg:Kita:2009} in a drawing response study, where participants had to draw a situation that they saw described in a video of a speaker using speech and gesture.
The speaker's gesture included a target gesture, that is, a gesture that displayed information not verbalized in speech (e.g., the direction of a movement).
The authors found that the drawings only included the information exclusively gestured more often if the speaker gazed at the target gesture.
Hence, interlocutors themselves make a distinction between (mostly peripheral) \emph{seeing} a gesture and \emph{interpreting} a gesture (gesture uptake, or informational evaluation).
This is reminiscent of the twofoldedness of gestures pointed out in the introductory quote: seeing gestures (movements) and seeing something in gesture (informational evaluation).\footnote{This confirms, to our minds, the diagnosis of \citet[p.~91\psq]{Giorgolo:2010-phd}, namely that the \enquote{meaning} (i.e., visuo-spatial percept) of iconic gestures has to be kept apart from communicating with these gestures (informational evaluation).}



With regard to (iii), mainstream semantics use linguistic labels to stipulate gestural meanings, as exemplified in (\nextx), taken from \citep[751]{Schlenker:2019-gestural}:
\ex 
\adjincludegraphics[scale=0.3, valign=m]{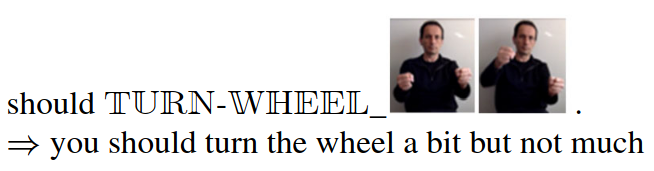} 
\xe 

In (\lastx) the presumed semantic contribution of the gesture is given by a verbal description, \textit{turn wheel} in this case.
The multimodal utterance is supposed to trigger the inference indicated after the double arrow. 
Note that the gesture in (\lastx) does not have an overt affiliate, it is a so-called pro-speech gesture, or \enquote{mixed syntax} \citep{Slama-Cazacu:1976}.
Furthermore, it is a gesture that is (close to) pantomime, a highly conventionalized form of gesture, see \cref{subsec:conventionalization}. 
We conjecture that it is no coincidence that the (constructed) examples of pro-speech gestures from the literature involve pantomime, because the increased degree of conventionalization explains observations about pro-speech gestures, which we return to in \cref{subsubsec:eval-labeling}.

\subsection{Exemplification and perceptual classification} 
\label{sec:perceptual-classification}

\Cref{subsec:reversed-denotation-exemplification} shows that informational evaluation rests on a \emph{direction} of meaning that is not part of possible world semantics. 
It is, however, related to notions of meaning developed in other areas of philosophy of language and computational semantics. 
It is therefore instructive to look at those areas (namely Goodman's exemplification and classifier-based semantics) and in which aspect they nonetheless fail to capture iconic gestures (namely by conflating gestures as event simulations with real-world events).
Section~\Cref{subsec:vector-space-semantics} introduces a visuo-spatial extension of (some) lexical entries.
This move allows to disentangle gestures miming events from real-world events: these visuo-spatially representations capture what gestures and the event they simulate have in common (how they \enquote{look alike}).
%


\subsubsection{Reversed denotation and exemplification}
\label{subsec:reversed-denotation-exemplification}

\begin{figure}[htb]
\centering 
        \begin{tikzpicture}
        \node[inner sep=0pt] (img) {\includegraphics[width=5cm]{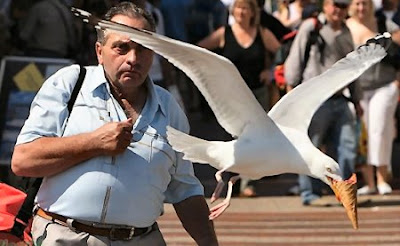}};
        \end{tikzpicture}
        \caption{Situation $s$: \enquote{Bird Stealing Icecream} (Gerard Vlemmings, 
        {CC BY-NC-ND 3.0}, 
        \protect\url{https://presurfer.blogspot.com/2009/03/birds-stealing-ice-cream.html}}
        \label{fig:bird-stealing-icecream}
\end{figure}

A gesture is an instance of visual communication.
Visual communication is related to the problem of \textit{how it is possible to talk about what one sees} 
\citep[p.~90]{Jackendoff:1987}.
Consider \cref{fig:bird-stealing-icecream} and its description in (\nextx a). 
The standard truth-conditional semantics of the assertion is given in (\nextx b) in terms of a property $m$ of situations (ignoring tense and definiteness information).
The assertion is true of the event $s$ depicted in \cref{fig:bird-stealing-icecream} iff (abbreviates \textit{if and only if}) $s \in m$.\footnote{Assuming a (possibilistic) situation semantics like the system of \citet{Kratzer:1989}.}
\pex 
\a \enquote{The bird is stealing icecream.}
\a $m=\lambda e [\text{steal}(e,x,y) \wedge \text{bird}(x) \wedge \text{icecream}(y)]$
\xe 

The utterance in (\lastx) involves one-place predicates (\textit{bird}, \textit{icecream}), whose meanings according to standard possible worlds semantics are (modelled as) functions from possible worlds (or world--time pairs, or situations) to entities.
Thus, predicates on this approach exhibit a \textit{word-to-world} direction of fit.\footnote{The direction-of-fit difference has been put forth by speech act theory \citep{Searle:Vanderveken:1985-illocutionary-logic}.}
Speaking about what one sees, however, involves a relation $f$ that exhibits a \textit{world-to-word} direction of fit: given a perceptual input $\alpha$, $f(\alpha)$ returns linguistic labels that classify $\alpha$.
Speaking about the scene depicted in \cref{fig:bird-stealing-icecream}, for instance, rests on two object classifications (in addition to a stealing event classification), which can be symbolized as follows:
\pex
\a $f(\text{\faIceCream})=$ icecream
\a $f(\text{\faCrow})=$ bird (seagull)
\label{ex:f}
\xe 

As indicated in (\lastx b), there are usually several possibilities but from a common taxonomy to classify an object, with a preferred or default level \citep{Rosch:et:al:1976-basic-objects}.
We will, however, largely ignore issues of natural categories in the following.
There are nevertheless two important differences between $f$ and $\llbracket\cdot\rrbracket$: 
\begin{itemize}
\item $f$ is a  \emph{relation}, not a unique function as $\llbracket\cdot\rrbracket$ is supposed to be; thus $f$ usually gives rise to several labels. 
\item $f$ and $\llbracket\cdot\rrbracket$ have contrary input--output domains (this is an alternative way of saying that they have antagonistic world--word directionality) and are therefore \emph{incompatible}. 
\end{itemize}

Hence, $f$ cannot be reduced to $\llbracket\cdot\rrbracket$.
But given a model for $\llbracket\cdot\rrbracket$, then $f$ can be thought of as an inverse relation over $\llbracket\cdot\rrbracket$. 
%
Philosophy of language knows a candidate for such a relation $f$, namely \emph{exemplification}, a converse denotation relation due to  \citet{Goodman:1976}.
As a simple example,\footnote{Goodman had no model-theoretic semantics at his disposal. Given his nominalist stance, much of his discussion pertains to inscriptions. We ignore the expressive dimension and metaphoric exemplification altogether.} consider the toy model in (\nextx a).
The denotation of \textit{green} is the set of three green objects.
Given this, any object within the denotation can be used to \emph{exemplify}, \exemplifies, \textit{green}, for instance, the green circle as in (\nextx b).
\pex 
\a $\denotation{green} = \{\tikz\node[rectangle, fill=green!80!black, minimum size=0.35cm]{}; , \tikz\node[circle,fill=green!70!black,minimum size=0.35cm]{}; , \tikz\node[isosceles triangle,minimum size=0.3cm, fill=green!75!black]{};\}$
\a \tikz\node[circle,fill=green!70!black,minimum size=0.35cm]{}; \exemplifies green
\xe 

Conversely, given this model, we can say that the predicate \textit{green} can be used to \emph{label} or \emph{classify} the green circle, since the green circle is part of \textit{green}'s denotation. 
The exemplification of transitive and $n$-place predicates in general is brought about by collections of $n$ objects.

While this seems to be on the right track, exemplification \textit{simpliciter} falls short of capturing iconic gestures, because gestures are not events but \emph{simulate} events \citep{Hostetter:Alibali:2008}. 
Consider again the throwing gesture from (\ref{ex:wheel-dagger}), re-given in (\nextx):
\ex 
\textit{You know when they go on that wheel and throw} \includegraphics[width=2cm]{gesture-throw-dagger} \textit{the dagger would you ever like to see that go wrong}?
\label{ex:throw-dagger2}
\xe 

The speaker in (\lastx) is not actually throwing something. 
While he mimes handshape and movement of a throwing event, no dagger is leaving his hand.
Given that the extension of the verb phrase \textit{throwing a dagger} are triples consisting of an event, an agent and a theme (a dagger), exemplification fails to deliver \textit{throw} as a label for the gesture since the theme is missing.
Trying to apply exemplification straightforwardly conflates gestures simulating actions with real-world actions. 
What is needed in this case is a means to disentangle throwing events from thrown objects in order to capture the simulative nature of acting gestures.\footnote{Gesture studies speak of \enquote{ad hoc abstraction} here and provide an interpretation drawing on metonymy \citep[p.~1747]{Mittelberg:Waugh:2014}.}
%

\subsubsection{The lexical meaning of motion and path verbs}
\label{subsec:lexical-meaning-of-motion-and-path}
 
Using geometric vector representations as representations for meanings is not new in natural language semantics and the philosophy of language
\citep[e.g.,][]{Warglien:Gaerdenfors:Westera:2012-events,Zwarts:Gaerdenfors:2016}.\footnote{\citet{Chalmers:2002-sense} makes the case for \emph{epistemic intensions}, but does not provide a formal (e.g., vector based) model.}
In cognitive science, it is commonplace that lexical items have both symbolic and visual meaning components \citep{Paivio:1986}.
A related dual coding approach has already been developed for the synthesis of three-dimensional iconic gestures in terms of \emph{imagistic description trees} \citep{Sowa:2006:a}.
Computational linguistics developed \enquote{words-as-classifiers} approaches \citep{Kennington:Schlangen:2015,Larsson:2015}, where the meaning of perception-related words is perceptually grounded \citep{Harnard:1990,Steels:Belpaeme:2005,Roy:2005}.
%
%
\citet{Bartsch:1998} developed a comparable approach in semantics and philosophy of language. 

%
Let us consider the example of biological motion and motion verbs.
Motion verbs vary along two dimensions: manner and path \citep{Engelberg:2000}.
The eigenmovement distinguishes motion verbs according to manner, regardless of the distance travelled: 
\ex
    \begin{tikzpicture}[baseline=(m-1-1)]
    \matrix[matrix of nodes,
    left delimiter=\{,
    right delimiter=\}] (m) {
    run \\ walk \\ stroll \\ saunter \\ \ldots \\
    };
    \end{tikzpicture} 
\xe 

Translational movement gives rise to a path that distinguishes motion verbs irrespective of the manner of motion:  
\ex
    \begin{tikzpicture}[baseline=(m-1-1)]
    \matrix[matrix of nodes,
    left delimiter=\{,
    right delimiter=\}] (m) {
    run \\ detour \\ circle \\ criss-cross \\ \ldots \\
    };
    \end{tikzpicture}
\xe

\begin{figure}[htb]
\subfloat[Stimulus for \textit{walking}\label{fig:walking-stimulus}]{%
\includegraphics[width=2cm, trim=0 360 0 0, clip]{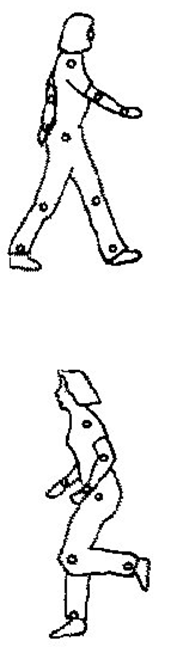}}
\hfill
\subfloat[Carriers of \textit{walking} movement\label{fig:walking-carrier}]{%
\includegraphics[width=2cm, trim=0 180 0 0, clip]{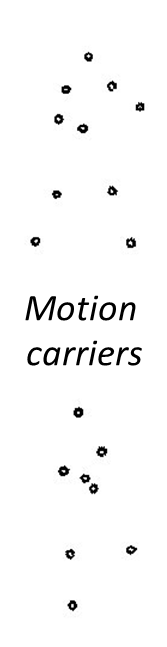}}
\hfill
\subfloat[Model of carrier movement\label{fig:carrier-movement}]{%
\begin{adjustbox}{scale=0.85}
           \begin{tikzpicture}[
        Carrier/.style={circle, inner sep=0pt, minimum size=7pt},
        Hip/.style={Carrier, fill=GU-Gruen},
        Knee/.style={Carrier, fill=GU-Goethe-Blau},
        Ankle/.style={Carrier, fill=GU-Orange},
        ]
        \begin{scope}[xshift=-2cm, scale=0.4, font=\scriptsize]
          \node [Hip, label=above:$A_1$] (A1) at (1,2) {};
          \node [Hip, label=above:$A_2$] (A2) at (2.5,2) {};
          \node [Hip, label=above:$A_3$] (A3) at (4,2) {};
          \node [Hip, label=above:$A_4$] (A4) at (5,2) {};
          \node [Hip, label=above:$A_1$] (A5) at (7,2) {};
          
          \node [Knee, label=above left:$B_1$] (B1) at (0,0) {};
          \node [Knee, label=above right:$B_2$] (B2) at (2.5,0) {};
          \node [Knee, label=above right:$B_3$] (B3) at (4.5,0) {};
          \node [Knee, label=above right:$B_4$] (B4) at (5.5,0) {};
          \node [Knee, label=above right:$B_1$] (B5) at (6.25,0) {};
          
          \node [Ankle, label=below:$C_1$] (C1) at (-1,-2) {};
          \node [Ankle, label=below:$C_2$] (C2) at (1,-1) {};
          \node [Ankle, label=below:$C_3$] (C3) at (5.5,-2) {};
          
          \draw (A1) to (B1) to (C1);
          \draw (A2) to (B2) to (C2);
          \draw (A3) to (B3) to (C3);
          \draw (A4) to (B4) to (C3);
          \draw (A5) to (B5) to (C3);
          \draw [dashed] (A1) to [bend right=20] (A2);
          \draw [dashed, decoration={snake, segment length=1cm, amplitude=0.05cm}, decorate] (A2) to (A3);
          \draw [dashed] (A3) to [bend right=15] (A4);
          \draw [dashed, decoration={snake, segment length=1.5cm, amplitude=0.05cm}, decorate] (A4) to (A5);
          \draw [dashed] (C1) to [bend left] (C2);
          \draw [dashed, decoration={snake, segment length=3cm, amplitude=0.15cm}, decorate] (C2) to (C3);
          
          \matrix [right=0.5cm of B5, nodes={anchor=west}, row sep=0.5cm] {
            \node {hip}; \\
            \node {knee}; \\
            \node {ankle}; \\
          };
        \end{scope}
      \end{tikzpicture}
      \end{adjustbox}
  }\hfill
  \subfloat[Abstract vector model of \textit{walking}\label{fig:walking-vector}]{%
  \begin{adjustbox}{scale=0.85}
            \begin{tikzpicture}[
        Carrier/.style={circle, inner sep=0pt, minimum size=7pt},
        Hip/.style={Carrier, fill=GU-Gruen},
        Knee/.style={Carrier, fill=GU-Goethe-Blau},
        Ankle/.style={Carrier, fill=GU-Orange},
        ]
        
        \begin{scope}[xshift=4cm, scale=0.4, font=\scriptsize]
          \node [Hip, label=above:$A$] (A) at (1,2) {};
          \node [Knee, label=above left:$B_1$] (B1) at (0,0) {};
          \node [Knee, label=above right:$B_4$] (B4) at (2,0) {};
          
          \draw (A) to (B1);
          \draw [dashed] (A) to (B4);
          \draw [->] (B1) to [bend right] (B4);
          
          \node [circle, draw=GU-Dunkelgrau, fit=(A) (B1) (B4), inner sep=1ex] (Move) {};
          
          \draw [->] (Move.east) -- ++(0:2cm) node [midway, above] {$A_1 \rightarrow A_4$};
        \end{scope}
      \end{tikzpicture}
      \end{adjustbox}
  }
\caption{Motion perception: \textit{walking}. (Taken from, respectively reproduced after \protect\citealp[p.~202 and 208]{Johansson:1973})}
\label{fig:walking}
\end{figure}

The path component -- the translational dimension of motions -- is already covered by the vector denotations within the spatial model from \cref{sec:main-spatial-gesture-semantics}, which truth-conditionally distinguish the verbs in (\lastx). 
But what about the manner dimension?
In an important series of experimental studies on the perception of biological motion, \citet{Johansson:1973,Johansson:1976,Johansson:et:al:1980} investigated, among others, the difference between \textit{walking} and \textit{running}.
How are we able to consistently tell both motion manners apart across different human (and presumably some non-human) individuals?
Is there an abstract perceptual commonality between running events on the one hand and walking events on the other hand?
Johansson and colleagues placed little lights at the anatomical joints that actually bring about the movement (the so-called \emph{motion carriers}) -- see \cref{fig:walking-stimulus}. 
The recorded light pattern (\cref{fig:walking-carrier}) is then shown to participants, who \enquote{saw} and correctly classified a walking event.
\enquote{How can 10 points moving simultaneously on a screen in a rather irregular way give such a vivid and definite impression of human walking?} \citep[p.~204]{Johansson:1973}.
An answer to this question was found in geometric analyses of the temporal stimulus pattern.
Walking is characterized by two horizontal trajectories (due to hip and knee carriers) and an up-and-down sequence (ankle) -- see
\cref{fig:carrier-movement}.
Factoring out common movement shares, the kernel percept of a walking event is the \emph{abstract vector model} shown in \cref{fig:walking-vector}.
If we observe something that looks like this vector model, we can classify it as \textit{walking}.

Now singling out walking events is exactly what the meaning of the verb \textit{walk} is supposed to achieve, and what is \enquote{pre-compiled} in model-theoretic semantics.
%
Accordingly, the model in \cref{fig:walking-vector} provides a representation of (a part of) the intensional meaning of \textit{walk}. 
We, following \citet{Luecking:2013:a}, refer to this visuo-spatial representation of an intension as \emph{conceptual vector meaning}, or \cvm for short.
Arguably, the lexical entry of any visuo-spatial expression comes with a \cvm (cf. \emph{dual coding}; \citealp{Paivio:1986}).\footnote{There is some related work in the framework of conceptual semantics, which, among others, addresses the problem of how we are able to talk about what we see \citep{Jackendoff:1987}. On this approach, a theory of language is connected to a theory of vision, namely the 3-D models of \citet{Marr:2010-vision}. Marr's model is specialized for object recognition and identification, vector models seem to be better suited for capturing biological movement \citep[but see][]{Marr:Vaina:1982}. A semantic rendering of Marr's model also underlies the logic of vision of \citet{van_der_Does:van_Lambalgen:2000}, but it remains purely extensional.}
As an illustration, consider (\nextx).
The meaning of the lexical entry in (\nextx a) is replaced by one involving a \textsc{cvm}, as illustrated in (\nextx b); the place path is due to the spatially extended model and distinguishes, for example, direct walks from detours.
\pex \label{ex:from-walk-to-walk-cvm}
\a 
$\denotation{walk} = \lambda x. \lambda e [\text{walk}(e) \wedge \text{agent}(e,x) \wedge \exists \bvec{v} [\text{place-path}(e,\bvec{v})]]$ 
\a 
$\denotation{walk} = \lambda x. \lambda e [\text{walk-\cvm}(e) = 1 \wedge \text{agent}(e,x) \wedge \exists \bvec{v} [\text{place-path}(e,\bvec{v})]]$ 
\xe 

(\lastx a) is standard (given a spatially extended vector model); 
%
(\lastx b) adds that the set of events $E$ characterized by the embedded function is such that each event $e \in E$ \enquote{looks like} the vector model encoded in {walk-\cvm}.\footnote{Hence, we relate semantics to non-linguistic cognitive activities such as perception, but contra \citet[p.~19]{Lewis:1970}, we do not think that \enquote{confusion comes of mixing these two topics}.}
That is, \cvm acts like a \emph{perceptual classifier} known from computational, classifier-based semantics \citep{Kennington:Schlangen:2015,Larsson:2015}.
A perceptual classifier maps perceptual input (from an object or a situation) to the interval $[0,1]$.\footnote{The \cvm-classifier in (\ref{ex:from-walk-to-walk-cvm}) is, however, a binary one since it just distinguishes successful (\enquote{1}) from unsuccessful (\enquote{0}) classifications. We make use of this simplification for heuristic purposes, but note that a serious computational classifier should be probabilistic and continuous.} 
For words $w$ and (perceptual data obtained from) objects or situations $x$, a classifier returns scores 
\ex 
$\sigma_w(x) \mapsto [0,1]$
\xe 

Following ideas in logic and formal philosophy of language where intensions are treated as algorithms for determining extensions \citep{Moschovakis:1993,Muskens:2005}, classifiers are employed as intensions of words, namely intensions that are independent of extensions.
The adaptation of intensional word meanings in the framework of \citet{Montague:1974} to classifiers can be done in the following, straightforward way (cf. \citealp{Kennington:Schlangen:2015}; \citealp{Larsson:2015} uses a different framework from the outset): 
\ex 
$\llbracket w \rrbracket = \lambda x . \sigma_w(x)$
\xe 

Hence, computational semantics provides a procedure for implementing the exemplification relation. 
Moreover, since the \cvm in (\ref{ex:from-walk-to-walk-cvm}) applies to motion patterns bound up with events and is independent of the agent of the event, it applies to iconic gestures, too: the extended lexical meaning of \textit{walk} including the abstract vector model shown in \cref{fig:walking-vector} can be used to interpret a gestural movement $\gamma$ that looks like walking as walking, as exemplified in (\nextx).
\ex \label{ex:extemplify-walk}
\begin{tikzpicture}[baseline=(x)]
            \node (walk) {\includegraphics[width=2cm]{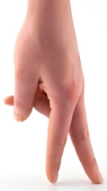}};

\draw [white, very thick] ([xshift=0.4cm]walk.center) -- +(255:1.7cm);
\draw [white, very thick] ([xshift=0.4cm]walk.center) -- +(285:1.7cm);
\draw [black!60, thick, dashed, <->] ([xshift=-0.3cm, yshift=0.1cm]walk.south east) arc[start angle=330, end angle=220, radius=0.5cm];
            
            \draw [thick, -stealth] ([yshift=0.2cm,xshift=-0.1cm]walk.south east) -- +(0:2cm);
            \node (x) at (1.5,1) {$x$};
            \draw [dashed, -latex] (x) -- +(180:0.9cm);
            \node (e) at (2.2,0) {$\text{walk-\cvm}(\gamma)=1$};
            \draw [dashed, -latex] (e) -- +(220:1.4cm);
            \node (v) at (2.1,-1) {\bvec{v}};
            \draw [dashed, -latex] (v) -- +(270:0.7cm);
\end{tikzpicture}
\xe
   
The gesture in (\lastx) is compatible with \textit{walk}, but fails for, e.g., \textit{stagger}, \textit{crawl}, \textit{give}, \textit{ride}, etc. because of different, incompatible {\cvm}s.\footnote{Of course, distinguishing gestures from real-world events follows from a double classification of the gesture according to a perceptual \cvm and a hand classifier (a hand-\cvm returns \textit{false} for real-world events).}

\subsubsection{Kind-denoting gestures}
\label{subsubsec:kind-gestures}

\citet{Fricke:2012} observed that some gestures do not depict their referent, but rather a prototype or concept associated with the referent. 
For example, when talking about a roof the gesturing person might produce a triangular iconic model (indicating a pointed roof, which is probably the most typical type of roof), even though the building she is actually talking about might happen to have a flat roof. 
We refer to such gestures as kind-denoting gestures. 
Kind-denoting gestures lack a truthful embedding in the affiliate's vector space. 
However, kind-denoting gestures are characterized by the fact that there is a another gesture with a different iconic model (\textit{modulo} rotation and scaling) that has a truthful embedding and that gives rise to to the same informational evaluation as the kind gesture. 
Thus, iconic gesture semantics provides a straightforward explanation of kind gestures without the need to establish additional mechanisms. 
%
%
An additional mechanism is employed by  \citet{Ebert:Hinterwimmer:2022,Walter:Ebert:Hinterwimmer:2025}, who suggest to explain kind-denoting gestures in terms of a the referent of a gesture (a rigid designator) which display prototypical aspects of the referent from the affiliate. 
Token-denoting gestures additionally enforce a similarity predicate which compares the referent of the gesture to the affiliate's referent. 
This is a problematic proposal for at least two reasons: Firstly, it has long been argued in philosophy that similarity is an insufficient concept for semantics, although in folklore it is assumed that iconic gestures are determined by similarity \citep[for a summary see][\S3]{Luecking:2013:a}.
Accordingly, the similarity predicate of \citet{Walter:Ebert:Hinterwimmer:2025} is rather vague, so that its explanatory value must be questioned.  
Secondly, the idea that gestures have a referent on their own has quite convincingly be refuted in semantic literature \citep[e.g.,][]{Lawler:Hahn:Rieser:2017}.

\subsection{Proof of concept: AI and gesture detection}
\label{subsec:ai}

Iconic hand gestures visually represent the spatial or physical properties of the referent, with the iconic model of the gesture reflecting the spatial aspects of the lexical entry of the affiliate.
%
If this is an accurate concept, it can be implemented using a gesture classifier.
A corresponding proof-of-concept model can be found at \url{https://github.com/texttechnologylab/SimpleActionGestureModel}.
The general approach is described in the following.

\subsubsection{Data Collection \& Preprocessing}

\begin{figure}
    \centering
    \includegraphics[trim=0 50 0 50, clip, width=0.35\linewidth]{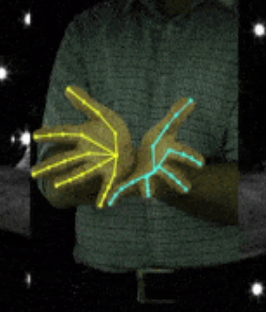}
    \caption{Vector-based hand recognition \protect\citep[figure taken from the InterHand2 dataset, \protect{https://mks0601.github.io/InterHand2.6M/},][]{Moon:et:al:2020-interhand2}}
    \label{fig:interhand}
\end{figure}

In order to build a gesture classifier, one must first gather a diverse dataset of target gestures. 
This can be done through video recordings or motion-capture data -- we use the \textit{Moments in Time} dataset (\url{http://moments.csail.mit.edu/}). 
It is important to ensure variability in performers, backgrounds, and conditions to ensure robustness \citep[Ch. 5]{Goodfellow:et:al:2016}.
Raw data often requires preprocessing, such as segmenting continuous recordings to isolate gesture events or normalizing and scaling coordinate data. 
To reduce data complexity, it is common to extract informative features -- for instance, pose estimation (like MediaPipe Holistic \citep{Lugaresi:et:al:2019} or OpenPose \citep{Cao:et:al:2017}) can be used to obtain the 3D coordinates of the hand and body's keypoints for each video frame.
This skeletal representation preserves the essential motion and shape of the gesture while filtering out background noise -- see \cref{fig:interhand} for an example. 
Representing gestures as sequences of these features (or derived images) enables the model to capture the spatiotemporal patterns that characterize each gesture.
%
%
%

\subsubsection{Self-supervised Pretraining}

Due to the limited size of many gesture datasets, self-supervised learning (SSL) techniques can be used to create robust feature representations prior to fine-tuning with labels. 
At this stage, a model (often a deep neural network) is trained using unlabeled gesture data and proxy tasks, such as predicting future motion, reconstructing masked parts of a sequence, or distinguishing between segments of the same and different gestures.
This pretraining step uses large amounts of unsupervised data, which is easier to collect, to give the model an understanding of generic \enquote{gestural} features and dynamics.
The benefit is that the model begins with a high-level understanding of hand motion patterns, making it less susceptible to overfitting when given actual gesture labels later on .
For example, a study showed that, after self-supervised pretraining on abundant unlabeled motion data, a classifier could recognize fine-grained hand gestures with over 95\% accuracy using only 5--10 labeled examples per class \citep{Kimura:2022}.
Similarly, in the closely related field of sign language recognition, a self-supervised framework that learns from pose sequences using masked motion modeling and hand-shape priors achieved state-of-the-art results despite limited labeled data \citep{Hu:et:al:2023}.
Incorporating a pretraining phase, especially with modern SSL methods, helps the model abstract salient movement features and spatial configurations. This effectively initializes the model to better recognize iconic gestures with minimal labeled data.

\subsubsection{Post-Training Fine-tuning}

During the fine-tuning stage, the network's parameters are adjusted using a curated, labeled dataset of gestures. 
Typically, either the pretrained backbone is frozen or updated at a lower learning rate while a final classification layer is trained to map the learned features to specific gesture labels \citep{Hu:et:al:2023}.
%
%
%
During this training phase, it is important to use data augmentation (e.g., mirroring videos, adding noise, and random temporal cropping) and regularization to combat overfitting, especially given the often small size of gesture datasets \citep{Habib:et:al:2025}.
%
%
%
%
The result of this stage is a model that can classify specific iconic gestures, linking learned motion and pose patterns to meaningful categories (\textit{drinking} and \textit{eating}, in our sample implementation).

\subsubsection{Evaluation}

Evaluation typically involves testing the model on a separate set of gesture data that was not used during training and measuring metrics such as classification accuracy, precision, recall, and F1-score for each gesture class.
While high overall accuracy indicates that the model recognizes gestures correctly, it is also crucial to ensure that it works well across different users and contexts due to variations in how people perform gestures. 
%
%
%
If the gestures are to be used in a real-time system, one must not only evaluate accuracy, but also latency (response time) and throughput (frames per second) to ensure that the model meets real-time requirements \citep{Cao:et:al:2019}.

\subsubsection{Deployment \& Iteration}

In the final stage, the trained model is deployed in a real-world application or experimental setup.
This may involve embedding the model in a system that receives live inputs, such as a webcam feed or sensors from an AR/VR device, and outputs gesture predictions to perform certain functions, such as interpreting user commands or analyzing behavior.
After deployment, the work is not done. 
An AI gesture system typically undergoes iterative improvement.
Users may provide feedback, or new edge cases may be discovered, prompting the collection of additional training examples and the refinement of the model \citep{Pouw:et:al:2024}.
This kind of ongoing iteration involves monitoring the system’s accuracy, fixing misclassifications or biases by incorporating new data, and possibly expanding the gesture set. 
An iterative deployment strategy ultimately ensures that the gesture classifier remains accurate and robust as it is exposed to more diverse usage while preserving real-time, spatial-semantic understanding of iconic hand gestures in practical scenarios \citep{Pouw:et:al:2024}.
Detailed reviews of existing gesture recognition approaches can be found in \citet{Sarowar:et:al:2025} and \citet{Tripathi:Verma:2024}.



In summary, computer vision and multimodal AI provide a computational implementation of the \cvm-based perceptual classifier from \cref{subsec:lexical-meaning-of-motion-and-path}, which was introduced primarily for empirical and theoretical reasons. 
Furthermore, pose detection algorithms can be used as continuous  computationally derived representations for gestures that would otherwise need to be encoded in the form of symbolic, feature-based annotations (see, however, \cref{subsec:kinematic-gesture-representation} for a discussion on continuity).

\subsection{Semantic toolkit: Extemplification heuristic and dynamic speech--gesture interaction}
\label{sec:toolkit}

We have argued in some detail that perceptual classification is key to understanding the informational evaluation of iconic gestures.
Classification of gestures, however, cannot be spelled out in terms of Goodmanian exemplification (which ignores the gestures' semiotic dimension). 
Extended exemplification on the other hand requires a dynamic, procedural approach to perceptual classification.
This in turn is incompatible with a semantic theory that rests on fixed, pre-given extensions like standard possible world semantics. 
We therefore propose an \emph{extended exemplification heuristic} for gesture interpretation.
That there is need for such a heuristic in semantic research is evinced in \cref{subsec:applications}, where we analyze concrete examples. 


\subsubsection{Informational evaluation}
\label{subsec:informational-evaluation}

The basic idea of informational evaluation has already been illustrated in relation to example (\ref{ex:extemplify-walk}) and the predicate \textit{walk}. 
Now consider \textit{throw}, as in the throwing-a-dagger example (\ref{ex:wheel-dagger}) in \cref{subsec:visuo-spatial}.
The lexicalized meaning of \textit{throw} is given in (\nextx).
\ex 
$\denotation{throw} = \lambda y . \lambda x . \lambda e [\text{throw-\cvm}(e)=1 \wedge \text{agent}(e,x) \wedge \text{theme}(e,y) \wedge \exists\bvec{v}[\placepath(y,\bvec{v})]]$
\label{ex:lex-throw}
\xe 

(\lastx) gives rise to the following Goodmanian exemplification (without a second \enquote{t}) conditions: if there is a body movement which looks like throwing (\enquote*{$\text{throw-\cvm}(e)=1$}), performed by $x$, and if there is something acted upon (\enquote*{theme($y$)}) and that something is dislocated (\enquote*{$\placepath(y,\bvec{v})$}; we abstract over time), we can classify this event $e$ as a throwing event.
Hence, a real-world event $e$ exemplifies a predicate constant $P$ if the event provides a witness for each of $P$'s arguments, and only for the arguments. 
We call this \emph{minimal exemplification}.
%
Any extended event $e'$ which includes $e$ will also exemplify that predicate, but not minimally.\footnote{Being confined to minimal exemplification exempts from considerations of situational upwards persistence \citep{Cooper:1991-persistence}.}
Note that more specific predicates can be exemplified, such as \textit{being thrown by Mike}, in which case Mike needs to be the agent of $e$, and so on.

\begin{figure}
    \centering
    \begin{tikzpicture}
    \node[](gest){\includegraphics[scale=0.3]{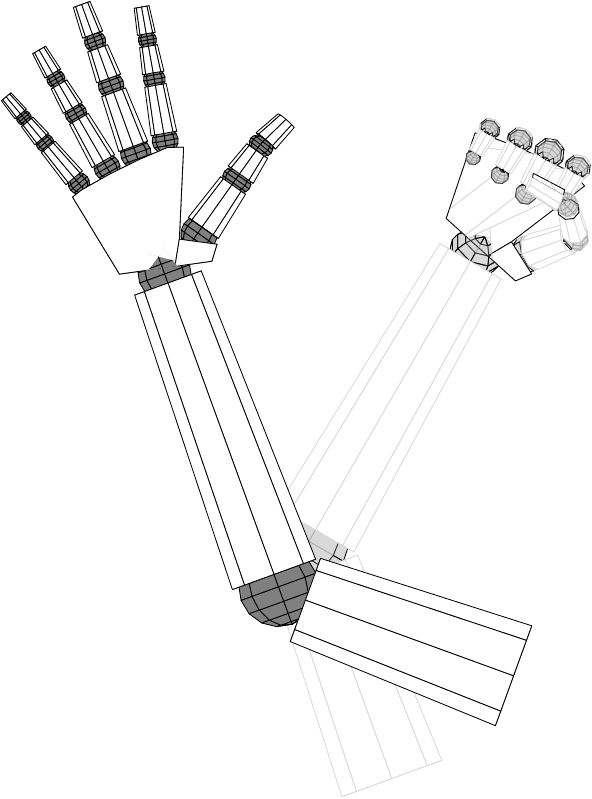}};
    \draw [dashed, very thick, gray, ->, shorten >=1.1cm, shorten <=0.6cm] ([yshift=1cm]gest.east) -- ([yshift=1.7cm]gest.west) node [pos=0.4] (cvm-anchor) {};
    \node [above=of cvm-anchor] (cvm) {throw-\cvm$(\gamma)=1$};
    \draw [dotted, ->] (cvm) -- (cvm-anchor);
    \draw [dotted, ->] (cvm) -- +(215:1.1cm);
    \draw [dotted, ->] (cvm) -- +(295:1.2cm);
    \node [right=of gest.south east, yshift=0.5cm] (agent) {agent $=$ speaker} edge [->, dotted] (gest.320);
    \draw [dashed, ->] ([yshift=-0.7cm]gest.north west) -- +(180:1.3cm) node [at end, left, draw] (theme) {y} node [midway] (v-anchor) {};
    \node [below left=0.6cm and 0.5cm of theme, align=center] {virtual\\ theme} edge [dotted, ->, shorten >=6pt] (theme);
    \node [below=of v-anchor, align=center] {virtual \\ trajectory \bvec{v}} edge [dotted, ->] (v-anchor);
    \end{tikzpicture}
    \caption{Exemplifying \textit{throw}.}
    \label{fig:ex-throw}
\end{figure}

Explicit representations of intensions in terms of \textsc{cvm}s provide the conceptual tools for expressing the required abstraction to move from exemplification to \emph{extended exemplification}, \extemplifies, or \emph{extemplification} (with a second \enquote{t}) as a short coinage.\footnote{This is a significant improvement over \citet{Luecking:2013:a}, who uses Goodmanian exemplification, inheriting the above-discussed problems.} 
The difference between exemplification and extemplification is that the latter acknowledges presupposed situational arguments. 
This is exemplified in \cref{fig:ex-throw}: the theme of the gestural throwing event simulation remains virtual, or presupposed, but this visually presupposed entity is allowed to fill a role within the schema of minimal exemplification, amounting to extended exemplification. 
Schematically: 
\pex \emph{Extended exemplification as informational evaluation of a gesture} \label{ex:exemplification-schema}
\a A gesture $\gamma$ extemplifies a predicate $p$, $\gamma$ \extemplifies $p$, if $\text{$p$-\cvm}(\gamma)=1$ and $\gamma$ is  minimal wrt. $p$.
\a $\gamma$ is  minimal wrt. $p$ iff there is a bijective mapping between (i) form features of $\gamma$, or (ii) visual, presupposed features of $\gamma$ and the arguments of $p$. 
\a If a. and b., that is, if $\gamma$ exemplifies $p$, we can use $p$ to informationally evaluate $\gamma$. 
\xe 

Steps (\lastx a,b) are to be brought about by the working semanticist, unless a computational classifier system is available (see \cref{subsec:vector-space-semantics}).
%
This is part of what makes (\lastx) a heuristic.\footnote{In human vision, perceptual classification is brought about by aligning a perceptual image with stored visual models \citep[in particular chapters 6 and 7]{Ullman:1996-vision}. See also the AI model in \cref{subsec:ai}.}
Step (\lastx c) delivers the input for the information-evaluation conditioned interpretation of multimodal utterances mentioned in \cref{subsec:visuo-spatial} and expressed explicitly in template (\ref{ex:gesture-conditioned-affiliation}) below.

As has been emphasized in \cref{subsec:visuo-spatial}, a gesture extemplifies several predicates. 
Now the working semanticist does not need to go through all possible extemplifiees.  
It is sufficient to use the affiliate for the informational evaluation of the gesture.
Let us look at an example of how (\lastx) works and offers a toolkit for linguistic interpretation within semantic gesture studies, namely the throwing gesture from example (\ref{ex:wheel-dagger}) in \cref{subsec:visuo-spatial}.
The relevant extract is given in (\nextx):
\ex
{[\ldots]} throw [\textit{throwing movement}] the dagger
\label{ex:throw-dagger-exemplification}
\xe 

Let \cref{fig:ex-throw} be an illustration of the gesture involved in (\lastx) -- it extemplifies \textit{throw} via the lexical meaning in (\ref{ex:lex-throw}). 
The movement \enquote{looks like} throwing (captured by the \cvm), it is a pantomimic action (the speaker mimics the agent of the simulated action), and the imagined continuation of the stopped gestural movement evokes a virtual trajectory which triggers a virtual theme.  
Both the virtual trajectory and the theme are presupposed by evaluating the gesture as throwing.
If this presupposition is not fulfilled, the gesture cannot \enquote{mean} throwing (it could extemplify some intransitive predicate instead, for instance, some direction instruction of a flight attendant, \enquote{Exit on the right}). 
Extended exemplification -- in addition to a gesture's preference to directly depict its affiliate --, thus, provides some justification to interpret the gesture in (\lastx) as \textit{throw}. 
%

There is another interpretation available, namely the VP figuring as affiliate, that is $\lambda x. \lambda e [\text{throw-\cvm}(e)=1 \wedge \text{agent}(e,x) \wedge \text{theme}(e,\text{dagger}) \wedge \exists\bvec{v}[\placepath(y,\bvec{v})]]$.
In this case, the gesture extemplifies \textit{throw the dagger}, which is brought about by the straightforward additional interpretive assumption that the virtual object $y$ is identified with the dagger. 
%
%
If also a subject were involved, a sentence-based extemplification can be derived, too.
Since in either of these cases the gesture remains informationally vacuous compared to speech, this leeway of interpretation can be neglected.

While direct extemplification of the affiliate is the default case of the linguistic interpretation of gestures (cf. \cref{subsec:visuo-spatial}), there are exceptions, such as, for example, the staircase example (\ref{ex:spiral-treppen}), re-given in (\nextx).
\ex 
A: I think that should be \staircases staircases 
\label{ex:spiral-staircases-regiven}
\xe 

The affiliate of the gesture is the noun \textit{staircases}. 
\textit{Staircases}, as hypernym, does not have a \cvm. 
Different kinds of stairs are distinguished by form, however.\footnote{See, for example, \url{https://en.wikipedia.org/wiki/Stairs\#Forms}, accessed \printdate{2024-10-31}.}
Accordingly, the gesture can be construed as extemplifying a shape property (e.g., \enquote{helical}; but there are other interpretations available, see \cref{subsec:visuo-spatial}), and thereby allows to infer a hyponym denoting a certain kind of stair.
That is,  extemplified predicate and affiliate expression can diverge. 
This means that there can be other semantic relationships between the extemplified predicate and the affiliate in addition to identity.   
Schematically, using the template $\alpha[\gamma/\beta]$ for a multimodal utterance $\alpha$ which includes an affiliate $\beta$ and gesture $\gamma$:
\ex \emph{Conditioned interpretation}: \label{ex:gesture-conditioned-affiliation}
\par 
If gesture $\gamma$ is informationally evaluated to mean $p$, then the utterance is interpreted as $\alpha[R(p,\beta)]$.
\xe 

\enquote*{$R(p,\beta)$} in (\lastx) means that $p$, the predicate extemplified by the gesture, is applied to the meaning of the gesture's affiliate expression $\beta$ via a relation $R$.
In the simplest case, $R$ is identity, namely when $p$ directly extemplifies $\beta$ and is informationally evaluated as $\beta$ (i.e., $p = \beta$).
This is the standard case of affiliation, in which the gesture intuitively \enquote{reduplicates} its affiliate -- the informational evaluation of the throwing gesture in (\ref{ex:throw-dagger-exemplification}) and \cref{fig:ex-throw} is an example.
Template (\ref{ex:gesture-conditioned-affiliation}) then gives \enquote*{If the gesture is interpreted as \textit{throw}, then  [\ldots] $R$(\textit{throw}, throw) the dagger}, which, of course, just amounts to \enquote{(someone) throw the dagger}.


For the staircases example (\ref{ex:spiral-treppen}), re-given in (\ref{ex:spiral-staircases-regiven}), several interpretations result from the information evaluation and the conditioned interpretation, including the following:
\pex If the meaning of the gesture is labelled as 
\a \textit{helical}, then the utterance is interpreted as \enquote{[\ldots] that should be $R$(\textit{helical}, staircases)}
\a \textit{tight}, then the utterance is interpreted as \enquote{[\ldots] that should be $R$(\textit{tight}, staircases)}
\a \textit{steep}, then the utterance is interpreted as \enquote{[\ldots] that should be $R$(\textit{steep}, staircases)}
\a \textit{upwards}, then the utterance is interpreted as \enquote{[\ldots] that should be $R$(\textit{upwards}, staircases)}
\xe 

Each informational evaluation brings about a different understanding of the multimodal utterance. 
(\lastx a) can be interpreted as being about a certain type of stairs, namely if $R$ is resolved to something like \textit{type\_of}. 
Resolving $R$ to \textit{form\_of} brings about interpretation of formal properties of stairs, as in (\lastx b,c).
The direction of stairs expressed in (\lastx d) results from resolving $R$ to \textit{direction\_of} (this could further involve to understand that the addressee has to use the stairs in the specified direction).

But how can one derive a relation $R$ in case if $p \neq \beta$? 
%
Here, standard mechanisms from dynamic semantics apply. 
%

\subsubsection{Lexicon-driven speech--gesture interpretation}
\label{sec:lexicon-speech-gesture-integration}

%
Usually, the gesture just extemplifies the affiliate, as seen in the previous section.
But what has extemplification to say about examples where the extemplified predicate does not match the overtly uttered one, as in the spiral staircases example?
We argue that the same kind of implicit meanings 
triggered by minimized contexts such as those in (\nextx) arise if extemplified and affiliated predicate diverge.  
\pex 
\a I can't ride my bike today. The back wheel's tire is flat.
\a The footage shows a man running on stage and stabbing Adamowicz [\ldots]. 
The assailant paces back and forth, arms aloft like a victorious boxer, still holding the 15cm (six-inch) knife.\footnote{Taken from BBC news, \url{https://www.bbc.com/news/world-europe-46878325}, accessed \printdate{2024-01-10}. Pawel Adamowicz was the mayor of Gdansk.}
\xe 

The tire in (\lastx a) is understood as the tire of the bike. 
The knife in (\lastx b) is understood as the instrument of the stabbing event, and pacing back and forth the stabbing action.
Such indirect anaphoric relations are known as \emph{bridging} \citep{Clark:1975}.
Accordingly, the same mechanisms used to resolve bridging anaphora apply in minimized speech--gesture affiliation pairs. 
An approach particularly well-suited to iconic gesture semantics is lexicon-driven and employs \emph{frames} \citep{Fillmore:1968,Fillmore:Baker:2010} to compute inferences on minimized content.\footnote{To extend lexical meaning to gesture integration has been proposed early on by \citet{Rieser:2008}.}
Corresponding lexical extensions have been developed within dynamic semantics, namely within \emph{Discourse Representation Theory} respectively \emph{Segmented Discourse Representation Theory} \citep{Bos:Nissim:2008,Irmer:2013} and \emph{Type Theory with Records} \citep{Cooper:2010-frames,Cooper:2023-ttr-book}. 
Frames can be conceived as stereotypical situation types which are connected to lexical items. 
A word form not only contributes its content, but it also \emph{evokes} the frames it is connected to.
Frame semantics is organized in a frame-base lexicon called FrameNet.\footnote{For a short introduction to FrameNet see \citet{Fillmore:Baker:Sato:2004}, the FrameNet resource is available at  \url{https://framenet.icsi.berkeley.edu/}.}
The lexical entry for \textit{staircase.n}, for instance, is linked to the  connecting\_architecture frame.
The connecting\_architecture frame has a \emph{Part} element as its core element, which is the connecting\_architecture in question, and resolves to the staircase in our example.
Additionally, there are nine non-core elements, namely
\emph{Connected\_locations},
\emph{Creator},
\emph{Descriptor}, 
\emph{Direction},
\emph{Goal},
\emph{Material},
\emph{Orientation},
\emph{Source}, and
\emph{Whole}.
Via \emph{frame evocation} \citep{Irmer:2013}, the content of \textit{staircase} in (\nextx a) is extended by frame elements as in (\nextx b) (using DRT's handy box notation; some frame elements are omitted for reasons of space):
\ex 
a. $\lambda x.$ \SDRS{x}{\text{staircase}(x)} \qquad 
b. $\lambda x.$ \SDRS{x \quad\mid\quad e, y_1, y_2, y_3, y_4}{e : \text{connecting\_architecture} \\ 
\text{Part}(e,x) \\ 
\text{staircase}(x) \\ 
\text{Creator}(e,y_1), y_1=? \\ 
\text{Descriptor}(e,y_2), y_2=? \\  
\text{Direction}(e,y_3), y_3=? \\
\text{Material}(e,y_4), y_4=?
}
\xe

Some remarks are in order here.
The first condition instantiates the connecting\_architecture frame as eventuality $e$.
$e$ as well as the discourse referents of the non-core arguments $y_{1\ldots4}$ are merely implicit and are separated from the regular discourse referents of the universe (vertical bar in the top row).
Implicit discourse referents were introduced by \citet{Kamp:Rossdeutscher:1994:a}, and a related distinction between discourse referents introduced by speech and those introduced by gesture has been argued for by \citet{Lascarides:Stone:2009:a}.
Frame evocation, thus, can be seen as a computational implementation of lexical presupposition triggering \citep{Kamp:Rossdeutscher:1994:b}. 
Since in this particular frame example, the part \emph{is} the connecting\_architecture talked about, all conditions that apply to $e$ also apply to $x$. 

If the gesture in 
(\nextx):
\ex 
Inside the hall was an imposing \staircases staircase. 
\xe 
is interpreted as \textit{spiral}, then information-evaluation conditioned utterance interpretation from (\ref{ex:gesture-conditioned-affiliation}) 
returns \enquote*{$R(\textit{spiral},\text{staircase})$}.
%
%
This information adds to the conditions of the frame-wise extended content of \textit{staircase}:
\ex 
$\lambda x.$ \SDRS{x, z \quad\mid\quad e, y_1, y_2, y_3, y_4}{e : \text{connecting\_architecture} \\ 
\text{Part}(e,x) \\ 
\text{staircase}(x) \\ 
\text{Creator}(e,y_1), y_1=? \\ 
\text{Descriptor}(e,y_2), y_2=? \\ 
\text{Direction}(e,y_3), y_3=? \\
\text{Material}(e,y_4), y_4=? \\
\text{spiral}(z) \\
R(\text{spiral}(z),\text{staircase}(x)), R=?
}
\xe 

The implicit discourse referents in (\nextx) are supposed to be filled by content of different kinds: $y_1$ is likely to be an individual, $y_4$ a substance, $y_3$ a direction, and $y_2$ some property.
%
Since \textit{spiral} 
is a shape predicate, the only plausible frame element to resolve $R$ is $R=$ \emph{Descriptor}.
We arrive at the following multimodal meaning, given that the gesture is informationally evaluated to mean \textit{spiral} and affiliated to \textit{staircase}:
\ex 
$\lambda x.$ \SDRS{x, z \quad\mid\quad e, y_1, y_2, y_3, y_4}{e : \text{connecting\_architecture} \\ 
\text{Part}(e,x) \\ 
\text{staircase}(x) \\ 
\text{Creator}(e,y_1), y_1=? \\ 
\text{Descriptor}(e,y_2), y_2= \text{spiral}(z), z=x \\ 
\text{Direction}(e,y_3), y_3=? \\
\text{Material}(e,y_4), y_4=?
}
\xe

The frame-extended predicate in (\lastx) is processed as usual in further semantic composition.\footnote{On harmonizing DRT and Montague-style semantics, see \citet{Muskens:1996,Zeevat:1989-drt}, although the natural framework of (\lastx) is SDRT \citep{Asher:Lascarides:2003-sdrt}.}
%

The same mechanism derives examples discussed elsewhere, such as (\nextx), taken from \citet[p.~303]{Schlenker:2018}.
\ex 
John [\textit{slapping gesture}] punished his son.
\xe 

If we interpret the gesture as slapping, with \textit{punished} being the lexical affiliate, then the multimodal information package \enquote*{$R(\textit{slapped},\text{punished})$} is obtained. 
The lexical unit \textit{punish.v} evokes the Rewards\_and\_punishment frame.
Thus, the lexical content in (\nextx a) is frame-wise extended to include the implicit content in (\nextx b) (slightly abbreviated; the agent role maps to frame element \enquote*{Agent}, the patient role to \text{Evaluee}; cf. \citealp{Irmer:2013}):
\pex 
\a $\lambda y. \lambda x. \lambda e.$ \SDRS{y,x,e}{\text{punish}(e) \\
\text{agent}(e,x) \\
\text{patient}(e,y)}
\a $\lambda y. \lambda x. \lambda e.$ \SDRS{y,x,e \quad\mid\quad z_1, z_2, z_3, z_4, z_5}{e : \text{rewards\_and\_punishment} \\
\text{punish}(e) \\
\text{Agent}(e,x) \\ 
\text{Evaluee}(e,y) \\ 
\text{Reason}(e,z_1), z_1=? \\
\text{Degree}(e,z_2), z_2=? \\
\text{Instrument}(e,z_3), z_3=? \\
\text{Manner}(e,z_4), z_4=? \\
\text{Means}(e,z_5), z_5=?
}
\xe 

Being an action-simulating gesture, slapping instantiates the non-core \textit{Means} frame element: punish by slapping, see the resolved content in (\nextx). 
\ex
$\lambda y. \lambda x. \lambda e.$ \SDRS{y,x,e,e',x',y' \quad\mid\quad z_1, z_2, z_3, z_4, z_5}{e : \text{rewards\_and\_punishment} \\
\text{punish}(e) \\
\text{Agent}(e,x) \\ 
\text{Evaluee}(e,y) \\ 
\text{Reason}(e,z_1), z_1=? \\
\text{Degree}(e,z_2), z_2=? \\
\text{Instrument}(e,z_3), z_3=? \\
\text{Manner}(e,z_4), z_4=? \\
\text{Means}(e,z_5), z_5=\text{slap}(e'), e'=e \\
\text{agent}(e',x'), x'=x \\
\text{patient}(e',y'), y'=y
}
\xe 

(\lastx) reads as $x$ punished $y$ by slapping $y$.
This captures Schlenker's intended reading for this example \citep[p.~303]{Schlenker:2018}, but is even stronger than the local context actually induced in his approach. 
The local context one gets without further stipulation is \textit{If John punished his son, then slapping would be involved} \citep[p.~318]{Schlenker:2018}.
But the slapping could be slapping someone else -- think of John punishing his son by slapping the son's pet.
In (\lastx) this is captured by a non-maximal interpretation and dispensing condition \enquote*{$y'=y$}.
(The same kind of objection applies to Schlenker's helping-by-lifting example, which has an additional comitative reading.)


Furthermore, presuppositional accounts do not seem to prevent speech--gesture mismatches.  
Take, for instance, (\nextx), where the spiral gesture is replaced by a slapping one:
\ex 
Inside the hall was an imposing [\textit{slapping gesture}] staircase. 
\xe 

This gesture obviously does not extemplify its affiliate \textit{staircase}.
Furthermore, slapping, denoting an action, is not a good candidate to fill any of the frame elements evoked by \textit{staircase} in (\blastx). 
Hence, frame-based dynamic semantics algorithms would fail to integrate speech and gesture in this case and signal a mismatch.
This does not seem to hold for other approaches such as those resting on local contexts and assertion-dependent presuppositions, since nothing prevents local contexts of the form \enquote{\textit{every world $w$ in which a staircase is in the hall is one in which slapping is involved}} from being computed \citep[cf.][p.~318\psq]{Schlenker:2018}.\footnote{Schlenker, by the way, puts a lot of weight on computability. But we would argue that his alleged \enquote{algorithm} for generating local contexts and iconic presuppositions is not computable at all. The main reason is that it operates on possible worlds. Possible worlds, however, are too much for human, cognitive processing \citep{Partee:1977-possible-worlds}, and since, according to philosophical arguments, they lack criteria of individuation and thereby countability \citep{Rescher:1999-possible-worlds}, they presumably resist any computational access. (We owe these arguments to \citet[p.~238\psqq]{Cooper:2023-ttr-book}.) But this might be okay from a purely technical point of view. The FrameNet approach at least \emph{is} computable (for an approach close to the present formal model see \citealp{Hou:Markert:Strube:2018-bridging}; for a recent overview including state-of-the-art neural models see \citealp{Poesio:et:al:2023-anaphora}).\label{fn:pw-computability}}

In sum, informational evaluation, conditioned interpretation, and, if required, frame-based integration of speech and gesture provides a systematic heuristic for analyzing iconic gesture in semantic research. 
It also avoids the pitfalls that inferential approaches tend to fall into: in contrast to the latter, it is computable, and provides a notion of multimodal incongruence. 


\subsubsection{Issueness}
\label{subsec:issueness}

Informational evaluation is a heuristic act, or a descriptive hypothesis of the interlocutor, gesture researcher, or bystander (see \cref{subsec:who-is-interpreting}) about the linguistic interpretation of a gesture.
This hypothesis can be confirmed or rejected in dialogue, as shown in \cref{subsec:visuo-spatial} and below.
The hypothetical character of the linguistic interpretation of multimodal utterances is reflected in the conditioned interpretation heuristic (\ref{ex:gesture-conditioned-affiliation}), which literally puts the understanding of a multimodal utterance in the consequence of an indicative conditional. 
This has consequences for the information status of informationally evaluated utterance interpretation.  
In particular, a gesture fails to contribute linguistic content -- at least unless interlocutors agree on an informational evaluation of the gesture, as has been pointed out in \cref{subsec:visuo-spatial}. 
This can be shown by using the nondeniability test for gestures \citep{Ebert:2014:a,Schlenker:2019-gestural}, as evinced in (\nextx) and (\anextx):
%
\pex \label{ex:staircase-nondeniability}
\a A: Inside the hall was an imposing \staircases staircase. 
\a \ljudge{\#} B: No, that is not true. The staircase was actually $\langle * \rangle$.
\xe 

As pointed out in \cref{subsec:visuo-spatial}, 
the correction lacks a linguistic antecedent, indicated by the \enquote{wild card}.
An antecedent can be established if interlocutors agree on an informational evaluation of the gesture:
\pex \label{ex:staircase-agreed}
\a A: Inside the hall was an imposing \staircases staircase.
\a B: Do you mean a spiral staircase? 
\a A: Yes.
\a B: No, that's not true. The staircase was actually straight.
\xe 

Utterances (\lastx b,c) lift the gesture to a quasi-linguistic level, where it can be taken up by speech.

Non-deniability is a feature of non-at-issue contributions.\footnote{And it has indeed been claimed from such tests that gestures contribute non-at-issue content \citep{Ebert:2014:a,Schlenker:2019-gestural}. However, this follows only if an informationally evaluation is stipulated \emph{and} if failing to contribute at-issue content is the same as contributing non-at-issue content; the latter has been pointed out by \citet[p.~326\psq]{Hunter:2019}. We do not subscribe to this implicit assumption, since according to iconic gesture semantics the gesture without being informationally evaluated simply remains sub-linguistic, visual.\label{fn:original-sin}}
However, other contexts induce non-deniable contents, too, in particular the antecedents of indicative conditional sentences.
Their consequences cannot be picked out by negation:
The negation of a sentence of the form \enquote{If $A$ then $C$} is either the conjunction \enquote{$A$ and not $C$} or the conditional \enquote{If $A$ then not $C$} \citep[cf.][]{Egre:Politzer:2013-conditionals}.\footnote{A weak version of denial of indicative conditionals has \enquote{[\ldots] then possibly not $C$} as consequence.}
A denying continuation of an indicative conditional targets the asserted implication:
\pex  A: If the staircase is spiral, it is an imposing one.
\a\ljudge{\#} B: No, that's not true. The staircase is imposing.
\a B: No, that's not true. The staircase is imposing even without being spiral. 
\xe 

Hence, we would expect contexts of conditioned, but not explicitly agreed, gesture interpretation to involve nondeniable consequences.
Accordingly, (\nextx) is in line with the predictions of the conditioned interpretation theory:
\pex  \textit{If \staircases is interpreted as \enquote{spiral}, then} in the hall was an imposing spiral staircase.
\a \ljudge{\#} No, that's not true. The staircase was actually straight.
\a No, that's not true. The staircase was actually straight, even if you interpret \coil[0.25] as \textit{spiral}. 
\xe 

Things are different if the antecedent condition is fulfilled, which happens if the gesture is lifted to a quasi-linguistic status due to explicitly agreed informational evaluation, as in  (\ref{ex:staircase-agreed}). 
From \enquote{If $A$ then $C$} and \enquote{$A$}, \enquote{$C$} follows and can be negated. 

\citet{Ebert:2024} distinguishes two additional non-at-issue tests for co-speech gestures, projection, and ellipsis.
Let us turn to each in turn.
The first one tests the fact that non-at-issue contents project across sentential operators:
\pex 
\a It is not the case that in the hall was an imposing \staircases staircase.
\a \ljudge{\#} No, the staircase was actually straight.
\xe 

Again, this behaviour is explained by conditioned interpretation:
\pex 
\a \textit{If \staircases is interpreted as \enquote{spiral}, then} it is not the case that in the hall was an imposing spiral staircase.
\a \ljudge{\#} No, the staircase was actually straight.
\xe 

The second test says that the co-speech gesture contribution is ignored in ellipsis constructions:
\ex
In the hall was an imposing \staircases staircase, and a window, too.
\xe 

Ignoring for now that resolving elliptical constructions is a complex process in itself \citep[see, e.g., ][]{Ginzburg:Cooper:2004}, then a conditional context produces a straightforward interpretation of the multimodal utterance, where \textit{spiral} does not need to take scope over \textit{window}:
\ex
\textit{If \staircases is interpreted as \enquote{spiral}, then} in the hall was an imposing spiral staircase, and a window, too.
\xe 

Hence, the informational evaluation of iconic gestures can \emph{explain} observations concerning the information status of iconic gestures wrongly attributed to (non-)at-issueness elsewhere.\footnote{Some authors had the intuition that the ellipsis test brings about a different result for pro-speech gestures \citep{Schlenker:Chemla:2018}. As will be discussed in \cref{sec:discussion-infeval} by the example of non-lexicalized iconic models, we show that there is no clear difference in the information status of co- and pro-speech gestures. The latter, when taken to be produced in purpose for the sake of communication (i.e., as \emph{foreground} not as \emph{background} gesture; \citealp{Cooperrider:2017}) may invoke a stronger obligation for informational evaluation, however.} 
%
Rather
a \enquote{linguistic lift} in terms of informational evaluation of a gesture is needed for the gesture to be able to interact with linguistic meaning within the interpretive context of the gesture researcher. 
%
%

\subsubsection{Evaluation of labeling approaches}
\label{subsubsec:eval-labeling}


\textit{Pace} \citet{Ebert:2024,Schlenker:2018} and others, there is no evidence that the semantic contribution of gestures pertain to the at-issue or non-at-issue level of linguistic meaning (see \cref{sec:introduction} and \cref{subsec:issueness}, and footnote \ref{fn:original-sin}).
Of course, this does not mean that multimodal information is not gradual.
We recognize two dimensions of graduality: (i) the prominence or salience of gesture uptake,
and (ii) the ease or necessity of informational evaluation.

In \cref{subsec:who-is-interpreting}, we saw that speakers and listeners do not always see and interpret the gestures of their conversation partners. 
However, there are feature of multimodal utterances that make gesture uptake more likely.
One of these is the prominence of gesture production: a gesture performed in the center of gesture space is considered more intentional and more important to the conversation \citep{Cooperrider:2017}. 
Another feature is a gesture produced in the context of a verbal demonstrative (e.g., \textit{like this}): a demonstration directs the attention of the addressee to an aspect of the context \citep{Luecking:2018-witness}.
A third feature is the timing of the gesture in relation to its affiliate (if any). 
Temporal alignment in the sense of an overlap between a gesture and an accentuated affiliate can be considered standard \citep[e.g.,][]{Alahverdzhieva:Lascarides:2011}.
Deviations from such co-speech gestures are therefore marked. 
Pre-speech gestures (gestures performed before the spoken affiliate) and post-speech gestures (gestures performed after their affiliate) are such deviations. 
Accordingly, they invite increased gesture uptake. 
The semantics of gesture uptake consists in the iconic model of a gesture instantiating a conceptual vector meaning and spatially restricting interpretation.
It should be noted that gesture uptake and the construction of a multimodal meaning (here: a visuo-linguistic meaning) can differ between interlocutors, which explains potential informational asymmetries between speaker and addressee.
This is because each dialogue partner has their own take on the conversation \citep{Traum:Larsson:2003,Ginzburg:2012}, which is particularly evident in misunderstandings.


In addition to (visual) gesture uptake, can informational evaluation can be performed, that is, the classification of a gesture based on a linguistic predicate.
Like gesture uptake, informational evaluation can also be more or less obvious or pressing. 
An important aspect is conventionalization, which we have characterized in \cref{subsec:conventionalization} as the strength of the association between an expression and a visual perception (in the case of iconic gestures, an iconic model).
The more conventional a gesture is, the easier it is to classify.
For example, pantomime is considered quite conventional.
Omitting an affiliate, as in pro-speech gestures, invites the verbal gap to be filled in to form a complete sentence. 
This already explains observations of pro-speech gestures mistakenly attributed to issueness elsewhere (\citealt{Schlenker:2018}, for instance, constructs his examples with conventionalized pantomimic gestures, cf. \cref{subsec:who-is-interpreting}).

As has been emphasized several times, agreement between interlocutors on a classification of gestures in the context of dialogue is the most reliable way to achieve an informational evaluation.
This contextual effect can be used in experiments involving \enquote{gesture guessing}. 
For example, the study described in \citet{Esipova:2019} introduces the verbal alternation \textit{small} and \textit{large} in the experimental instructions, which is then exploited by participants to interpret holding gestures in terms of sizes (see \cref{fig:hold-carriers}).
Accordingly, the priming of participants explains gesture interpretation and not the alleged degree of issueness of gestures.

\subsubsection{Applications}
\label{subsec:applications}

\begin{figure}[htb]
\begin{adjustbox}{center}
    \begin{tikzpicture}
    \node [rectangle] (fig) {\includegraphics[width=4.5cm]{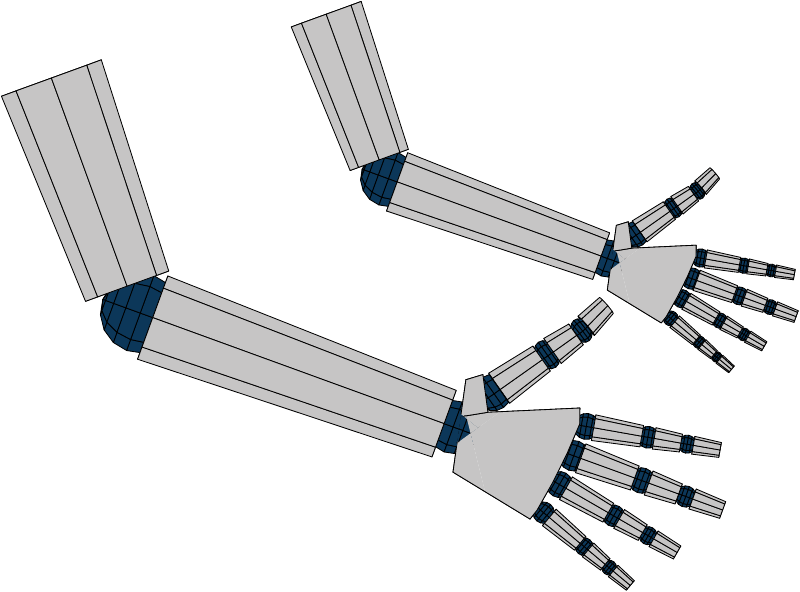}};
    \draw [red!80!black, very thick, shorten >=1.4cm] ([xshift=-0.8cm]fig.east) -- (fig.south) node [pos=0.3] (x) {};
    \node [right=1.6cm of x, align=left, yshift=0.75cm] (theme) {\emph{Space between hands:} \\ virtual theme}; 
    \draw [->, dotted] (theme.west) -- +(215:1cm);
    \node [right=of x, align=center, yshift=-0.45cm] {\emph{Free ride:} \\ distance $d$} edge [->, dotted] (x);
    \node [above=1.3cm of x, xshift=0.5cm, align=center] (img) {\emph{Visual image:} \\ hold-\cvm$(\gamma)=1$};
    \draw [dotted, ->] (img) -- +(270:0.9cm);
    \draw [dotted, ->] (img) -- +(247:1.8cm);
    \node [above=1.7cm of x, xshift=-2.4cm, align=center] (img) {\emph{Agent:} \\ speaker};
    \end{tikzpicture}
\end{adjustbox}
\caption{Visual image of holding or touching an item. A distance (red line) comes as a \emph{Free Ride} as studied in diagrammatic reasoning \protect\citep[e.g.,][]{Shimojima:2015}.}%
\label{fig:hold-carriers}
\end{figure}

In this section, we apply the labeling heuristic to some examples in order to illustrate how it is supposed to work and contribute to multimodal semantic analyses.
To start with, the gesture displayed in \cref{fig:hold-carriers} evokes the visual image of a holding event. 
\textit{Holding} is a two-place predicate with the meaning represented in (\nextx a) and the extemplification mappings are explicated in (\nextx b).
Being a static action, no vector path is involved in this case. 
\pex 
\a $\denotation{hold} = \lambda y. \lambda x. \lambda e [\text{hold-\cvm}(e)=1 \wedge \text{agent}(e,x) \wedge \text{theme}(e,y)]$
\a Bijective iconic mappings:\\
-- hold-{\cvm}($\gamma$) $\mapsto 1$ (the gesture looks like a holding posture) \\
-- speaker/gesturer $\mapsto \text{agent}(e)$\\ 
-- space between hands $\mapsto \text{theme}(e)$ (i.e., the theme remains virtual, or presupposed)
\xe 

Every content part of \textit{hold} can be mapped onto morphological or visually presupposed features of the gesture.
Hence, the gesture successfully extemplifies \textit{hold}.

Now, this type of gesture has been claimed to mean \textit{large} (the interpretations following \enquote*{$\Rightarrow$} are taken from   \citealp[p.~118]{Esipova:2019}, respectively \citealp[p.~304]{Schlenker:2018}):
\ex \label{ex:larg-mistake}
S. might bring her \gestlarge\ \bringbox{dog.} \\
\#$\Rightarrow$\enquote{S. might bring her large dog} / \enquote{If S. brings her dog, it will be large}
\xe 
This interpretation does not seem to live up to iconic gesture interpretation, however, hence the \enquote{\#}.
There is, apart from implementational details, consensus that \textit{large} is an adjective that is lexically associated with a measurement scale and a standard of comparison, as is expressed in (\nextx a) \citep[see][]{Kennedy:2007,Morzycki:2009}.
But can this meaning be extemplified by the sample gesture? 
The gesture, in any case, extemplifies the property of \emph{being this large} (\textit{d-large} from (\nextx a)) by means of the spatial distance $d$ between the hands (this from a \emph{Free Ride} type of inference from diagrammatic reasoning, \citealp{Shimojima:2015}).
Comparing (\lastx) and the visual image displayed in \cref{fig:hold-carriers}, however, it is apparent that the gesture does not extemplify the standard of comparison.
The standard is not an \emph{intrinsic} property of (virtual or real) sizing actions, hence \textit{large} is not a fully visual property (as is already indicated by a lack of a \textsc{cvm}).
\pex 
\a $\denotation{large} = \lambda x [\text{standard}(\text{large}) \leq \text{large}(x)]$, \\
where \textit{large} is a measure function $\lambda x. \rotatebox[origin=c]{180}{\ensuremath{\iota}} d. [x \textit{ is } d\textit{-large}]$ of type $\langle e, d \rangle$
\a Bijective iconic mappings: \\
-- there is no {large-\cvm}! \\
-- (from \emph{Free Ride}) distance $d \mapsto d$-\textit{large}\\
-- \textbf{?} $\mapsto$ standard (and it is unclear what to do with the agent)
\xe

Extemplification of \textit{this large} already involves a visual inference \textit{via} a free ride from \textit{holding something} to \textit{sizing}.\footnote{The sizing mode of representation has been observed by \citet{Kendon:2004} in the Grappolo gesture family; it has also been used in iconic gesture annotation \citep{B1:Sagaland}.} 
Arriving at \textit{large} -- if possible at all without further information in context
-- needs further inferential processing.\footnote{Note that the verbal affiliates in such examples are concrete nouns like \textit{bottle} or \textit{dog}. The lexical meaning of both expressions does not refer to tallness (or even size). This analysis confirms the judgements reported by \citet[p.~322]{Hunter:2019}, namely that a \textit{large} reading of the gesture is unavailable for such examples.}

An additional trigger for pragmatic reasoning is the absence of a \enquote{lexical affiliate} \citep{Schegloff:1984}, which is usually associated with an iconic gesture and guides its interpretation \citep{Hadar:Krauss:1991}.
The imagined affiliate in case of (\ref{ex:larg-mistake}) is the word \textit{large} and its absence already indicates that the example is somewhat deviant: if \textit{large} actually \emph{is} the important property in question, then it should be produced in an information-structurally distinguished way, that is, being the focus expression.
Omitting the focused predicate leads to a pragmatically infelicitous utterance.
Pragmatically infelicitous utterances in turn trigger specific implicatures (e.g., evasion moves; \citealp{Ginzburg:et:al:2023-response-space}).
Hence, such examples are particularly ill-suited to tell apart asserted and implied contents.

Example (\nextx) is taken from \citep[p.~761]{Schlenker:2019-gestural}.\footnote{The original example has a picture of a body posture labelled with \textsc{unscrew-ceiling}. We believe that our comparatively neutral verbal description is more likely to prevent us from a premature informational evaluation.}
\ex 
This light bulb, are you going to [\textit{speaker stretches arm overhead and rotates hand}]
\xe 

Schlenker interprets the gesture as unscrewing at the ceiling.
How can we derive this interpretation?
\textit{unscrew.v} gives us the following lexical information and evokes the \emph{Closure} frame:
\ex 
$\denotation{unscrew} = \lambda y. \lambda x. \lambda e [\text{unscrew-\cvm}(e) =1 \wedge \text{agent}(e,x) \wedge \text{theme}(e,y) \wedge e : \text{Closure}]$
\xe 

The \emph{Closure} frame contributes a non-core \emph{Place} element, which the gesture extemplifies to \textit{overhead}.
This is what we get from gesture semantics.
Schlenker claims further that the multimodal utterance involving the pro-speech gesture in (\blastx) triggers the presupposition that the light bulb is at the ceiling. 
Since this is not part of the multimodal meaning, the location of the lamp should not be a presupposition -- speaking in terms of FrameNet: the \emph{Ground} element of the \emph{Location\_of\_light} frame is not extemplified.
Accordingly, we expect to find a situation where it does not hold.
And indeed, stretching the arm over the head might also be necessary to unscrew light bulbs from other kinds of tall lamps, as shown in \crefrange{subfig:floorlamp}{subfig:tablelamp}.
Hence, the impression that the light bulb is at the ceiling as in \cref{subfig:ceilinglamp} cannot be a presupposition of (\lastx) -- it is just an artifact of gesture interpretation (which in turn might be influenced by a defeasible abductive inference to a common location of light bulbs fixed overhead, though).

\begin{figure}[htb]
\centering
\subfloat[Ceiling lamp\label{subfig:ceilinglamp}]{%
\begin{tikzpicture}
    \draw [thick] (0,0) rectangle (3,3);
    \draw [ultra thick, line cap=butt] (1.5,3) -- +(225:0.65cm);
    \draw [ultra thick] (1.5,3) -- +(315:0.65cm);
    \node [scale=1.2, rotate=180, anchor=south, inner sep=1pt] at (1.5,3){\faLightbulb[regular]};
 
    \filldraw [fill=black!60, draw=black!80, very thick] (1.5,0.75) -- (3,0.75) -- (3,1.5) -- (2.25,1.5) -- cycle;
    \draw [line width=2pt] (1.8,0) -- (1.8,0.75);
    \draw [line width=2pt] (2.9,0) -- (2.9,0.75);
\end{tikzpicture}
}
\qquad
\subfloat[Floor lamp\label{subfig:floorlamp}]{%
\begin{tikzpicture}
    \draw [thick] (0,0) rectangle (3,3);
    \draw [ultra thick, line cap=butt] (1.1,2.7) -- +(225:0.65cm);
    \draw [ultra thick] (1.1,2.7) -- +(315:0.65cm);
    \node [scale=1.2, rotate=180, anchor=south, inner sep=1pt] at (1.1,2.7){\faLightbulb[regular]};

    \draw [ultra thick] (0.2,0) -- (0.2,2.1) arc[start angle=180, end angle=74, radius=0.7cm] -- +(270:0.1cm);

    \filldraw [fill=black!60, draw=black!80, very thick] (1.5,0.75) -- (3,0.75) -- (3,1.5) -- (2.25,1.5) -- cycle;
    \draw [line width=2pt] (1.8,0) -- (1.8,0.75);
    \draw [line width=2pt] (2.9,0) -- (2.9,0.75);
\end{tikzpicture}
}
\qquad
\subfloat[Table Lamp\label{subfig:tablelamp}]{%
\begin{tikzpicture}
    \draw [thick] (0,0) rectangle (3,3);
    \draw [ultra thick, line cap=butt] (1.7,2.8) -- +(225:0.65cm);
    \draw [ultra thick] (1.7,2.8) -- +(315:0.65cm);
    \node [scale=1.2, rotate=180, anchor=south, inner sep=1pt] at (1.7,2.8){\faLightbulb[regular]};

    \filldraw [fill=black!60, draw=black!80, very thick] (1.5,0.75) -- (3,0.75) -- (3,1.5) -- (2.25,1.5) -- cycle;
    \draw [line width=2pt] (1.8,0) -- (1.8,0.75);
    \draw [line width=2pt] (2.9,0) -- (2.9,0.75);

    \filldraw [ball color=black!30, thin] (2.4,1.2) -- (2.7,1.2) arc[start angle=0, end angle=180, radius=0.3cm] -- cycle;
    \draw [line width=2pt, black!90] (2.5,1.5) -- (2.8,2.7) -- (1.7,2.8);
\end{tikzpicture}
}
\caption{Different lamps whose lightbulbs have to be unscrewed overhead.}
\label{fig:lamps}
\end{figure}

This shows, we believe, the importance of a semantic toolkit for interpreting gestures in semantics, in particular, if different kinds of inferential meanings are to be kept apart.

\subsection{Discussion}
\label{sec:discussion-infeval}

\subsubsection{Non-lexicalized percepts}
\label{subsec:non-lexicalized}

Since a gesture's linguistic meaning contribution is conditioned on its informational evaluation, we expect no difference in at-issue contexts.
The demonstrative context \textit{like this} in (\nextx) is supposed to shift its referent to at-issue content \citep[p.~303]{Schlenker:2019-gestural}.
\ex 
The staircase looked like this: \coil[0.3]
\xe 

While this works for speech, what is the at-issue contribution of the gesture in (\lastx)? 
You have to informationally evaluate it first, as usual.
That gestures without informational evaluation fail to contribute at-issue (or non-at-issue, for that matter) content can be seen by gestural movements that resist perceptual classification because the trajectory has no lexicalized label.
An example is given in (\nextx).\footnote{At least the authors do not know how to call this shape; it might be possible that some readers think that it resembles something they know and can name. We rely on these readers to appreciate the point of the example nonetheless.}
\ex 
The inscription looked like this: \tikz [baseline, scale=0.3] \draw [thick] (1.5,0.5) .. controls (2,0) and (2.5,0.5) .. (3.5,1) .. controls (4,1.5) and (3,2) .. (2,2) .. controls (1.5,2.5) and (2,3) .. (3,3) -- (4,3) -- (4,1);
\xe 

We do think that the at-issue test still does work, and that it works for gestures.
A gesture that is such that it resists perceptual classification in terms of single words just contributes its iconic model.

Other gesture uses that can be subsumed under non-lexical ones are metaphoric ex(t)emplifications. 
We ignored them so far, but just want to mention that a frame-based approach provides a modelling clue in terms of the blending of frames \citep[MetaNet, ][]{Petruck:Dodge:2016-metanet}.

\subsubsection{Gesture holds}
\label{subsubsec:holds}

While gesture timing in terms of pre-, co-, post-speech gestures does not pose serious challenges for gesture semantics, gesture holds arguably do. 
\citet[64\psq]{Rieser:2024} discusses, among other things, the SaGA example in \cref{fig:teich-1}.
The speaker describes a path through a park; there is a pond in this park. 
The pond is represented by the left hand, which shows a static, circular shape (i.e., a modeling gesture in terms of \cref{sec:gesture-primer}). 
The speaker then uses the right hand to indicate the path around the pond, while continuing to represent the pond with their left hand (gesture hold). 
The algorithmic challenge for compositional semantics is that the pond representation is available over a longer period of time; standard compositionality \enquote{consumes up} its arguments only once. 
\citet{Rieser:2024} therefore develops a $\lambda\Psi$ process algebra, where speech and gesture are modeled by agents that communicate over time.

\begin{figure}
\centering
\IncGest{0.3\textwidth}{V5-Teich-7_57-1}\hfill%
\IncGest{0.3\textwidth}{V5-Teich-7_57-3}\hfill%
\IncGest{0.3\textwidth}{V5-Teich-7_57-5}
\caption{[Du fährst] \enquote{um den Teich herum} ([You drive] \textit{around the pond}): Index finger and thumb of left hand form a circle and  right hand with stretched index finger is moved to three quarters around left hand.}
\label{fig:teich-1}
\end{figure}

\subsubsection{Repercussions for semantic theories}
\label{subsec:repercussions}


%
%
%
%
%
%
%
Arguably, the reversed denotation relation needed for ex(t)emplification and perceptual classifications cannot be reconciled with a textbook possible worlds semantics (cf. \cref{sec:perceptual-classification}). 
If this is right,\footnote{And there is ample supporting evidence. To name a few: the representability problem of possible worlds clashes with cognitive tractability \citep{Lappin:2015-curry}, which is a  characterizing feature of extemplification/classification (see also footnote~\ref{fn:pw-computability}); the extensions of classifier-based semantics are (slightly) indeterminate, which is not compatible with fixed universes $D_e$ of quantification of traditional models \citep{Larsson:2020}; the learnability problem in Montagovian models \citep{Zimmermann:2022-extensions} is at odds with classifier learning.} then formal semantic thinking about iconic meaning ultimately requires looking for a different semantic theory.  
A suitable candidate, to our minds, is a \emph{Type Theory with Records} \citep[TTR,][]{Cooper:2023-ttr-book}.
%
TTR incorporates words-as-classifiers \citep{Larsson:2015}, and can be given a probabilistic interpretation \citep{Cooper:Dobnik:Lappin:Larsson:2015-probttr}, which is needed for learning and graded judgements (cf. \textit{above} from \cref{fig:above}).
TTR's modal theory does not assume possible worlds and hence fares better in terms of computability and cognitive interpretability.
%
%
%
%
It also provides the ontology for dialogue semantics \citep{Ginzburg:2012} (recall the importance of clarification interaction discussed in \cref{sec:introduction}). 
After all, as is widely recognized, dialogue is \enquote{ecological niche} of multimodal interaction \citep{Holler:Levinson:2019,Luecking:Ginzburg:2023-leading-voices}.
Thus, the contour not only of a compositional but also of a \emph{computational} (which implies cognitive) semantic theory of iconic gestures is emerging, which will be further elaborated in future work.



\section{Conclusions}
\label{sec:conclusion}

Visual communication turns out to be a big obstacle to formal semantic modelling; in fact, it forces us to leave the well-trodden paths of Frege/Montague-style possible worlds semantics and to adopt a procedural, classifier-based notion of meaning. 
It is this repercussion of iconic gestures, we have argued, that makes visual communication a theoretically interesting object of research in formal semantics and the philosophy of language in the first place. 

As a workaround for interpreting gestures in semantic analyses, a heuristic has been developed. 
The heuristic is informed by extemplification, an extended variant of Goodmanian exemplification, and consists of three ingredients: 1. \emph{informational evaluation} of gestures according to the meaning associated with extemplified predicates, 2. \emph{conditioned interpretation} of multimodal utterances based on step 1., and 3. \emph{frame-based} resolution of potentially indirect relations involved in step 2.
This heuristic paves the way for systematic, empirically informed gesture studies within formal semantics and related fields. 
The linguistic interpretation of iconic gestures complements visuo-spatial, truth-functional models of the content of iconic gestures, which we only alluded to here. 
Together, however, a computationally implementable iconic gesture semantics emerges, that enables cross-talk between formal gesture semantics and cognitive science. 

This is, of course, not to say that the iconic gesture semantic theory is a theory of speech--gesture production. 
But it should be emphasized that it is compatible with most multimodal production models (apart from the important fact that it incorporates empirical findings on (the lack of) informational evaluation, see \citet{Gullberg:Kita:2009} and the above discussion). 
For instance, the \emph{Sketch Model} \citep{de_Ruiter:2000} assumes an abstract spatio-temporal representation alongside verbal ones.
This clearly corresponds to our separation of iconic models and spatial frames of reference, respectively the lexicon. 
The \emph{Lexical Access Model} \citep{Krauss:Chen:Gottesmann:2000} emphasizes a gesture's facilitation of word retrieval \citep[see also][]{Hadar:1989}; the underlying connection between iconic models and verbal items is precisely captured by the extemplification relation. 
The \emph{Interface Model} rests on a continuous \enquote{negotiation process} between verbal and gestural production channels \citep{Kita:Ozyurek:2003}. 
The iconicity relation mediates between these two generation streams, but as it stands is not a temporal, incremental process.
It is less clear to our minds, however, how iconic gesture semantics relates to the heterogeneous \enquote{multimodal idea units} postulated by the \emph{Growth Point Model} \citep{McNeill:Duncan:2000}. 
In any case, we believe that it is an advantage that our iconic gesture semantics not only captures semantic facts about gesture but is also compatible with a wide range of empirical findings and cognitive, psycholinguistic models.

\appendix  


\printbibliography

\end{document}